\documentclass[sigconf, nonacm, pdfa]{acmart}

\usepackage[a-2b]{pdfx}
\usepackage{amsmath}
\usepackage{bm}
\usepackage{graphicx}
\usepackage{apxproof}
\usepackage{etoolbox}
\usepackage{enumitem}
\usepackage{algorithmic}
\usepackage{multirow}
\usepackage{amsthm}
\usepackage{booktabs}
\usepackage{xspace}
\usepackage{amsmath,stmaryrd,pict2e,picture}
\usepackage{xcolor}
\usepackage{colortbl}
\usepackage{tikz}
\usepackage{bbold}
\usepackage{multirow}
\usepackage{diagbox}
\usepackage{balance}
\usepackage{float}
\usepackage[noend,lined,boxed,vlined,ruled,linesnumbered]
{algorithm2e}
\usepackage{makecell}
\usepackage{array}
\usepackage{listings}

\newcolumntype{C}[1]{>{\centering\arraybackslash}p{#1}}

\definecolor{bestgreen}{RGB}{200, 230, 201}

\definecolor{myyellow}{HTML}{fcefd2}
\definecolor{mydarkyellow}{HTML}{FAD02E}

\definecolor{mygreen}{HTML}{d2fce6}
\definecolor{mypurple}{HTML}{f2d2fc}
\definecolor{peach}{HTML}{FFDAB9}
\definecolor{fpred}{RGB}{255,99,99}

\definecolor{rank1}{HTML}{388E3C} 
\definecolor{rank2}{HTML}{FBC02D} 
\definecolor{rank3}{HTML}{E64A19} 

\newcommand{\pill}[2]{%
    \tikz[baseline=(X.base)] \node[
        fill=#1, 
        rounded corners=3pt, 
        inner sep=2pt, 
        minimum width=3.2em, 
        text height=1.5ex, 
        text depth=0.25ex, 
        align=center, 
        anchor=base,
        text=black,
    ] (X) {#2};%
}
\newcommand{\sci}[2]{%
    \ensuremath{#1{\cdot}\scriptscriptstyle 10^{-#2}}%
}

\newcommand{\first}[1]{\pill{rank1!35}{\textbf{\boldmath{#1}}}}
\newcommand{\second}[1]{\pill{rank2!35}{#1}}
\newcommand{\third}[1]{\pill{rank3!35}{#1}}

\newcommand{\firstT}[1]{\colorbox{rank1!35}{\textbf{#1}}}
\newcommand{\secondT}[1]{\colorbox{rank2!35}{#1}}
\newcommand{\thirdT}[1]{\colorbox{rank3!35}{#1}}

\usepackage{caption, subcaption}

\usepackage[labelformat=simple]{subcaption}
\SetKwInOut{Input}{input}
\SetKwInOut{Output}{output}

\newcommand{\codeq}[1]{{\tt {\small ``#1''}}}
\newcommand{\qq}{\textquotedbl}

\DeclareMathOperator*{\argmax}{arg\,max}
\DeclareMathOperator*{\argmin}{arg\,min}

\newcommand{\code}[1]{\texttt{\small #1}}
\newcommand{\hide}[1]{}

\newcommand{\at}{\textsc{Auto-Fill}\xspace}

\newcommand{\mk}{$\mathcal{M}_K$\xspace}
\newcommand{\mr}{$\mathcal{M}_R$\xspace}
\newcommand{\mc}{$\mathcal{M}_C$\xspace}
\newcommand{\T}{$\mathcal{T}$\xspace}

\newcommand{\excelFiftyK}{Pub-XLS\xspace}
\newcommand{\gov}{Gov-CSV\xspace}
\newcommand{\pbi}{Pub-BI\xspace}
\newcommand{\webwiki}{Pub-Wiki\xspace}
\newcommand{\parquet}{Git-Parquet\xspace}
\newcommand{\gitcsvsmall}{Git-CSV\xspace}

\newcommand{\enterprisecosmos}{Ent-CSV\xspace}
\newcommand{\enterpriseexcel}{Ent-XLS\xspace}
\newcommand{\webtable}{Pub-Web\xspace}
\newcommand{\ar}{Rel-AR\xspace}
\newcommand{\fd}{Rel-FD\xspace}
\newcommand{\st}{Rel-ST\xspace}

\newcommand{\ignore}[1]{}

\newtheorem{df}{Definition}
\newtheorem{ex}{Example}
\newtheorem{pr}{Proposition}
\newtheorem{lemm}{Lemma}
\newtheorem{lem}{Theorem}

\newenvironment{example}{\begin{ex} \nopagebreak
\begin{rm}}{{\hfill$\Box$}\end{rm}\end{ex}} 

\newenvironment{definition}{\begin{df} \nopagebreak
\begin{rm}}{{}\end{rm}\end{df}}

\newcommand{\myparagraph}[1]{\vspace{0.25em}\noindent \textbf{#1.}}

\def\withnotes{1} 
\ifnum\withnotes=1
    \newcommand{\addcite}[1]{\textcolor{red}{[cite]}}
    \newcommand{\todo}[1]{\textcolor{red}{TODO: #1}}
    \newcommand{\yurong}[1]{\textcolor{orange}{#1}}
    \newcommand{\new}[1]{\textcolor{red}{#1}}
\else
    \newcommand{\addcite}[1]{\xspace}
    \newcommand{\todo}[1]{\xspace}
    \newcommand{\yurong}[1]{\xspace}
    \newcommand{\new}[1]{\xspace}
\fi

\newtoggle{full}
\toggletrue{full}

\newcommand\vldbdoi{XX.XX/XXX.XX}
\newcommand\vldbpages{XXX-XXX}
\newcommand\vldbvolume{19}
\newcommand\vldbissue{11}
\newcommand\vldbyear{2026}
\newcommand\vldbauthors{\authors}
\newcommand\vldbtitle{\shorttitle} 
\newcommand\vldbavailabilityurl{https://github.com/lyrain2001/auto-fill}
\newcommand\vldbpagestyle{empty}

\newcommand{\rev}[1]{{\color{black}#1}}
\colorlet{BLUE}{blue}
\colorlet{BLACK}{black}

\definecolor{claudered}{rgb}{0.80,0.05,0.25}

\colorlet{CLAUDERED}{claudered}

\begin{document}


\title{Auto-Fill: Learning to Predict Missing Values Accurately with Specialist Language Models}

%
\author{Yurong Liu}
\authornote{Part of work done while at Microsoft Research.}
\affiliation{%
  \institution{New York University}
}
\email{yurong.liu@nyu.edu}


\author{Yeye He}
\affiliation{%
  \institution{Microsoft Research}
}
\email{yeyehe@microsoft.com}

\author{Haoyu Dong}
\affiliation{%
  \institution{Microsoft Research}
}
\email{hadong@microsoft.com}

\author{Junjie Xing}
\affiliation{%
  \institution{Microsoft Research}
}
\email{junjiexing@microsoft.com}

\author{Shi Han}
\affiliation{%
  \institution{Microsoft Research}
}
\email{shihan@microsoft.com}

\author{Dongmei Zhang}
\affiliation{%
  \institution{Microsoft Research}
}
\email{dongmeiz@microsoft.com}

\author{Surajit Chaudhuri}
\affiliation{%
  \institution{Microsoft Research}
}
\email{surajitc@microsoft.com}

\begin{abstract}

Predicting missing cell values in tabular data is a fundamental problem in data cleaning. 
While state-of-the-art reasoning models show great promise in predicting missing values in tables, by reasoning holistically across rows and columns, they are costly to deploy at scale and tend to be overconfident, often generating hallucinated or false-positive predictions.

In this paper, we observe that achieving high precision missing-value prediction in tables requires a distinct combination of three capabilities: (1) world knowledge, 
(2) text-based reasoning, 
and (3) code-based reasoning. 
We systematically explore design choices for combining these capabilities,  and propose an \at approach that post-trains three specialist small language models (SLMs), each optimized for one capability. We develop a calibrated ensemble mechanism that either dynamically selects the most confident specialist or abstains, ensuring high accuracy.

Extensive experiments on 11 benchmarks with 2200 real tables drawn from diverse domains show that \at achieves superior accuracy compared to state-of-the-art reasoning models (e.g., o3-pro, Gemini 3 Pro, and DeepSeek R1), while operating at a fraction (less than 1\%) of the cost of these frontier models. Our results highlight the effectiveness of specialization and calibrated abstention in the important domain of tabular data. 
\iftoggle{full}
{
\at is publicly available at \href{https://github.com/lyrain2001/auto-fill}{https://github.com/lyrain2001/auto-fill}.
}
{

}
\end{abstract}

\maketitle

\iftoggle{full}
{

}
{
\pagestyle{\vldbpagestyle}
\begingroup\small\noindent\raggedright\textbf{PVLDB Reference Format:}\\
\vldbauthors. \vldbtitle. PVLDB, \vldbvolume(\vldbissue): \vldbpages, \vldbyear.\\
\href{https://doi.org/\vldbdoi}{doi:\vldbdoi}
\endgroup
\begingroup
\renewcommand\thefootnote{}\footnote{\noindent
This work is licensed under the Creative Commons BY-NC-ND 4.0 International License. Visit \url{https://creativecommons.org/licenses/by-nc-nd/4.0/} to view a copy of this license. For any use beyond those covered by this license, obtain permission by emailing \href{mailto:info@vldb.org}{info@vldb.org}. Copyright is held by the owner/author(s). Publication rights licensed to the VLDB Endowment. \\
\raggedright Proceedings of the VLDB Endowment, Vol. \vldbvolume, No. \vldbissue\ %
ISSN 2150-8097. \\
\href{https://doi.org/\vldbdoi}{doi:\vldbdoi} \\
}\addtocounter{footnote}{-1}\endgroup

\ifdefempty{\vldbavailabilityurl}{}{
\vspace{.3cm}
\begingroup\small\noindent\raggedright\textbf{PVLDB Artifact Availability:}\\
The source code, data, and/or other artifacts have been made available at \url{\vldbavailabilityurl}.
\endgroup
}
}

\begin{sloppy}

\section{Introduction}
\label{sec:intro}
Missing values are prevalent in tabular datasets,  which can arise due to missed data entry, unavailable data observation, or errors introduced during data integration~\cite{cleandata-survey-1, cleandata-survey-2}. Prior studies report that up to 45\% of real-world tables contain missing cells~\cite{missing-val-frac-1, missing-val-frac-2}, making predicting missing values one of the key tasks in data cleaning~\cite{cleandata-survey-1, cleandata-survey-2, hua2007cleaning, chai2020importance}.

Recent advances in end-user data cleaning, such as the built-in cleaning capabilities in widely used spreadsheet software like Excel~\cite{excel-clean-data} and Google Sheets~\cite{google-smart-cleanup}, create new opportunities for deploying missing-value prediction directly within tables used by billions of users, making the problem especially important. Example~\ref{ex:motivating-ex} shows an example of this capability in spreadsheet settings.

\begin{example}
\label{ex:motivating-ex}
User Alice is viewing the spreadsheet in Figure~\ref{fig:ex-knowledge}, which lists statistics for past Super Bowl games, including \codeq{Date}, \codeq{Winning team}, and \codeq{Score}, etc. 
She notices a missing value in cell \codeq{C6} and wants to fill it to improve data quality for analytics, a task that traditionally requires substantial manual effort.

Prediction from surrounding table context can simplify this task. 
For \codeq{C6}, the system can predict the winning team and its overall record, format the result consistently with the column (e.g., \codeq{Seattle Seahawks (2, 1-1)}), and present it as a ``card'' in the side pane. Alice can then quickly review and accept the recommendation rather than researching and entering the value manually.
\end{example}

\begin{figure}[!t]
\centering
    \includegraphics[width=1\columnwidth]{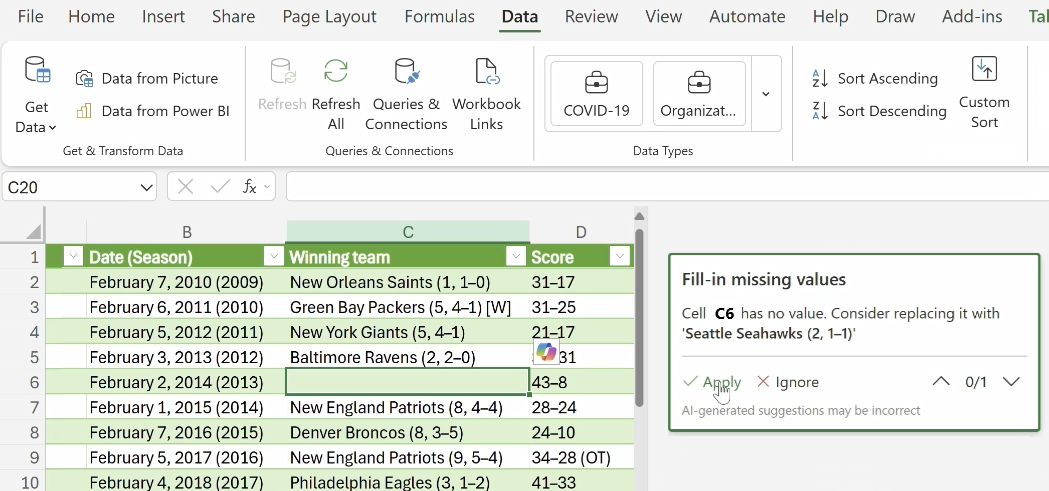}
\caption{Example of \at in spreadsheet software: confident missing-cell predictions appear as ``suggestion cards'' (right pane) for users to review and accept. Correctly predicting \codeq{C6} (winner of Super Bowl 2014) requires [\underline{Knowledge}].}
\label{fig:ex-knowledge}
\end{figure}

\begin{figure*}[t]
\centering
\begin{subfigure}[t]{1\columnwidth}
    \centering
    \includegraphics[width=\linewidth]{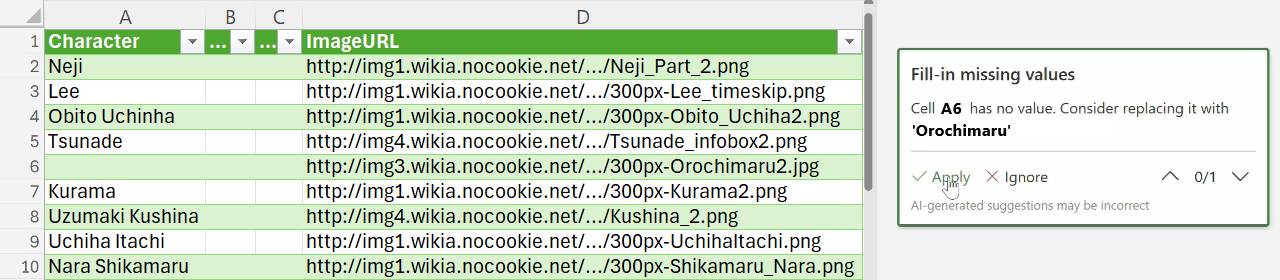}
\end{subfigure}
\hfill
\begin{subfigure}[t]{1\columnwidth}
    \centering
    \includegraphics[width=0.95\linewidth]{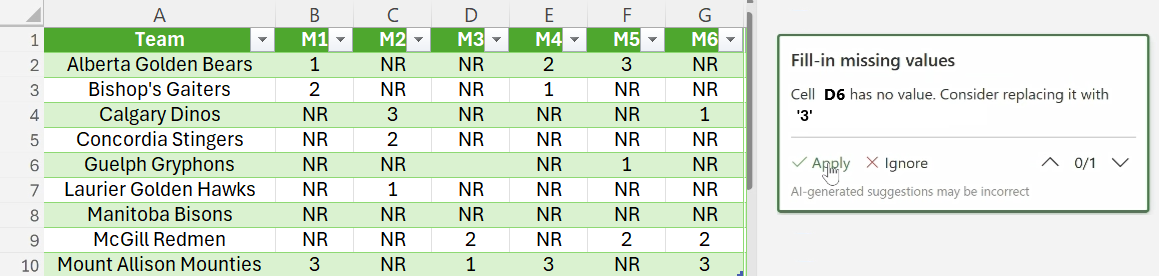}
\end{subfigure}
\caption{Real tables where [\underline{Reasoning}] is required (see Example~\ref{ex:reasoning-figure}). \underline{Left (inter-column)}: the missing name in \code{A6} (\codeq{Orochimaru}) is embedded ad-hoc within the \codeq{ImageURL} string. \underline{Right (intra-column)}: each column holds \codeq{1}/\codeq{2}/\codeq{3} once, else \codeq{NR}, so \code{D6}=\codeq{3}.}
\label{fig:ex-reasoning}
\end{figure*}

\begin{figure}[t]
\centering
    \includegraphics[width=1\columnwidth]{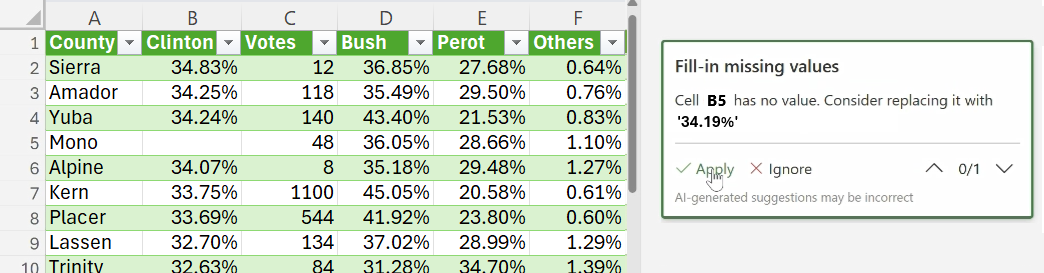}
\caption{Real table where [\underline{Coding}] is required (see Example~\ref{ex:coding-figure}): cell \code{B5} (\codeq{34.19\%}) is computed from the implicit relation \codeq{Clinton}=\codeq{100\%}$-$\codeq{Bush}$-$\codeq{Perot}$-$\codeq{Others}.}
\label{fig:ex-coding}
\end{figure}

\textbf{Unique combination: Knowledge, reasoning, and coding.}
While the example above requires relevant ``knowledge'', which seems well suited for today's language models, we emphasize that filling missing values in tabular data is \emph{far more than retrieving facts} from models' internal knowledge. Rather, it is a task that requires a distinctive combination of knowledge, reasoning, and coding capabilities, as we show below.

\begin{example}
Figure~\ref{fig:ex-reasoning} presents two real tables containing missing cells that require non-trivial reasoning to resolve. In the left table, cell \code{A6} in the \codeq{Characters} column is missing. Notably, several columns to the right include an \codeq{ImageURL} field, whose URL strings embed character names in ad-hoc formats. By leveraging this subtle ``\emph{inter-column}'' relationship, one can infer the missing character name in \code{A6} (in this case, the correct value is \codeq{Orochimaru}).

In the right table, cell \code{D6} is missing. Although this may not be immediately apparent, each column actually follows a consistent pattern: the values \codeq{1}, \codeq{2}, and \codeq{3} each appear exactly once, representing the rankings of the top-3 teams in a given event, while the remaining team is labeled \codeq{NR} (not ranked). By recognizing this implicit ``\emph{intra-column}'' pattern, one can confidently infer that the missing value in \code{D6} is \codeq{3}.
\label{ex:reasoning-figure}
\end{example}

In both examples above, factual “knowledge” plays little role in predicting the missing values, as these are really niche facts that language models either cannot recall exactly or tend to hallucinate~\cite{sun2024head}. Instead, it is far more effective to predict by leveraging implicit inter-column and intra-column patterns that exist in the table in these cases, based on the surrounding table context.

Given the need to ``reason'', and the recent rise of reasoning models (e.g., OpenAI o1, DeepSeek-R1, Gemini 3 Pro), in our initial tests, we found reasoning models to be quite effective for examples like those in Figure~\ref{fig:ex-reasoning}. For instance, DeepSeek-R1 is able to generate detailed, step-by-step textual reasoning for the right-hand example in Figure~\ref{fig:ex-reasoning}, ultimately arriving at the correct missing value. Figure~\ref{fig:prompt} (Left) illustrates a representative reasoning trajectory 
(``\textit{Okay, let's check \ldots Wait, for each column, we see exactly one team ranked as 1, 2, 3 \ldots So the only missing value for Guelph should be 3}'').

In addition to text-based reasoning, code is another important ``mode'' that needs to be used to infer missing values in tables. 

\begin{example}
Figure~\ref{fig:ex-coding} shows another example table, where cell \code{B5} is missing. In this case, based on the values in the table, one could infer an implicit inter-column relationship, namely \codeq{Clinton} + \codeq{Bush} + \codeq{Perot}  + \codeq{Others} =  \codeq{100\%}. It is therefore possible to create a small code snippet to calculate the missing value in \code{B5}.

It is worth noting that, unlike the example in Figure~\ref{fig:prompt} (Left), which relies on text-based reasoning, the example in Figure~\ref{fig:ex-coding} involves precise numerical computation and is better suited for code-based inference, as illustrated in Figure~\ref{fig:prompt} (Right). In this case, the generated code must be executed over the table to produce the predicted missing value. (Although text-based reasoning can also carry out calculations in natural language, it is susceptible to calculation errors, particularly when the computations are complex).
\label{ex:coding-figure}
\end{example}

\textbf{Challenges and key requirements.} The task of filling missing values in tables we study \rev{poses} the following unique requirements: 

(1) \underline{High precision}:  In our task of imputing missing values in tabular data (e.g., user spreadsheets or business-critical tables), prioritizing high precision is essential. The system should generate predictions only when they are highly likely to be correct. Repeated inaccurate suggestions not only burden users (since each suggestion requires manual verification) but can also contaminate the underlying tables, causing issues in downstream analytics.

\rev{Achieving high precision necessitates reliable confidence estimation, yet vanilla LLMs are known to be systematically overconfident and tend to generate hallucinated predictions even under substantial uncertainty~\citep{xiong2023can, zhang2024calibrating}. We therefore need to train models to not only predict missing values, but also produce well-calibrated confidence scores. Producing calibrated confidence is a well-known challenge in LLMs~\citep{geng2024survey}, and a key focus of this work.}

(2) \underline{Low cost}: Given that tables with missing values are prevalent, and predictions like those in Figure~\ref{fig:ex-knowledge} need to be generated at scale (e.g., for all tables containing missing cells so users can review them), maintaining low inference cost is essential, which makes a direct use of frontier models too costly\footnote{Applying frontier reasoning models to predict missing values for all active spreadsheet tables in Excel is estimated to cost over tens of millions of dollars per day.}. At the same time, while smaller language models (e.g., 8B open-source models) are significantly cheaper, they often yield substantially lower quality. Achieving the best of both worlds, with high-quality predictions at low cost, is therefore a key goal that we aim to achieve in this work.

(3) \underline{{Multi-mode}}: 
As the examples in Figure~\ref{fig:ex-knowledge},~\ref{fig:ex-reasoning} and~\ref{fig:ex-coding} show, predicting missing values in different types of tables requires approaches of different ``modes'' (knowledge/reasoning/coding). An effective solution must therefore seamlessly integrate the complementary ``modes'' in order to make accurate predictions.

\textbf{Our approach: \at.} In this work, we develop an \at approach that post-trains small language model (SLM) specialists \rev{, each specializing} in knowledge/reasoning/coding, respectively, for the missing value prediction task. These specialist models are then combined holistically using confidence-based ensembles, ensuring high quality while at very low costs as shown in Figure~\ref{fig:recall_vs_cost}.

Our work makes the following contributions:    \begin{itemize}[leftmargin=*]
        \item We develop and release a suite of 11 benchmark datasets curated from diverse tabular sources, enabling systematic comparison of missing-value prediction methods.
        \item We design \at, a high-precision, low-cost, multi-mode method that trains and combines specialist SLMs for knowledge, reasoning, and coding.
        \item We test \at using multiple families of language models (Qwen3 and GPT-4.1), demonstrating state-of-the-art performance that surpasses frontier models such as o3-pro and Gemini 3 Pro, while operating at less than 1\% of their cost, which is highlighted in our main result in Figure~\ref{fig:recall_vs_cost}.
        \item We map the design space through extensive experiments, documenting negative results to inform future work.
        \item While reasoning models excel at math and code, their use in tabular settings remains limited. We are among the first to demonstrate the potential of training and adapting reasoning models to real-world tabular use cases. 
    \end{itemize}

\begin{figure}[!t]
\centering
    \includegraphics[width=1\columnwidth]{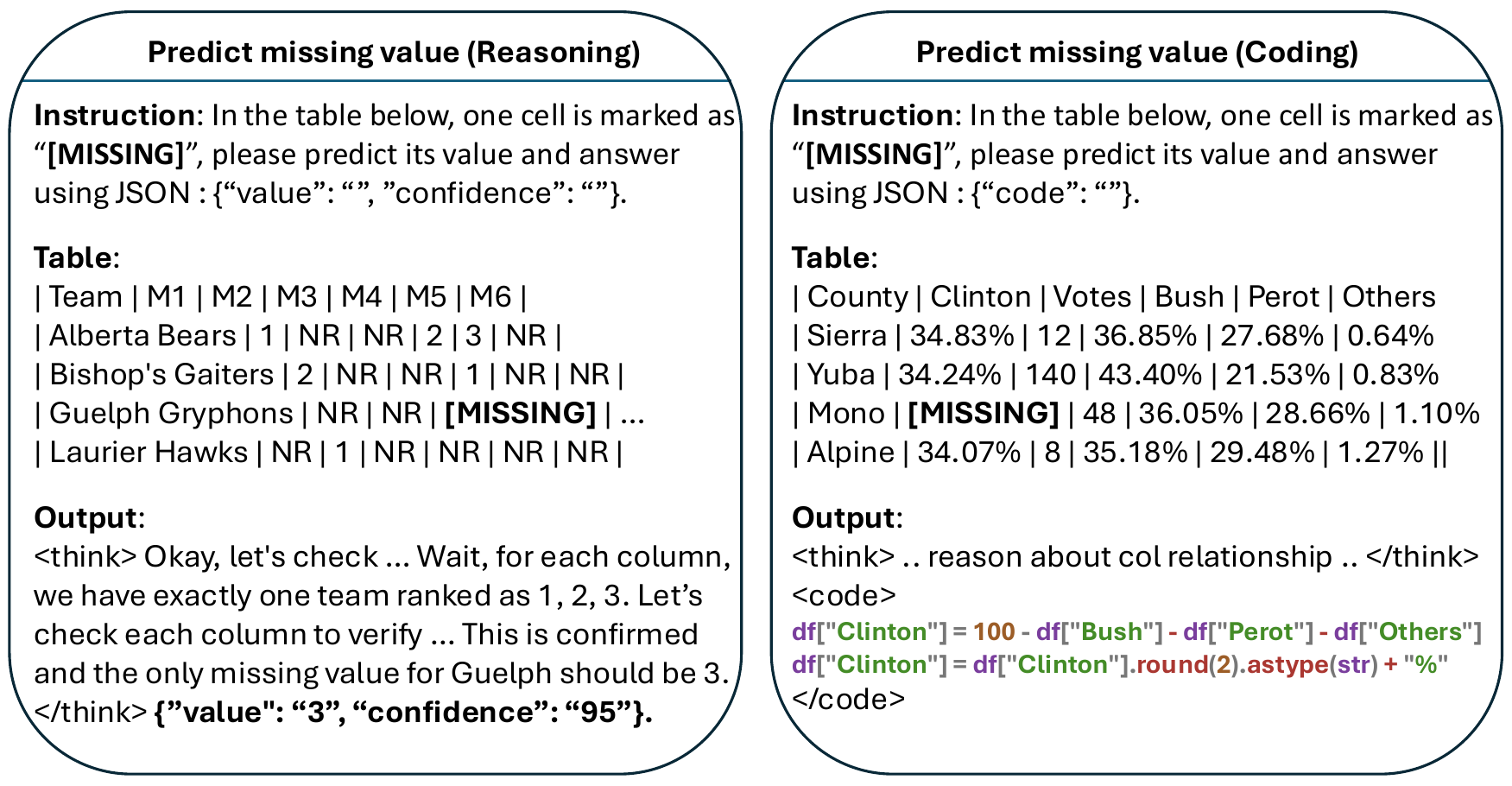}
\caption{Example LLM responses to predict missing values. %
(\underline{Left}): an example prompt corresponding to Figure~\ref{fig:ex-reasoning} (Right), which shows that the LLM needs to ``reason'' before it can answer correctly. \rev{(\underline{Right}): an example corresponding to Figure~\ref{fig:ex-coding}, where the LLM needs to ``code'' to answer correctly.} 
}
\label{fig:prompt}
\end{figure}


\section{Related work}
\label{sec:related}

We review existing research related to our problem in this section.

\textbf{Traditional data cleaning in tables.} There is a long and fruitful line of research on data cleaning~\cite{cleandata-survey-1, cleandata-survey-2, hua2007cleaning, chai2020importance} that leverages formal constraints, such as Functional Dependencies (FDs) and Conditional Functional Dependencies (CFDs), to detect errors and propose fixes (i.e., values to fill in cells). 
While this line of research is highly influential, these methods typically require constraints to be known and provided a priori on specific datasets, which therefore do not generalize to open-ended, spreadsheet-like scenarios (Figure~\ref{fig:ex-knowledge}).
Furthermore, formal constraints are inherently limited in their ability to incorporate world knowledge or perform complex reasoning, thereby restricting their coverage. As we will show experimentally, these limitations place a low upper bound on the achievable recall of constraint-based methods.

\begin{figure}[!t]
\centering
    \includegraphics[width=.9\columnwidth]{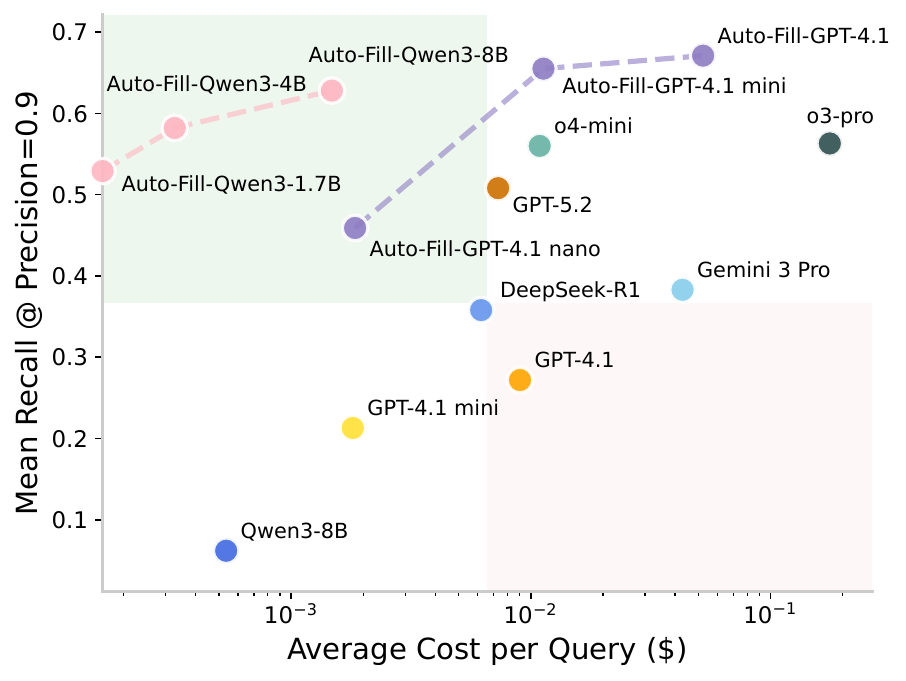}
\caption{Quality vs. cost comparisons: \at variants based on Qwen3 (pink) and GPT-4.1 (purple) form the quality--cost Pareto frontier among the tested models, offering different trade-offs between quality (y-axis, higher is better) and cost (x-axis in log scale, lower is better). \colorbox{bestgreen!60}{\rule{0pt}{6pt}\rule{6pt}{0pt}} indicates the best quadrant (high quality, low cost).}
\label{fig:recall_vs_cost}
\end{figure}

\textbf{Language models for value filling}. Large language models (LLMs), including recent reasoning-oriented models~\cite{r1, o1}, show strong potential for filling missing values in tabular data, and recent works adapt them to diverse table tasks~\citep{zhang2024jellyfish, li2024table, zhou2026can, zhang2024tablellama, xing2024table}. However, they are often overly confident, always producing predictions even when uncertainty is high~\citep{xiong2023can,zhang2024calibrating}, and are prohibitively expensive to deploy at scale (e.g., across all spreadsheet tables with missing cells). 
These limitations make it impractical to apply LLMs out of the box, motivating our work to adapt them into specialized models that are both more accurate and cost-efficient for our task.


\rev{\textbf{Data-lake-based imputation.} A separate line of work fills missing values using an external corpus of related tables: LakeFill~\citep{yang2025lakefill} retrieves candidate values from a data lake, and CESID~\citep{luo2026missing} hybridizes retrieval with model-based estimation. While these approaches are effective when a large data lake contains many similar tables from which missing values can be retrieved, such resources are often unavailable in the general tabular settings we target.
}

\rev{\textbf{Data repair.} The broader repair literature spans two paradigms~\citep{ni2024automatic}: \emph{rule-driven} methods that learn repair actions on top of declared constraints (e.g., BUNNI~\citep{mecca2024bunni}), and \emph{rule-free} methods that learn from observed data distributions (e.g., SCARE~\citep{yakout2013don}, BoostClean~\citep{krishnan2017boostclean}, Baran~\citep{mahdavi2020baran}).  Both, however, rely on intra-table signals and cannot leverage world knowledge or complex reasoning.
}

\textbf{Reasoning models.}
Recent advances in reasoning-oriented models, such as DeepSeek-R1~\cite{r1} and OpenAI o1~\cite{o1}, demonstrate that models post-trained from LLMs using reinforcement learning with verifiable rewards (RLVR) techniques, such as GRPO~\cite{shao2024deepseekmath}, can achieve strong performance on math and coding tasks. 

While there is substantial research on adapting and post-training reasoning models for math and coding tasks, training reasoning models in tabular settings has been limited so far. We are among the first to demonstrate the potential of post-training reasoning models in important real-world tabular scenarios. 

\textbf{Data imputation in Machine Learning literature.} A related ML literature on ``data imputation''~\cite{imputation-1, imputation-2, wang2025llm, he2025llm} aims to fill missing cells with values \emph{as close as possible to the ground truth}, rather than predicting the exact missing value. The guiding principle is statistical utility: for example, mean imputation may estimate a \codeq{sales-quantity} value that is close but not exactly correct, which still improves downstream ML training since most models benefit from approximate but complete inputs.

This sharply contrasts with our setting, where a prediction must be \emph{exactly correct} or the system should abstain. Predicting \codeq{98} for a true \codeq{sales-quantity} of \codeq{100} is unacceptable in spreadsheet and analytics scenarios, since even small inaccuracies propagate into downstream reports. Our problem therefore requires both exact correctness (numeric closeness is insufficient) and high precision.

\section{problem formulation}
\label{sec:problem}
We now formally define the problem of missing value prediction studied in this work.

\begin{definition}[\emph{High-Precision Missing-Value Prediction}]
\label{def:problem}
Let $\mathbf{T} = \{(T_i, r_i, c_i)\}_{i=1}^{N}$ be a set of prediction tasks, where each task
specifies a relational table $T_i$ and a target missing cell $(r_i, c_i)$ with unknown ground-truth value $v_i$. For each task $(T_i, r_i, c_i)$, an uncertainty-aware model $\mathcal{M}$ produces $\mathcal{M}(T_i, r_i, c_i)$, which is either a predicted value $\hat{v}_i$, or an abstention that is denoted by the symbol $\bot$.

Let $\mathcal{P}(\mathcal{M}) = \{i \mid \mathcal{M}(T_i, r_i, c_i) \neq \bot\}$ denote the set of non-abstained predictions produced by $\mathcal{M}$, and
let correctness be determined by exact match, $\mathbf{1}[\hat{v}_i = v_i]$. We define precision and
recall of $\mathcal{M}$ over $\mathbf{T}$ as:
\begin{align}
    \text{Precision}(\mathcal{M}) &= |\{i \in \mathcal{P} (\mathcal{M}) \mid \hat{v}_i = v_i\}| \;/\; |\mathcal{P} (\mathcal{M})|, \\
    \text{Recall}(\mathcal{M})    &= |\{i \in \mathcal{P} (\mathcal{M})\mid \hat{v}_i = v_i\}| \;/\; |\mathbf{T}|.
\end{align}
Given a user-specified precision threshold $\tau \in [0, 1]$, the problem of \emph{High-Precision Missing-Value Prediction} is to find a model $\mathcal{M}$, such that the precision of $\mathcal{M}$ is over the required threshold $\tau$, while $\mathcal{M}$'s recall is maximized, written as:
\begin{equation}
    \mathcal{M}^* = \argmax_{\mathcal{M}} \; \textup{Recall}(\mathcal{M}) \quad \textup{s.t.} \quad \textup{Precision}(\mathcal{M}) \geq \tau.
\end{equation}
\end{definition}

In our high-precision setting, model $\mathcal{M}$ must produce calibrated confidence and \emph{abstain when uncertain}, avoiding hallucinations or ``wild guesses'' that corrupt business-critical tabular data.
\rev{This differs from traditional ML data imputation~\cite{imputation-1, imputation-2} in two fundamental ways. First, we require ``\emph{strict exact match}'' rather than ``numerical closeness'', since even minor deviations (e.g., 99 instead of 100) can distort downstream results. Second, whereas conventional approaches aim to fill all missing values, our high-precision formulation explicitly \emph{encourages abstention} when evidence is insufficient for reliable prediction.}
\rev{The central challenge is therefore obtaining reliable confidence estimates from heterogeneous models (knowledge-, reasoning-, and code-based), enabling abstention that maximizes recall while maintaining high precision, as formalized in Definition~\ref{def:problem}. Section~\ref{sec:overview} presents our approach.}

\section{Overview: Design Space Exploration}
\label{sec:overview}

\begin{figure*}[!t]
\centering
    \includegraphics[width=\textwidth]{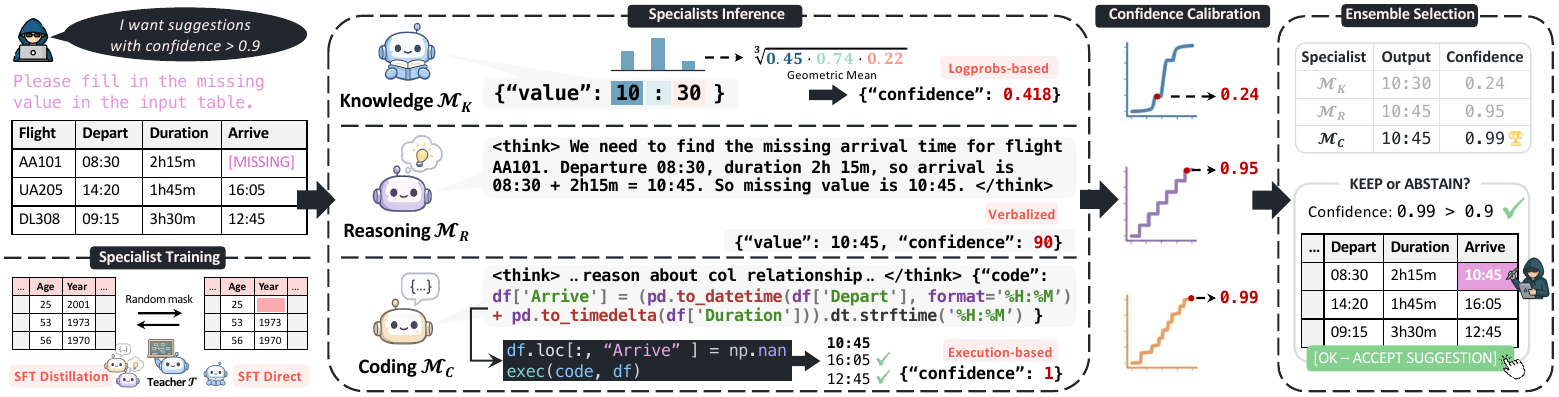}
\caption{\at: our proposed architecture based on an exploration of the design choices in Table~\ref{tab:architecture-options}. \rev{It trains specialist models (middle)}, which are dynamically selected at inference time based on calibrated confidence estimates (right).}
\label{fig:autofill-framework}
\end{figure*}

We present \at, a confidence-aware tabular missing-value prediction framework built on an ensemble of specialized small language models (SLMs). 
We will give a high-level overview of the design space in this section before delving into technical details in subsequent sections. Table~\ref{tab:architecture-options} summarizes the design space options.


\begin{table}[t]
\centering
\small
\caption{\at: Design space choices
}
\label{tab:architecture-options}
\resizebox{\columnwidth}{!}{%
    \begin{tabular}{|c|l|c|c|c|}
    \hline
    \multicolumn{2}{|c|}{\rule{0pt}{8pt}\textbf{Architecture}} &
    \multicolumn{3}{c|}{\rule{0pt}{8pt}\textbf{Options}} \\
    \hline

    \multirow{2}{*}{\makecell{\\One hybrid \\ vs.\ multiple \\ specialist models}}

    & (1) One hybrid model
    & \multicolumn{3}{l|}{
    \rule{0pt}{8pt}\textbf{Training strategy:} (1) Mixed SFT (2) Mixed SFT+RL
    }
    \\
    \cline{2-5}

    & (2) Multiple specialists
    & \rev{\makecell[l]{\rule{0pt}{8pt}\textbf{Modes:}\\
    (1) Knowledge\\
    (2) Reasoning\\
    (3) Coding}}
    & \makecell[l]{\rule{0pt}{8pt}\textbf{Training strategy:}\\
    (1) SFT direct complete\\
    (2) SFT distillation \\
    (3) SFT + RL}
    & \makecell[l]{\rule{0pt}{8pt}\textbf{Ensemble strategy:}\\
    (1) Learned router\\
    (2) Prob. calibration\\
    (3) Classical ML}
    \\
    \hline
    \end{tabular}
}
\end{table}


\underline{One hybrid model vs. multiple specialist models?} 
At the architectural level, we can either (1) train a \emph{single hybrid model} capable of natively switching between knowledge-, reasoning-, and coding-based modes, or (2) use \emph{multiple specialist models} each specializing in one mode, that are then dynamically combined through an ensemble mechanism. 

Our findings indicate that combining specialized models consistently outperforms a single hybrid model. When we train a small model that combines these capabilities in a shared parameter set, the model must alternate between fundamentally different generation strategies, which leads to interference~\citep{yu2020gradient,shen2024mome}.
In contrast, it is easier to train separate specialist models focusing on one capability with strong performance. We will give detailed experimental results in this area (Table~\ref{tab:composition_ablation} of our experiments).

\rev{\underline{Which specialist modes?} Our knowledge/reasoning/coding split mirrors the three model archetypes the LLM community has independently converged on — instruction-tuned chat models for direct completion, reasoning models with chain-of-thought~\citep{wei2022chain, o1, r1}, and code-specialized models~\citep{roziere2023code, guo2024deepseek}. Prior modular LLM systems also draw the same split between knowledge retrieval, reasoning, and code execution~\citep{karpas2022mrkl, yao2023react, chen2022program}.
A post-hoc error analysis of our final system supports this split empirically: among 100 sampled failures, 89\% fall inside the knowledge/reasoning/coding decomposition rather than indicating a missing fourth mode, with the remaining 11\% being unrecoverable cases
\iftoggle{full}
{~(Appendix~\ref{appendix:error-analysis})}
{~(see our full paper~\citep{full})}.
}
%

\underline{How to train specialist models?} 
Given that we employ multiple specialist models, the next design decision concerns the training strategy. We consider three options shown in the middle of Table~\ref{tab:architecture-options}: (1)~\emph{direct supervised fine-tuning (SFT)}, in which the model is trained to predict answers directly; (2)~\emph{distillation-based SFT}, where the model is trained on chain-of-thought reasoning traces generated by a stronger teacher reasoning model; and (3)~\emph{SFT + reinforcement learning (RL)}, where SFT is followed with RL to further enhance the models' reasoning abilities. 

Overall, our findings suggest that direct SFT is sufficient when answers use models' parametric knowledge, or require simple pattern matching over tables. However, distillation-based SFT becomes essential for problems involving complex multi-step reasoning or non-trivial code generation. While reinforcement learning (RL) yields modest further gains, it is significantly more expensive to train, so we treat it as an optional enhancement. 

\underline{How to ensemble specialist models?} Given multiple specialists, we need a mechanism to select among their predictions, as the same missing cell may be solvable by different specialist models. We explored three approaches, shown on the right of Table~\ref{tab:architecture-options}: (1)~a \emph{learned router} that predicts which specialist to invoke for a given input (analogous to router models used in ChatGPT~\cite{gpt5-router} to select reasoning vs. chat models); (2)~a \emph{classical ML ensemble} (XGBoost) that takes the confidence scores from specialist models as input and uses classifications to select a prediction; and (3)~\emph{calibrated confidence selection}, where each specialist produces a principled confidence score that is calibrated to true probabilities, with the most confident specialist being selected. 

We find that calibrated confidence selection is the most principled and effective strategy: it directly captures each specialist’s self-estimated prediction reliability, thereby enabling natural abstention when no specialist is sufficiently confident. In contrast, the learned router’s confidence measures certainty about \emph{which specialist to invoke}, rather than \emph{whether the resulting prediction is correct}, which leads to degraded performance in our high precision settings. 
Classical ML XGBoost improves upon the router, but it does not generalize consistently across datasets. Detailed evaluations of these design choices will be presented in our experiments.

\underline{\at: Final design.} Building on the above explorations, our proposed \at adopts a \emph{multi-specialist} architecture, as illustrated in Figure~\ref{fig:autofill-framework}. Our approach uses an ensemble of three complementary models:
(1)~a \emph{knowledge specialist} \mk trained via direct SFT, with confidence estimated from token-level log-probabilities;
(2)~a \emph{reasoning specialist} \mr trained via distillation-based SFT, whose confidence is distilled from sampled reasoning distributions; and
(3)~a \emph{coding specialist} \mc, also trained via distillation-based SFT, which generates Python code and derives confidence through self-validation against the observed column values.

The distillation of \mr and \mc relies on a teacher model \T. In our implementation, we instantiate \T with DeepSeek-R1~\citep{r1}, as it produces open reasoning trajectories (unlike closed-source reasoning models), though any sufficiently capable reasoning model could serve as the teacher.

Confidence signals from \mk, \mr, and \mc are then mapped onto a unified probabilistic scale using principled isotonic calibration to reflect their true probabilities~\citep{zadrozny2002transforming}. At inference time, \at either selects the prediction from the most confident specialist, or abstains if none surpasses a target precision threshold.  

We will now describe the training stage and ensemble stage of \at, in Section~\ref{sec:specialists} and Section~\ref{sec:confidence}, respectively.
\section{Training Specialist Models}
\label{sec:specialists}

This section details the training of three specialist SLMs, each tailored to a distinct capability: knowledge, reasoning, and coding. For each specialist, we describe both its post-training procedure and its corresponding confidence estimation mechanism. Confidence calibration is particularly critical, as LLMs are known to be systematically overconfident, and different reasoning paradigms necessitate different estimation strategies~\citep{geng2024survey, xiong2023can, zhang2024calibrating, yang2024alignment}.


At a high level, all three specialists share a common training data construction process. A key advantage of missing value prediction for tabular data as a learning problem is that ground-truth training data can be generated automatically from complete tables: \rev{for each complete table in our corpus, we randomly select a non-empty cell $(r, c)$ and replace its value with \texttt{[MISSING]}. The masked table serves as input, and the original cell value serves as the ground-truth label. 
Cells already empty in the original table are kept, so the model still sees naturally co-occurring missing values in context.}

\begin{example}
\label{ex:masking} [Masking]. On the left of Figure~\ref{fig:autofill-framework}, we have an example table with flight information. In the self-supervised ``masking'' procedure, we sample a cell from the table, in this case the top-right cell, replace its original value (\codeq{10:45}) with the mask \codeq{[MISSING]}. 

Because we know the ground-truth should always be the original value \codeq{10:45}, which comes “for free”, it enables us to systematically construct training examples for the knowledge, reasoning, and coding specialists, where the final answer is always \codeq{10:45}. This serves as a unified way to ``self-supervise'' and construct diverse training examples for all specialists.
\end{example}

This self-supervised paradigm allows us to construct large-scale training corpora without human annotation. 
From a large shared pool of masked tables, we develop techniques to automatically generate training data for knowledge, reasoning, and coding, respectively, which we will describe in the next three subsections.


\subsection{Knowledge Specialist} 
Not all missing values require complex reasoning. For instance, the missing value in Figure~\ref{fig:ex-knowledge} (the winning team of a Super Bowl game) can be answered using the model’s knowledge, together with necessary formatting, so that the predicted value is consistent with other values in the same column to precisely match the ground-truth. Models can answer cases like this 
without complex reasoning.

%
%

Interestingly, we in fact find that for tasks requiring knowledge, employing chain-of-thought reasoning models can actually harm prediction quality. For example, when we sample Wikipedia tables with well-known facts as test cases, reasoning-oriented models often perform noticeably worse than chat-style models that directly generate completions; this trend persists even after post-training.\footnote{We hypothesize that generating additional intermediate tokens introduces more opportunities for error accumulation and shifts probability mass away from the correct answer token~\citep{liu2024mind}. Moreover, predicted values in tabular settings must adhere to formatting patterns established by other values in the same column (e.g., in Figure~\ref{fig:ex-knowledge}). Reasoning models can fail to maintain such consistency when they are distracted by intermediate reasoning tokens.}

\underline{Capabilities beyond Knowledge.}  While we refer to this specialist as “knowledge” for simplicity, we observe that post-training enhances the model beyond merely retrieving relevant facts from its internal memory. In particular, post-training here improves two capabilities that are critical yet often lacking in vanilla small models.
First, the model must be able to ``align'' a missing cell with other values in the same column, effectively “reading vertically” in the column direction, which is non-trivial especially in large and wide tables~\cite{li2024table,yang2026carelessly}. 
If the model mis-aligns the missing cell and associates it with a different column (which happens often with small models), the resulting prediction can be entirely incorrect.
Second, the model must generate predictions that conform to the formatting patterns established within the column -- e.g., in Figure~\ref{fig:ex-knowledge}, where win/loss/total records follow a specific format, the predicted output must also use the exact same format, a skill that small models can learn to improve.
We observe that post-trained knowledge specialists $\mathcal{M}_K$ substantially enhance both capabilities in addition to better knowledge retrieval.

\underline{Training data generation: direct ground-truth completion.} For the knowledge specialist, \mk,  we therefore train the model to directly predict the ground-truth value in a structured response of the form $\{\text{``value''}: v\}$, where $v$ is the ground-truth that we can automatically construct using our masking procedure in Example~\ref{ex:masking}. 

\underline{Confidence extraction: log probability aggregation.} 
Since \mk generates $v$ as its first substantive output, the token-level log-probability $\text{logprob}(v)$ reflects the model's raw belief over its output vocabulary, unconditioned on any potentially incorrect intermediate predictions~\citep{lanham2023measuring}. 
Log-probabilities in the direct-answer setting have been shown to correlate with factual correctness, making them a reliable confidence signal~\citep{kadavath2022language, xiong2023can}. 

When the predicted value $v$ spans multiple tokens (e.g., multi-word phrases or numeric strings), we compute confidence as the geometric mean of token-level probabilities. Concretely, let $\{t_i\}_{i=1}^n$ denote the tokens comprising $v$. We extract the log-probability of each token from the model’s output distribution and aggregate them to compute the overall confidence, denoted by $\text{conf}_K$, as:
\begin{equation}
\label{eq:knowledge-logprob}
  \text{conf}_K = \exp\!\left(\frac{1}{n}\sum_{i=1}^{n} \text{logprob}(t_i)\right) = \left(\prod_{i=1}^{n} P(t_i)\right)^{1/n},
\end{equation}
where $n$ is the number of tokens in the model-produced answer $v$. 
 
The geometric mean is appropriate because the joint probability of generating the full value factorizes as a product of per-token probabilities in auto-regressive models. Averaging in log-space normalizes for sequence length, preventing longer values from being systematically penalized. The resulting score lies in $[0,1]$ and is the model’s intrinsic confidence in the predicted value.

\underline{Training.} We use standard supervised fine-tuning directly on (table, answer) pairs constructed via the masking procedure in Example~\ref{ex:masking}, similar to table instruction-tuning style training \citep{li2024table, zhang2024tablellama}.

After training, the model learns to predict missing values by not only retrieving relevant facts from its internal knowledge, but also formatting the predicted value in ways consistent with other values in the same column, as in the example of Figure~\ref{fig:ex-knowledge} (where the predicted value needs to follow a particular data format). This is important in our tabular setting, and is an ability that small models learn to improve during training. 

\begin{algorithm}[!t]
\begin{small}
\SetKwComment{Comment}{/* }{ */}
\SetKw{Not}{not}
\SetKw{Or}{or}
\caption{Trace Construction for \mr}
\label{alg:trace-reasoning}
\KwIn{Masked table $T$, target cell $(r,c)$, ground truth $v^*$, teacher \T, sample size $k$}
\KwOut{Training trace with supervision}
\vspace{1pt}
Generate reasoning traces $\mathcal{R}_1 \gets \{\mathcal{T}(T)\}_{i=1}^{k}$ without confidence \;
$\mathcal{C}_1 \gets \{r \in \mathcal{R}_1 : \text{answer}(r) = v^*\}$ \;
\If{$\mathcal{C}_1 = \emptyset$}{\Return $\perp$}
$r^* \gets \argmin_{r \in \mathcal{C}_1} |r|$ \tcp{Shortest correct trace}
Generate reasoning traces $\mathcal{R}_2 \gets \{\mathcal{T}(T)\}_{i=1}^{k}$ with confidence \;
$\text{conf} \gets \mathbb{E}[c \mid \text{answer}(r)=v^*] \cdot P(\text{answer}(r)=v^*)$ over $\mathcal{R}_2$ \;
\Return $(r^*, v^*, \text{conf})$ \;
\end{small}
\end{algorithm}

\subsection{Reasoning Specialist} 
In many other cases, predictions require multi-step complex reasoning, where direct completion can fall short.
Figure~\ref{fig:ex-reasoning} shows two illustrative examples in this category, as discussed earlier.
For these cases, chain-of-thought reasoning is essential.

\underline{Training data generation: reasoning distillation.} 
We train \mr by distilling high-quality chain-of-thought reasoning traces from a teacher model \T. Chain-of-thought prompting~\citep{wei2022chain} substantially improves multi-step reasoning by encouraging models to decompose problems into intermediate steps, but effective chain-of-thought generation typically requires either large models or explicit training on reasoning. We address this through distillation: \T generates reasoning demonstrations, which are used to fine-tune the small \mr student model. Following the \texttt{<think>...</think>} convention~\citep{r1, yang2025qwen3}, \mr generates its reasoning trace within a structured block before producing its final prediction. This separation forces the model to commit to an explicit logical path before answering and makes the reasoning process inspectable.

For each training instance constructed via the masking procedure (Example~\ref{ex:masking}), we prompt \T to generate $k$ independent reasoning traces (with $k=10$, balancing sample diversity and generation cost). We then compare the predicted outcomes from these $k$ traces against the ground-truth value known from masking, retaining only those traces that produce the correct answer. Among the correct traces, we select the one with the shortest reasoning length for training, as it typically reflects a clearer logical structure and lowers inference cost in the resulting \mr.

\underline{Confidence extraction: two-stage sampling.} To obtain reliable confidence estimates for reasoning, a straightforward approach is to use token-level log-probabilities of the final answer, as we do for \mk. However, we find this signal to be unreliable in the reasoning settings. Unlike \mk, the final answer token in a reasoning model is conditioned on the entire preceding chain-of-thought reasoning, which is post-hoc, often making its log-probability artificially high. In fact, we observe that this probability is almost always close to \code{1.0}, rendering it unusable as a confidence signal.

An alternative is to have \T generate reasoning traces that verbalize confidence self-assessment, then use those traces directly for training. However, this creates inconsistencies: the confidence mentioned within a reasoning trace reflects a single sample's self-assessment, whereas the true reliability of that reasoning path can only be estimated from the distribution of outcomes across multiple independent attempts~\citep{podolak2025read,xiong2023can}. If \T verbalizes high confidence within a single reasoning trace but empirically succeeds on that case only occasionally in $k$ attempts, we would be training on a trace that claims certainty while the true reliability is low. 

We address this by decoupling trace selection from confidence computation through a two-stage procedure. For each training case, we first generate the $k$ reasoning traces described above without confidence, yielding a clean distribution of prediction outcomes. Second, we prompt \T to generate $k$ \emph{additional} completions with explicit confidence verbalization. 
From this second set, we compute an aggregate confidence score:
\begin{equation}
    \text{conf}^{\text{train}}_R = \mathbb{E}[\text{conf} \mid \text{correct}] \times P(\text{correct}),
    \label{eq:reason}
\end{equation}
where $\mathbb{E}[\text{conf} \mid \text{correct}]$ is the mean self-assessed confidence across correct completions, and $P(\text{correct})$ is the fraction of attempts that were correct. This product captures two complementary dimensions: how confident the model is when it succeeds, and how reliably it succeeds. A case where \T is consistently correct and consistently confident yields a high score; a case with occasional success or low confidence when correct yields a lower score.

\begin{example}[Two-stage confidence generation for $\mathcal{M}_R$].
Consider the table in Figure~\ref{fig:autofill-framework} (lower left), where the \codeq{Year} for \codeq{Age = 25} is masked (ground truth: \codeq{2001}).

In \underline{Stage 1}, we prompt \T to generate $k=10$ independent reasoning traces without confidence. Eight correctly deduce $2026 - 25 = 2001$. We retain the shortest correct trace for training on this case. 
In \underline{Stage 2}, \T generates $k=10$ additional completions with verbalized confidence, producing a mean confidence of 95. With eight correct traces and the mean verbalized confidence, using Eq.~\ref{eq:reason}, we obtain $95 \times 0.8 = 76$, which is the target confidence we use in the training data for \mr to learn to fit.
\end{example}

This process for training trace generation is detailed in Algorithm~\ref{alg:trace-reasoning}, and
\iftoggle{full}
{
    example \T traces for both stages are in Appendix~\ref{appendix:teacher-traces}.
}
{
    example \T traces for both stages are in our full paper~\citep{full}.
}
Importantly, we pair the selected reasoning trace (from the first sampling round, generated \emph{without} confidence) with the aggregate confidence score (computed from the second sampling round, generated \emph{with} confidence), 
where the confidence is presented as an externally-derived label. Figure~\ref{fig:prompt} (Left) shows an example from this generation process -- given the table from Figure~\ref{fig:ex-reasoning} (Right) with a missing cell, the teacher model \T produces a reasoning trace that arrives at the correct prediction \codeq{3}, together with a confidence estimate \codeq{0.95} following Algorithm~\ref{alg:trace-reasoning}.

\underline{Training.} During distillation-based SFT, \mr is trained to reproduce the reasoning trace before outputting the predicted value:
\begin{equation*}
    \texttt{<think>}\textit{...}\texttt{</think>} \; \texttt{\{``value'': } v\texttt{, ``confidence'': } c \texttt{\}}.
\end{equation*}
Note that the confidence label $c$ is learned from \T's sampling distribution, 
ensuring that \mr learns to produce confidence that reflects true uncertainty rather than post-hoc rationalization~\citep{kadavath2022language}.

\begin{algorithm}[!t]
\begin{small}
\SetKwComment{Comment}{/* }{ */}
\SetKw{Not}{not}
\SetKw{Or}{or}
\caption{Trace Construction for \mc}
\label{alg:trace-coding}
\KwIn{Masked table $T$, target cell $(r,c)$, ground truth $v^*$, teacher \T, sample size $k$, observed row indices $\mathcal{I}_{\text{obs}}$ in column $c$}
\KwOut{Training trace with supervision}
\vspace{1pt}
\For{$i = 1$ \KwTo $k$}{
    Generate code $code_i \gets \mathcal{T}(T)$ \;
    Execute on column $c$ to get result $V_i$ \;
    \If{execution fails \Or $V_i[r,c] \neq v^*$}{\textbf{continue} \;}
    $\text{acc} \gets \frac{1} {|\mathcal{I}_{\text{obs}}|}\sum_{j \in \mathcal{I}_{\text{obs}}} \mathbb{1}[V_i[j,c] = T[j,c]]$ \tcp{Column-level accuracy on observed rows}
    \If{$\text{acc} \geq 0.8$} {\Return $(c_i, v^*)$} \tcp{$code_i$ generalizes to column}
}
\Return $\perp$ \tcp{No valid code found}
\end{small}
\end{algorithm}

\subsection{Coding Specialist} 
A third class of missing values follows column relationships best expressed as code, as opposed to verbose and imprecise text-based reasoning. For instance, in Figure~\ref{fig:ex-coding}, the missing value in cell \code{B5} can be computed via the relationship \codeq{Clinton} = \codeq{100\%} $-$ \codeq{Bush} $-$ \codeq{Perot} $-$ \codeq{Others}. Describing such a relationship and performing the computation in natural language is verbose and error-prone; but in code, they are direct and precise:
\begin{small}
\begin{verbatim}
df['Clinton'] = 1 - df['Bush'] - df['Perot'] - df['Others']
\end{verbatim}
\end{small}

Similarly, mathematical computations (e.g., calculating tax rates based on different regions), domain-specific formulas (e.g., BMI calculations), or unit conversion, etc., are all best expressed as executable code snippets, rather than text-based reasoning. 

\underline{Training data generation: column-level code.} 
We generate training data for \mc by prompting a teacher model \T to first reason about programmatic relationships that exist in a masked table (Example~\ref{ex:masking}), before producing Python code snippets to instantiate the inferred relationships and predict the missing value. The format of the training traces generated by \T is:
\begin{equation*}
    \texttt{<think>}\textit{...}\texttt{</think>} \; \texttt{\{``code'': } c \texttt{\}}
\end{equation*}
Figure~\ref{fig:prompt} (Right) shows a concrete example of a trace with reasoning and a code snippet, for the example in Figure~\ref{fig:ex-coding}.

We note that naive code generation using \T can often produce ``hard-coded'' programs that directly overwrite the predicted value in the missing cell (e.g., \rev{\begin{small}\texttt{df.loc[5, \qq B\qq] = \qq 34.19\%\qq}\end{small}} for Figure~\ref{fig:ex-coding}), which is ad-hoc and does not reflect column-level relationships.

We therefore constrain \T to generate \emph{column-level} code that computes the entire target column as a function of other columns (e.g., 
\rev{\begin{small}\texttt{df[\qq Clinton\qq] = 1 - df[\qq Bush\qq] - df[\qq Perot\qq] ...}\end{small}}).
This has two benefits: it forces the model to discover and leverage column relationships rather than making ad-hoc point-based predictions; and it naturally generalizes to cases where multiple cells in the same column are missing.
Our target code snippet $c$, when executed against the input table, should then compute the entire target column, as in the case of Figure~\ref{fig:prompt} (Right).

For each training case that is a masked table, we iteratively prompt \T to generate reasoning traces and code for up to $k$ times, accepting the first candidate that satisfies three criteria: (1) the code executes without errors, (2) it produces the correct value in the target missing cell, and (3) it produces correct values for at least 80\% of cells in the same target column. The third criterion is crucial as it ensures the generated code generalizes to the entire column (note that we do not require a 100\% match with the target column, because in real tables, a small fraction of cells, such as total and sub-total rows, can deviate from the dominant column pattern). Cases where no valid code is found within $k$ attempts are discarded, as they likely have no programmatic relationships, and are unfit to be used as training data to train $\mathcal{M}_C$. 

\underline{Confidence extraction: execution-based validation.}
Because \mc produces code rather than direct predictions, token-level log-probabilities do not meaningfully reflect whether the code implements the correct transformation. Moreover, unlike \mr, where the model can reason about answer quality during its thinking process, \mc cannot execute code during generation to verify correctness. We therefore derive confidence from \emph{execution-based self-validation}: after generating code, we execute it against the table and measure how well it reproduces the \emph{observed} (non-missing) values in the target column. Formally, let $\mathcal{I}_{\text{obs}}$ denote the set of observed row indices in column $c$, and let $f_{\text{code}}$ represent the function computed by the generated code. The confidence is:
\begin{equation}
    \text{conf}_C = \frac{1}{|\mathcal{I}_{\text{obs}}|} \sum_{i \in \mathcal{I}_{\text{obs}}} \mathbb{1}\bigl[f_{\text{code}}(T[i, :]) = T[i, c]\bigr].
    \label{eq:code}
\end{equation}
%
%
This mechanism is grounded entirely in comparing execution results and observed values in tables. Requiring column-level code makes self-validation meaningful: if the generated code reconstructs 90\% of observed values in the target column, it gives strong evidence that the code snippet reliably captures the underlying column relationship. Conversely, low validation accuracy signals that the code is either incorrect or applies to only a subset of rows, lowering our confidence. When code execution fails entirely (due to syntax errors or runtime exceptions), we assign $\text{conf}_C = 0$. 
Algorithm~\ref{alg:trace-coding} provides the complete process for coding trace generation. 


\begin{example}[Confidence extraction for $\mathcal{M}_C$].
Consider the table in Figure~\ref{fig:prompt} (Right), where the \codeq{Clinton} value for Mono County is missing, and $\mathcal{M}_C$ generates the column-level code shown there. To compute confidence $\text{conf}_C$ via Eq.~\ref{eq:code}, we execute this code and compare its output against the three observed \codeq{Clinton} values (Sierra: \codeq{34.83\%}, Yuba: \codeq{34.24\%}, Alpine: \codeq{34.07\%}). Since the execution results match the cell values across all three rows in the table, we get $\text{conf}_C = 3/3 = 1.0$.
\label{ex:coding-confidence}
\end{example}

\underline{Training.} Using the training data prepared above, \mc is trained similar to \mr. The resulting \mc learns to first generate reasoning trace, and then the corresponding code. 
Confidence estimates are computed via execution-based validation (Eq.~\ref{eq:code}), as in Example~\ref{ex:coding-confidence}.  This ensures that \mc's confidence reflects execution success 
rather than the model's subjective assessment of its own code quality.

\subsection{Optional RL-based Training}
\label{sec:grpo}
\iftoggle{full}{
The distillation-based SFT described above trains \mr and \mc to reproduce the teacher model's reasoning traces and confidence signals. A natural question is whether reinforcement learning (RL) can further improve the specialists on top SFT, similar to what the training of DeepSeek-R1 has shown~\cite{r1}.

We explore this using a particular form of RL known as Group Relative Policy Optimization (GRPO)~\citep{shao2024deepseekmath}, that is shown to be effective in LLMs. GRPO scores multiple sampled completions for the same input and uses relative rankings as the training signal, avoiding the need for a separate reward model. The key design challenge here is the reward function: standard RL for language models typically optimizes for correctness alone, but our setting requires both the prediction and the confidence to be accurate. 

\underline{Calibration-aware reward}
We use RLCR (Reinforcement Learning with Calibration Rewards)~\citep{damani2025beyond}, which augments the standard correctness reward with a Brier-score-based calibration term. For a prediction $\hat{v}$ with confidence $q \in [0,1]$ and binary correctness indicator $\mathbb{1}[\hat{v} = v^*]$, the reward is:
\begin{equation}
    R = \mathbb{1}[\hat{v} = v^*]- \bigl(q - \mathbb{1}[\hat{v} = v^*]\bigr)^2,
    \label{eq:rlcr}
\end{equation}
where the calibration penalty is minimized when $q$ matches the empirical correctness---i.e., when the model states high confidence on cases it gets right and low confidence on cases it gets wrong. This reward encourages the model to jointly improve its predictions and its self-assessment.
For \mr, $q$ is the verbalized confidence (an integer 0--100, normalized to $[0,1]$) from the model's output. For \mc, $q$ is the execution-based column accuracy $\text{conf}_C$ defined in Equation~\ref{eq:code}, which serves as an implicit confidence signal grounded in observable correctness. We do not apply RL to \mk, as its confidence derives from token-level log-probabilities rather than verifiable output signals suitable for reward-based training.

As we will show in our experiments (Table~\ref{tab:grpo_ablation}), RL yields only modest improvements. We believe this is due to our need to balance two objectives (correctness and confidence), which makes the reward function difficult to design and potentially less effective.
}
{
A natural question is whether reinforcement learning (RL) can further improve \mr and \mc on top of distillation-based SFT, similar to DeepSeek-R1's recipe~\cite{r1}. We explore this with GRPO~\citep{shao2024deepseekmath} using a calibration-aware reward (RLCR~\citep{damani2025beyond}) that augments standard correctness with a Brier-score calibration penalty, so the model is rewarded for being both \emph{correct} and \emph{well-calibrated}. Unlike standard RL where a single correctness signal suffices, jointly optimizing prediction accuracy \emph{and} confidence calibration is a harder target, where gains on one can come at the expense of the other~\cite{geng2024survey,leng2024taming,stangel2025rewarding,zhang2024calibrating}. Empirically, RL yields only modest gains in our setting; we therefore treat it as an optional enhancement.
}
\iftoggle{full}
{
    Further training details are in Appendix~\ref{appendix:rl-training}.
}
{
    Full reward formulation and training details are in our technical report~\citep{full}.
}

\section{Calibrated Ensemble of Specialists}
\label{sec:confidence}

The three specialist models we described so far produce confidence signals through  disparate mechanisms: \mk via token log-probabilities, \mr via verbalized probabilities, and \mc via execution, which are not directly comparable. In this section, we describe how to calibrate these into principled true probabilities (Section~\ref{sec:calibration}), in order to perform dynamic selection and abstention (Section~\ref{sec:ensemble}).

\subsection{Confidence Calibration}
\label{sec:calibration}
In the three specialists we trained, token log-probabilities from \mk tend toward binary extremes for common factual queries; verbalized confidence from \mr reflects uncertainty from the teacher model's sampling distribution; and execution accuracy from \mc is inherently discretized by the number of observed rows. Naively selecting the specialist with the highest raw score is therefore unreliable, since a log-probability of 0.8 from \mk does not carry the same meaning as an 80\% execution accuracy from \mc.

To address this, we use isotonic regression~\citep{zadrozny2002transforming}, to calibrate confidence from each specialist into true probabilities. 
%
%
For each specialist $\mathcal{M}_i \in \{$\mk, \mr, \mc$\}$, on a held-out validation set $\mathcal{D}_{\text{val}}$ disjoint from both training and test data, we collect $n$ pairs of raw confidence $c_j$ (normalized to $[0,1]$) and binary correctness label $y_j$, and fit a non-decreasing $g_i : [0,1] \to [0,1]$ that minimizes the empirical calibration error:
\begin{equation}
\label{eq:isotonic}
  \min_{g_i} \sum_{j=1}^{n} \bigl(g_i(c_j) - y_j\bigr)^2 \quad \text{s.t.} \quad g_i(c_j) \leq g_i(c_k) \;\; \forall\, c_j \le c_k.
\end{equation}
The resulting $g_i$ is piecewise-constant, making no parametric assumption about the raw-confidence distribution, which differs sharply across specialists (log-probabilities, verbalized scores, execution accuracies). The calibrated probability $\hat{\text{conf}}_i = g_i(\text{conf}_i)$ then approximates the true correctness probability of $\mathcal{M}_i$, making scores directly comparable across specialists.

\subsection{Ensemble Selection}
\label{sec:ensemble}
At inference time, given a table $T$ with a missing value at position $(r, c)$, we invoke all three specialists in parallel, each producing a prediction and calibrated probability: $(v_K, \hat{\text{conf}}_K)$, $(v_R, \hat{\text{conf}}_R)$, $(v_C, \hat{\text{conf}}_C)$. We select the output from the most confident specialist, or abstain if none of them exceeds a user-specified threshold $\tau$:
\begin{equation}
  i^* = \argmax_{i \in \{K, R, C\}} \hat{\text{conf}}_i, \quad
  v_{\text{final}} =
  \begin{cases}
    v_{i^*} & \text{if } \hat{\text{conf}}_{i^*} \geq \tau \\
    \textsc{Abstain} & \text{otherwise.}
  \end{cases}
\end{equation}

Note that unlike adaptive routing used in ChatGPT~\cite{gpt5-router} that trains a separate routing model to send queries to appropriate models (reasoning vs. chat), our confidence-based selection provides implicit routing based on true probabilities: a specialist suited for a test case naturally produces higher calibrated probability than others, and by selecting the specialist with the highest probability, the system dynamically ``routes'' to an appropriate specialist leveraging its unique specialties (knowledge, reasoning, vs. coding).

\underline{Principled abstention.}
Missing cells cannot always be reliably inferred when inter- or intra-column dependencies are weak or absent. In such cases, all three specialists produce calibrated probabilities below the target precision threshold $\tau$ (Definition~\ref{def:problem}), and \at abstains naturally, a behavior explicitly desirable in business-critical scenarios, in contrast to the hallucinations or wild guesses vanilla LLMs are prone to.

\iftoggle{full}
{
    Example predictions from each specialist model can be found in Appendix~\ref{appendix:specialist-examples}.
}
{
    Example predictions from each specialist model can be found in our full paper~\citep{full}.
}
We use the following example to demonstrate the end-to-end workflow of \at with the specialist ensemble.
\begin{example}[End-to-end ensemble selection]
Consider the flight table in Figure~\ref{fig:autofill-framework}, where the \codeq{Arrive} time for AA101 is missing and the target precision threshold is $\tau = 0.9$.
The three specialists are invoked in parallel. 
\mk produces an (incorrect) answer \codeq{10:30} directly, where its raw log-probability confidence (Eq~\eqref{eq:knowledge-logprob}) of $0.418$ is calibrated into a true probability of 0.24 using Eq~\eqref{eq:isotonic}. \mr reasons to the correct value \codeq{10:45}, with its verbalized confidence calibrated to $\hat{\text{conf}}_R = 0.95$. \mc generates column-level code that derives arrival times from departure times and durations, validates it on the observed rows, and predicts \codeq{10:45} with $\hat{\mathrm{conf}}_C=0.99$.
Because $\hat{\mathrm{conf}}_C$ is the highest and exceeds $\tau$, \at returns \mc's prediction; if all three confidences are below $\tau$, it abstains.
\end{example}

\section{Experiment}
\label{sec:exp}

\newcommand{\pillsm}[2]{%
    \tikz[baseline=(X.base)] \node[
        fill=#1, 
        rounded corners=2pt, 
        inner sep=1pt, 
        minimum width=1.8em, 
        text height=1.3ex, 
        text depth=0.2ex, 
        align=center, 
        anchor=base,
        text=black,
    ] (X) {\small #2};%
}
\newcommand{\firstS}[1]{\pillsm{rank1!35}{\textbf{\boldmath{#1}}}}
\newcommand{\secondS}[1]{\pillsm{rank2!35}{#1}}
\newcommand{\thirdS}[1]{\pillsm{rank3!35}{#1}}

\begin{table*}[!t]
    \centering
    \caption{Quality results of all methods. Each cell reports R@P=0.9\,/\,\rev{pAUPRC}. 
    The \emph{FD upper-bound} row reports full-distribution recall. Cost reflects LLM inference per query only. \firstT{First}, \secondT{second}, and \thirdT{third} best results per column are highlighted.}
    \label{tab:abstain_combined}
    \resizebox{\textwidth}{!}{%
        \begin{tabular}{l|ccccc|cccccc|c|c}
            \toprule
            & \multicolumn{5}{c}{\textbf{In-Distribution (ID)}} & \multicolumn{6}{c}{\textbf{Out-of-Distribution (OOD)}} & & \\
            \textbf{Method} & \textbf{\excelFiftyK} & \textbf{\pbi} & \textbf{\webwiki} & \textbf{\gov} & \textbf{\parquet} & \textbf{\enterprisecosmos} & \textbf{\enterpriseexcel} & \textbf{\webtable} & \textbf{\ar} & \textbf{\fd} & \textbf{\st} & \textbf{Mean ($\uparrow$)} & \textbf{Cost (\$) ($\downarrow$)} \\
            \midrule
            \rowcolor{black!5}\multicolumn{14}{c}{\textbf{\textit{Non-LLM baselines}}} \\
            FD upper-bound$^*$ &
            0.06\,/\,--- & 0.16\,/\,--- & 0.08\,/\,--- & 0.07\,/\,--- & 0.23\,/\,--- &
            0.34\,/\,--- & 0.13\,/\,--- & 0.07\,/\,--- & 0.06\,/\,--- & 0.48\,/\,--- & 0.00\,/\,--- & 0.15\,/\,--- & --- \\
            LakeFill-small & 
            0.00\,/\,0.00 & 0.00\,/\,0.00 & 0.00\,/\,0.00 & 0.00\,/\,0.00 & 0.00\,/\,0.00 & 
            0.00\,/\,0.00 & 0.00\,/\,0.00 & 0.00\,/\,0.00 & 0.00\,/\,0.00 & 0.83\,/\,0.77 & 0.00\,/\,0.00 & 0.08\,/\,0.07 & --- \\
            LakeFill-full & 
            0.00\,/\,0.00 & 0.00\,/\,0.00 & 0.00\,/\,0.00 & 0.00\,/\,0.00 & 0.00\,/\,0.00 & 
            0.00\,/\,0.00 & 0.00\,/\,0.00 & 0.00\,/\,0.00 & 0.00\,/\,0.00 & 0.77\,/\,0.71 & 0.00\,/\,0.00 & 0.07\,/\,0.06 & --- \\
            \rev{TabPFN} & 
            \rev{0.00\,/\,0.00} & \rev{0.12\,/\,0.11} & \rev{0.00\,/\,0.00} & \rev{0.16\,/\,0.15} & \rev{0.26\,/\,0.25} &
            \rev{0.46\,/\,0.45} & \rev{0.30\,/\,0.30} & \rev{0.00\,/\,0.00} & \rev{0.12\,/\,0.12} & \rev{0.68\,/\,0.66} & \rev{0.03\,/\,0.03} & \rev{0.19\,/\,0.19} & \rev{---} \\
            \rev{Baran} & 
            \rev{0.00\,/\,0.00} & \rev{0.07\,/\,0.07} & \rev{0.08\,/\,0.00} & \rev{0.26\,/\,0.25} & \rev{0.30\,/\,0.28} &
            \rev{0.53\,/\,0.52} & \rev{0.48\,/\,0.48} & \rev{0.00\,/\,0.00} & \rev{0.11\,/\,0.11} & \rev{0.70\,/\,0.69} & \rev{0.03\,/\,0.03} & \rev{0.23\,/\,0.22} & \rev{---} \\
            \rev{SCARE} & 
            \rev{0.00\,/\,0.00} & \rev{0.07\,/\,0.06} & \rev{0.00\,/\,0.00} & \rev{0.17\,/\,0.16} & \rev{0.24\,/\,0.22} & 
            \rev{0.45\,/\,0.44} & \rev{0.39\,/\,0.39} & \rev{0.00\,/\,0.00} & \rev{0.12\,/\,0.12} & \rev{0.63\,/\,0.62} & \rev{0.03\,/\,0.03} & \rev{0.19\,/\,0.19} & \rev{---} \\
                        \midrule
            \rowcolor{black!5}\multicolumn{14}{c}{\textbf{\textit{LLMs with verbalized confidence}}} \\
            Qwen3-8B & 
            0.00\,/\,0.00 & 0.43\,/\,0.40 & 0.00\,/\,0.00 & 0.00\,/\,0.00 & 0.00\,/\,0.00 & 
            0.00\,/\,0.00 & 0.26\,/\,0.24 & 0.00\,/\,0.00 & 0.00\,/\,0.00 & 0.00\,/\,0.00 & 0.00\,/\,0.00 & 0.06\,/\,0.06 & \firstS{\sci{5.36}{4}} \\
            GPT-4.1 mini & 
            0.08\,/\,0.08 & 0.27\,/\,0.25 & 0.02\,/\,0.02 & 0.00\,/\,0.00 & 0.15\,/\,0.14 & 
            0.00\,/\,0.00 & 0.12\,/\,0.11 & 0.02\,/\,0.02 & 0.00\,/\,0.00 & 0.81\,/\,0.75 & 0.90\,/\,0.84 & 0.21\,/\,0.20 & \thirdS{\sci{1.81}{3}} \\
            GPT-4.1 & 
            0.31\,/\,0.28 & 0.42\,/\,0.39 & 0.00\,/\,0.00 & 0.00\,/\,0.00 & 0.00\,/\,0.00 & 
            0.00\,/\,0.00 & 0.45\,/\,0.43 & 0.00\,/\,0.00 & 0.00\,/\,0.00 & 0.84\,/\,0.80 & 0.99\,/\,0.97 & 0.27\,/\,0.26 & \sci{9.02}{3} \\
            DeepSeek-R1 & 
            0.21\,/\,0.19 & 0.59\,/\,0.56 & 0.09\,/\,0.08 & 0.16\,/\,0.15 & 0.00\,/\,0.00 & 0.13\,/\,0.12 & 0.00\,/\,0.00 & 0.00\,/\,0.00 & 0.93\,/\,0.88 & 0.87\,/\,0.86 & 0.98\,/\,0.96 & 0.36\,/\,0.34 & \sci{6.21}{3} \\
            o4-mini & 
            0.43\,/\,0.42 & \thirdS{0.67}\,/\,\thirdS{0.64} & 0.13\,/\,0.12 & 0.41\,/\,0.38 & 0.46\,/\,0.44 & 0.42\,/\,0.41 & \secondS{0.66}\,/\,0.62 & 0.16\,/\,0.14 & 0.98\,/\,0.98 & 0.88\,/\,0.85 & 0.99\,/\,0.98 & \thirdS{0.56}\,/\,0.54 & \sci{1.09}{2} \\
            GPT-5.2 & 
            0.02\,/\,0.02 & 0.57\,/\,0.55 & 0.25\,/\,0.23 & 0.49\,/\,0.46 & 0.44\,/\,0.42 & \thirdS{0.59}\,/\,\secondS{0.57} & 0.50\,/\,0.47 & 0.00\,/\,0.00 & 0.88\,/\,0.83 & \thirdS{0.89}\,/\,\secondS{0.89} & 0.98\,/\,0.98 & 0.51\,/\,0.49 & \sci{7.31}{3} \\
            Gemini 3 Pro & 
            0.00\,/\,0.00 & 0.66\,/\,0.63 & 0.00\,/\,0.00 & 0.00\,/\,0.00 & 0.00\,/\,0.00 & 0.00\,/\,0.00 & 0.65\,/\,0.58 & 0.00\,/\,0.00 & 0.98\,/\,0.95 & \firstS{0.94}\,/\,\firstS{0.92} & \firstS{1.00}\,/\,\firstS{1.00} & 0.38\,/\,0.37 & \sci{4.30}{2} \\
            o3-pro & 
            \firstS{0.55}\,/\,\firstS{0.53} & 0.02\,/\,0.02 & \firstS{0.43}\,/\,\firstS{0.41} & \secondS{0.52}\,/\,\firstS{0.50} & 0.54\,/\,0.51 & \firstS{0.63}\,/\,\firstS{0.62} & 0.64\,/\,\thirdS{0.63} & 0.00\,/\,0.00 & \firstS{1.00}\,/\,\firstS{1.00} & \secondS{0.90}\,/\,\secondS{0.89} & \firstS{1.00}\,/\,\firstS{1.00} & \thirdS{0.56}\,/\,\thirdS{0.55} & \sci{1.77}{1} \\
            \midrule
            \rowcolor{black!5}\multicolumn{14}{c}
            {\rev{\textbf{\textit{LLMs with logprob confidence}}}} \\
            \rev{GPT-4.1 mini} & 
            \rev{0.00\,/\,0.00} & \rev{0.55\,/\,0.54} & \rev{0.07\,/\,0.06} & \rev{0.39\,/\,0.37} & \rev{0.47\,/\,0.44} & \rev{0.49\,/\,0.46} & \rev{0.00\,/\,0.00} & \rev{0.21\,/\,0.20} & \rev{0.58\,/\,0.57} & \rev{0.81\,/\,0.79} & \rev{0.95\,/\,0.94} & \rev{0.41\,/\,0.40} & \rev{\sci{1.82}{3}} \\
            \rev{GPT-4.1} & 
            \rev{0.41\,/\,0.40} & \rev{0.00\,/\,0.00} & \rev{0.26\,/\,0.24} & \rev{0.38\,/\,0.35} & \rev{0.52\,/\,0.51} & \rev{0.53\,/\,0.50} & \rev{0.53\,/\,0.50} & \rev{0.00\,/\,0.00} & \rev{0.72\,/\,0.68} & \rev{0.88\,/\,0.86} & \rev{0.98\,/\,0.97} & \rev{0.47\,/\,0.46} & \rev{\sci{9.33}{3}} \\
            \rev{GPT-5.2} & 
            \rev{\thirdS{0.48}\,/\,\thirdS{0.46}} & \rev{0.63\,/\,0.60} & \rev{\secondS{0.38}\,/\,\secondS{0.37}} & \rev{0.03\,/\,0.03} & \rev{0.58\,/\,\thirdS{0.57}} & \rev{0.04\,/\,0.04} & \rev{\secondS{0.66}\,/\,0.62} & \rev{\secondS{0.35}\,/\,\secondS{0.34}} & \rev{0.90\,/\,0.90} & \rev{\thirdS{0.89}\,/\,0.88} & \rev{0.99\,/\,0.99} & \rev{0.54\,/\,0.53} & \rev{\sci{7.22}{3}} \\
            \midrule
            \rowcolor{black!5}\multicolumn{14}{c}{\rev{\textbf{\textit{LLMs with self-consistency confidence}}}} \\
            \rev{DeepSeek-R1} & 
            \rev{0.00\,/\,0.00} & \rev{0.65\,/\,0.58} & \rev{0.00\,/\,0.00} & \rev{0.00\,/\,0.00} & \rev{0.55\,/\,0.50} & \rev{0.00\,/\,0.00} & \rev{0.53\,/\,0.48} & \rev{0.00\,/\,0.00} & \rev{0.94\,/\,0.93} & \rev{\thirdS{0.89}\,/\,0.86} & \rev{0.99\,/\,0.98} & \rev{0.41\,/\,0.39} & \rev{\sci{5.90}{2}} \\
            \rev{o4-mini} & 
            \rev{0.00\,/\,0.00} & \rev{\firstS{0.74}\,/\,\secondS{0.68}} & \rev{0.00\,/\,0.00} & \rev{0.00\,/\,0.00} & \rev{\secondS{0.59}\,/\,0.54} & \rev{0.57\,/\,0.51} & \rev{0.00\,/\,0.00} & \rev{0.00\,/\,0.00} & \rev{\firstS{1.00}\,/\,\firstS{1.00}} & \rev{\thirdS{0.89}\,/\,0.86} & \rev{\firstS{1.00}\,/\,\firstS{1.00}} & \rev{0.43\,/\,0.42} & \rev{\sci{1.23}{1}} \\
            \midrule
            \rowcolor{black!5}\multicolumn{14}{c}{\textbf{\textit{Ours}}} \\
            \textbf{\at-Qwen} & 
            \secondS{0.53}\,/\,\secondS{0.48} & 0.62\,/\,0.61 & 0.29\,/\,0.28 & \thirdS{0.50}\,/\,\thirdS{0.49} & \secondS{0.59}\,/\,\firstS{0.58} & \thirdS{0.59}\,/\,0.56 & \secondS{0.66}\,/\,\secondS{0.64} & \thirdS{0.28}\,/\,\thirdS{0.23} & \thirdS{0.99}\,/\,\thirdS{0.99} & \thirdS{0.89}\,/\,\secondS{0.89} & \firstS{1.00}\,/\,0.99 & \secondS{0.63}\,/\,\secondS{0.61} & \secondS{\sci{1.48}{3}} \\
            \textbf{\at-GPT} & 
            0.47\,/\,0.44 & \secondS{0.71}\,/\,\firstS{0.69} & \thirdS{0.33}\,/\,\thirdS{0.32} & \firstS{0.55}\,/\,\firstS{0.50} & \firstS{0.60}\,/\,\firstS{0.58} & \secondS{0.61}\,/\,\secondS{0.57} & \firstS{0.69}\,/\,\firstS{0.67} & \firstS{0.42}\,/\,\firstS{0.42} & \thirdS{0.99}\,/\,0.98 & \thirdS{0.89}\,/\,\secondS{0.89} & 0.99\,/\,0.99 & \firstS{0.66}\,/\,\firstS{0.64} & \sci{1.13}{2} \\
            \bottomrule
        \end{tabular}%
    }
\end{table*}

\subsection{Experimental Setup}
\label{sec:experiment-setup}

\myparagraph{Benchmarks}
To rigorously evaluate the performance of \at, we built a comprehensive set of 11 benchmarks from 11 diverse tabular data sources. We select 200 test tables per benchmark from each tabular source, for a comprehensive set of 2,200 test tables in total. 
We test both \textit{in-distribution} and \textit{out-of-distribution} settings. 

\underline{Training Data.} We train our specialist models using training data built with tables from six tabular sources: 
\begin{itemize}[leftmargin=*,noitemsep]
  \item \underline{\emph{\excelFiftyK}} is a large collection of 467K  relational tables parsed from spreadsheet files (.xlsx) crawled from a search engine index.
  \item \underline{\emph{\pbi}} is a collection of 12K relational tables extracted from business intelligence  (BI) models obtained from a prior study~\cite{lin2023auto}.
  \item \underline{\emph{\webwiki}} is a set of 292K Wikipedia tables extracted from a recent snapshot of Wikipedia.
    \item \underline{\emph{\gov}} is a collection of 1.6K CSV files crawled from a government portal (nationalarchives.gov.uk)~\cite{song2021auto}, following a similar crawling procedure as~\cite{gov-crawler}.
  \item 
  \underline{\emph{\gitcsvsmall}} is a recent crawl of 635 CSV files from GitHub. Compared to \gov, which is often data statistics manually created by government agencies, \gitcsvsmall are more developer-centric, with data programmatically generated by code on GitHub. 
  \item \underline{\emph{\parquet}} is a recent crawl of 29k Parquet files from GitHub, which is similar in nature to \gitcsvsmall, but in the Parquet format.
\end{itemize}
This collection spans public web tables, government statistics, and spreadsheet repositories, providing exposure to diverse table content and structures. We sample from this corpus to construct training cases using our masking procedure (Example~\ref{ex:masking}).

\underline{Test Data: In-Distribution (ID).}
For in-distribution tests, we sampled held-out test tables from five of our training sources: \emph{\excelFiftyK, \gov, \pbi, \parquet,} and \emph{\webwiki}\footnote{\gitcsvsmall has too few tables and could not be used for tests. 
}
These test tables are fully disjoint from those used during training.

\underline{Test Data: Out-of-Distribution (OOD).} In addition, we randomly sample tables from six additional tabular sources completely unseen during training, hereby introducing novel data content and previously unobserved domains, to test models' generalizability: 
\begin{itemize}[leftmargin=*,noitemsep]
  \item \underline{\emph{\enterprisecosmos}} is a collection of 100k enterprise CSV files obtained from a large enterprise's data lake behind a corporate firewall. Unlike \gov or \gitcsvsmall, most files are proprietary and unavailable on the public web, making it a strong ``OOD'' test set.
  \item \underline{\emph{\enterpriseexcel}} is a collection of 100k enterprise Spreadsheet files (.xlsx), obtained from a large enterprise. Similar to \enterprisecosmos, \enterpriseexcel are proprietary in nature and reserved for OOD evaluation.
  \item \underline{\emph{\webtable} \citep{xing2025mmtu}} is a large collection of 1k general web tables extracted from public HTML pages. 
  \item \underline{\emph{\ar}} is a collection of 6k real tables with known column-level arithmetic relationships (AR), e.g., column ``Margin'' = ``Profit'' / ``Cost'' in the same table~\cite{han2026autorelate}. When a cell in an arithmetic relationship is missing, the missing value can be reverse-engineered from the remaining values in the same row.
  \item \underline{\emph{\st}} includes 1.1k tables with known string-based relationships (e.g. column ``full-name'' = column ``first-name'' concatenate column ``last-name''~\cite{han2026autorelate}). Like \ar, missing values in a string relationship can be inferred from values in the same row.
  \item \underline{\emph{\fd}}\footnote{Note that in the case of \ar, \st, \fd, even if relationships are known, they are not provided as input, therefore requiring models to use their reasoning capabilities to infer implicit relationships before they can fill values in correctly.} is a set of 167k tables with real Functional Dependencies (FD) (e.g., $ProductID \to ProductName$)~\cite{han2026autorelate}. When a value in a dependent column is missing, the missing value can also be inferred using the FDs in the table.
\end{itemize}


\underline{Data Filtering and Preprocessing.} During pre-processing, we remove low-information tables, or ones with a missing-value ratio exceeding 70\%. We also filtered out cells exceeding 1,000 characters (which likely contain long natural-language texts), to focus on structured tabular content rather than long-form text.
From each ID and OOD source, we sample 200 test tables, each masking out exactly one cell  (marked as \texttt{[MISSING]}, as in Figure~\ref{fig:autofill-framework}). 

\myparagraph{Training Details} Our primary experiments use Qwen3-8B as the base architecture for all three specialists~\citep{yang2025qwen3, subramanian2025small, belcak2025small};
we refer to this configuration as \at-Qwen. We also train a GPT-4.1 mini variant, another SLM-class model~\citep{openai2025gpt41}, via the Microsoft Azure AI Foundry fine-tuning API\footnote{\url{https://learn.microsoft.com/en-us/azure/ai-services/openai/how-to/fine-tuning}.}, referred to as \at-GPT.  
We additionally train models of different sizes in the same family (Qwen3-8B, 4B, 1.7B; GPT-4.1-full, mini, nano) for comparisons. 

For \at-Qwen, \mk is trained via standard supervised fine-tuning on 30K direct question-answer pairs without reasoning, while \mr and \mc are each fine-tuned on 50K examples from the DeepSeek-R1 distillation. All three specialists are then calibrated on 2,000 held-out cases from the ID training corpus.
\hide{\at-Qwen models are fine-tuned for 2 epochs on 4 $\times$ A100 (80GB) GPUs, using a learning rate of 1e-5 for \mk and \mc, and 2e-5 for \mr. Training \mk requires approximately 11 hours, while \mr and \mc each require $\sim$40 hours due to long chain-of-thought traces.}

During inference, \mr and \mc use a sampling temperature of 0.8 to encourage diverse reasoning paths, while \mk uses 0.1, reflecting its role as a deterministic knowledge retriever.

\myparagraph{Evaluation Metrics}
As defined in our problem formulation (Section~\ref{sec:problem}), we target high-precision scenarios, where models must abstain whenever their calibrated confidence falls below a predefined precision threshold $\tau$ (e.g., $\tau = 0.9$). Accordingly, our evaluation focuses on the \emph{high-precision regime}.
Following~\cite{chen2025auto}, we use two metrics to assess result quality in this high-precision regime: 


\emph{R@P=0.9} (\emph{Recall@Precision=0.9}) measures recall at the target precision threshold\footnote{Similar to~\cite{chen2025auto}, predictions are sorted by descending calibrated confidence, and all predictions sharing the same confidence score are evaluated as a group. Recall is recorded at the highest-confidence prefix of the ranked list where precision, computed over all predictions from the top down to that point, remains at or above 0.9}. The reported recall thus reflects a high-precision segment from the top of the ranked list, without crediting recall at lower confidence levels, which directly aligns with our high-precision problem setting in Definition~\ref{def:problem}.
\iftoggle{full}
{
    Additional results for R@P=0.8 and full accuracy can be found in Appendix~\ref{appendix:additional_results}.
}
{
    Additional results for R@P=0.8 and full accuracy can be found in our technical report~\citep{full}.
}

\emph{pAUPRC} (partial Area Under the Precision-Recall Curve) measures the area under the curve for precision $\ge 0.9$. While R@P=0.9 identifies the maximum coverage before precision drops below the threshold, pAUPRC evaluates the stability of the model's accuracy leading up to that point, rewarding models that maintain near-perfect precision across their most confident predictions. 




\begin{figure*}[!t]
\centering
    \includegraphics[width=.95\textwidth]{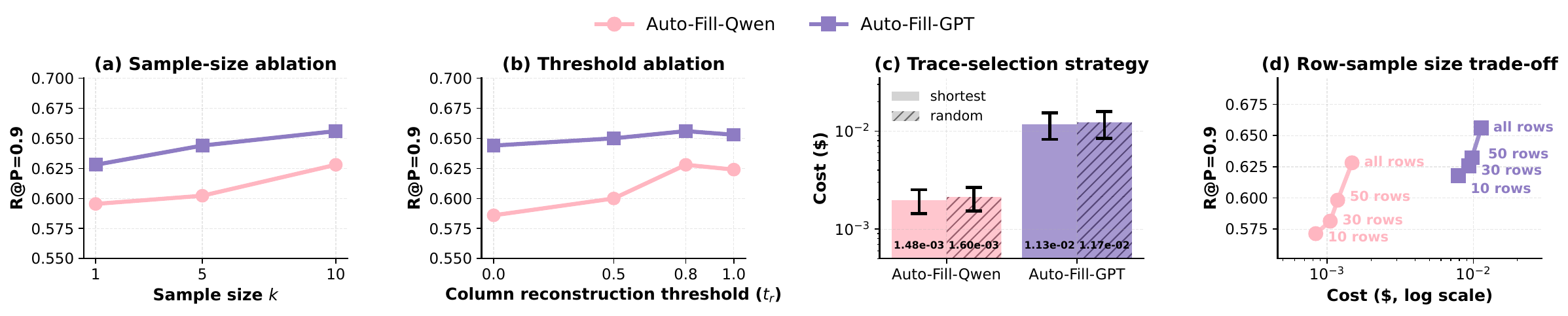}
\caption{%
  \rev{Sensitivity of \at to trace-construction and inference-time choices: (a)~\mr's teacher sample size $k$, (b)~\mc's column-reconstruction threshold $t_r$, (c)~shortest vs.\ random trace selection, and (d)~input-table row sampling at inference.}
}
\label{fig:sensitivity}
\end{figure*}

\myparagraph{Baselines}
We compare \at with a diverse set of methods: 

\underline{Vanilla Language Models.} We evaluate a broad set of open- and closed-source models at various scales. Closed-source baselines include GPT-4.1 and GPT-4.1 mini, frontier OpenAI models (o4-mini, o3-pro, GPT-5.2), as well as Gemini 3 Pro. Open-source baselines include Qwen3-8B (with thinking) and the frontier open-source reasoning model DeepSeek-R1. \rev{By default, we prompt all vanilla LLMs to \emph{verbalize} a confidence score alongside the prediction. We also evaluate \emph{logprob} (next-token probability of the predicted answer) for non-reasoning models, and \emph{self-consistency}~\citep{wang2022self} ($k{=}10$ sample agreement) for the more affordable reasoning models (DeepSeek-R1, o4-mini); we exclude premium reasoning models (e.g., o3-pro) here as $k$-sample inference becomes prohibitively expensive. Logprob is inapplicable to reasoning models, since chain-of-thought breaks token-level calibration~\citep{nakkiran2025trained} and closed-source reasoning APIs do not expose logprobs.}  

\underline{Retrieval-Augmented Imputation.} We compare against LakeFill~\citep{yang2025lakefill}, a state-of-the-art retrieval-based imputation system for data lakes. LakeFill retrieves related tuples from a data lake and leverages an LLM to perform imputation. We use the authors’ original implementation with their confidence-aware variant for abstention. \textbf{LakeFill-full} runs on a lake of all 50K training tables used by \at; since retrieval often fails at this scale, we also test \textbf{LakeFill-small}, which uses only the 200 test tables per benchmark.

\rev{\underline{Tabular ML models.} We compare against \textbf{TabPFN}~\citep{hollmann2022tabpfn}, a pretrained transformer for in-context tabular prediction. We adapt it by treating filled rows as in-context training examples and predicting the masked cell in a query row, using the multiclass extension for high-cardinality columns and predicted probability as confidence.}

\rev{\underline{Data Repair Approaches.} We adapt two repair systems to single-cell imputation: \textbf{SCARE}~\citep{yakout2013don}, with one masked cell per row, reduces to a Naive-Bayes predictor over the non-target columns (class probability as confidence); and \textbf{Baran}~\citep{mahdavi2020baran}, where, lacking user-labeled corrections, we hold out observed target-column cells as simulated corrections and use its meta-classifier score as confidence.}
%
%
\rev{
\iftoggle{full}
{
Variants of these baselines are in Appendix~\ref{appendix:baselines}.
    }
{
    }
}


\begin{table}[!t]
    \centering
    \caption{Quality and cost trade-offs across model sizes. Metrics are averaged over ID and OOD datasets.}
    \label{tab:model_size}
    \resizebox{\columnwidth}{!}{%
    \begin{tabular}{@{}cccccccc@{}}
        \toprule
        \multirow{2}{*}{\textbf{Model}} & \multirow{2}{*}{\textbf{Size}} & \multicolumn{2}{c}{\textbf{ID}} & \multicolumn{2}{c}{\textbf{OOD}} & \multirow{2}{*}{\textbf{Cost (\$)}} \\
        \cmidrule(lr){3-4} \cmidrule(lr){5-6}
         & & \textbf{R@P=0.9} & \textbf{pAUPRC} & \textbf{R@P=0.9} & \textbf{pAUPRC} & \\ \midrule
        \multirow{3}{*}{\textbf{Qwen3}}   & 1.7B & 0.362 & 0.352 & 0.668 & 0.663 & --- \\
         & 4B   & 0.423 & 0.411 & 0.715 & 0.710 & \first{\sci{3.27}{4}} \\
         & 8B   & \first{0.503} & \first{0.488} & \first{0.733} & \first{0.717} & \sci{1.48}{3} \\ \midrule
        \multirow{3}{*}{\textbf{GPT-4.1}} & Nano & 0.279 & 0.240 & 0.609 & 0.588 & \first{\sci{1.85}{3}} \\
         & Mini & 0.530 & 0.508 & \first{0.762} & \first{0.752} & \sci{1.13}{2} \\
         & Full & \first{0.569} & \first{0.555} & 0.757 & 0.750 & \sci{5.24}{2} \\ \bottomrule
    \end{tabular}
    }
\end{table}

\iftoggle{full}
{
    \begin{table}[t]
\centering
\caption{Effect of RL fine-tuning on specialist models in \at-\texttt{Qwen3-1.7B}. 
The baseline (---) uses SFT-only specialists.}
\label{tab:grpo_ablation}
\resizebox{\columnwidth}{!}{%
\begin{tabular}{l cc cc}
\toprule
\multirow{2}{*}{\makecell[l]{\textbf{Specialist} \\ \textbf{w/ RL}}} & \multicolumn{2}{c}{\textbf{ID}} & \multicolumn{2}{c}{\textbf{OOD}} \\
\cmidrule(lr){2-3} \cmidrule(lr){4-5}
& \textbf{R@P=0.9} & \textbf{pAUPRC} & \textbf{R@P=0.9} & \textbf{pAUPRC} \\
\midrule
$\mathcal{M}_R$
    & 0.354 {\scriptsize($\downarrow$0.008)} & 0.345 {\scriptsize($\downarrow$0.007)}
    & 0.673 {\scriptsize($\uparrow$0.005)} & 0.667 {\scriptsize($\uparrow$0.004)} \\
$\mathcal{M}_C$
    & 0.377 {\scriptsize($\uparrow$0.015)} & 0.366 {\scriptsize($\uparrow$0.014)}
    & 0.668 {\scriptsize($\uparrow$0.000)} & 0.662 {\scriptsize($\downarrow$0.001)} \\
$\mathcal{M}_R + \mathcal{M}_C$
    & 0.369 {\scriptsize($\uparrow$0.007)} & 0.359 {\scriptsize($\uparrow$0.007)}
    & 0.678 {\scriptsize($\uparrow$0.010)} & 0.671 {\scriptsize($\uparrow$0.008)} \\
\bottomrule
\end{tabular}
}
\end{table}

}
{

}

\underline{Functional Dependency (FD)-based Methods.} FD-based methods exploit deterministic column dependencies (e.g., employee-ID $\Rightarrow$ employee-name) widely used in data cleaning. Since FDs yield no calibrated confidence, we adopt an optimistic \textbf{FD-upper-bound}: if a missing cell lies on the RHS of an FD whose LHS appears elsewhere, we count it as ``solved by FD'' without imposing the $\geq 0.9$ precision threshold required of other methods.


\underline{\at Variants.} We also compare against alternative design choices from Table~\ref{tab:architecture-options}, including a hybrid model trained on mixed data, and a learned router or a classical classifier for ensemble selections. We report these results in our ablation study (Section~\ref{sec:ablation}).


\subsection{Overall comparisons}

\underline{Quality comparisons.}
Table~\ref{tab:abstain_combined} reports R@P=0.9 and pAUPRC across all 11 datasets. \at-Qwen achieves 0.628 and 0.613, and \at-GPT further improves to 0.656 and 0.641, both outperforming \emph{all} baselines on both metrics.

The FD upper-bound row 
reports the maximum recall achievable under perfect abstention. Even so, recall stays low, showing that formal constraints cannot capture the knowledge or reasoning needed to fill missing values.
%
\rev{
Both LakeFill variants score low, often because the target is absent from the retrieved tuples (or the lake itself) and their confidence is unreliable; LakeFill-small does slightly better, suggesting a larger lake adds more noise than signal. Likewise, TabPFN and classical repair methods (Baran, SCARE) rely on statistical association across rows—succeeding on \fd and enterprise tables where targets recur as lookup values, but failing on \ar\,/\,\st\ (computed, unique per row) and on knowledge-intensive benchmarks whose answers lie outside the table.}

\begin{table}[t]
    \centering
    \caption{Ablation study on specialist combinations, with R@P denoting R@P=0.9.}
    \label{tab:abstain_90_specialist_ablation_avg}
    \resizebox{\columnwidth}{!}{%
        \begin{tabular}{l | cc cc | cc cc}
            \toprule
            & \multicolumn{4}{c|}{\textbf{Qwen3-8B}} & \multicolumn{4}{c}{\textbf{GPT-4.1 mini}} \\
            \cmidrule(lr){2-5} \cmidrule(lr){6-9}
            \textbf{Specialist(s)} & \multicolumn{2}{c}{\textbf{ID}} & \multicolumn{2}{c|}{\textbf{OOD}} & \multicolumn{2}{c}{\textbf{ID}} & \multicolumn{2}{c}{\textbf{OOD}} \\
            \cmidrule(lr){2-3} \cmidrule(lr){4-5} \cmidrule(lr){6-7} \cmidrule(lr){8-9}
            & \textbf{R@P} & \textbf{pAUPRC} & \textbf{R@P} & \textbf{pAUPRC} & \textbf{R@P} & \textbf{pAUPRC} & \textbf{R@P} & \textbf{pAUPRC} \\
            \midrule
            \mk only         & 0.428 & 0.421 & 0.637 & 0.629 & 0.501 & 0.488 & 0.705 & 0.697 \\
            \mr only         & 0.258 & 0.246 & 0.551 & 0.538 & 0.338 & 0.325 & 0.547 & 0.525 \\
            \mc only         & 0.160 & 0.151 & 0.497 & 0.491 & 0.187 & 0.179 & 0.430 & 0.430 \\
            \cmidrule{1-9}
            \mk + \mr        & 0.466 & 0.451 & 0.692 & 0.678 & 0.512 & 0.499 & 0.729 & 0.713 \\
            \mk + \mc        & 0.467 & 0.452 & 0.718 & 0.707 & 0.526 & 0.449 & 0.748 & 0.697 \\
            \mr + \mc        & 0.371 & 0.356 & 0.598 & 0.591 & 0.347 & 0.330 & 0.555 & 0.550 \\
            \cmidrule{1-9}
            \textbf{All (Ours)} & \first{0.503} & \first{0.488} & \first{0.733} & \first{0.717} & \first{0.530} & \first{0.508} & \first{0.762} & \first{0.752} \\
            \bottomrule
        \end{tabular}%
    }
\end{table}

Vanilla LLMs are known to be overconfident and to hallucinate under uncertainty, and this is on full display on our benchmarks. While stronger frontier models perform well when answers are certain and directly derivable from the table context -- e.g., o3-pro and Gemini 3 Pro both reach near-perfect quality on relational datasets (\ar, \fd, \st), they all drop sharply on other less deterministic benchmarks that require abstention (overconfident errors can quickly push precision below 0.9 and collapse R@P=0.9). 
Even DeepSeek-R1, our distillation teacher \T, achieves only modest mean scores, confirming that raw reasoning capability alone does not yield reliable uncertainty estimation. Smaller models such as Qwen3-8B and GPT-4.1 mini collapse to near-zero recall on almost all benchmarks. \rev{Replacing verbalized confidence with logprob or self-consistency yields mixed results. Logprob helps weaker non-reasoning models like GPT-4.1 mini substantially, but only marginally improves GPT-5.2. Self-consistency helps DeepSeek-R1 yet hurts o4-mini, while inflating cost $\sim10\times$. As no alternative applies uniformly across all backbones, and none matches \at, we adopt \emph{verbalized confidence} as the default in all other results.}


In comparison, \at addresses this using each specialist's trained confidence (tailored to its mode), and calibration for true probabilities. 
On relational datasets, both \at variants achieve near-perfect recall via \mc's execution-based self-validation. On the noisier and more challenging \webtable, \at-GPT reaches 0.42 versus 0.00 for most baselines, benefiting from GPT-4.1's richer world knowledge and strong abstention.
\iftoggle{full}
{
    PR curves for all methods are provided in Appendix~\ref{appendix:pr-curve}
}
{
    PR curves for all methods are provided in the full version~\citep{full}.
}

\underline{Cost comparisons.} The last column of Table~\ref{tab:abstain_combined} reports per-query LLM inference cost, computed using the lowest publicly available API prices.\footnote{Prices sourced from \href{https://openrouter.ai/}{OpenRouter} for DeepSeek-R1 and Gemini 3 Pro, \href{https://azure.microsoft.com/en-us/pricing/details/azure-openai/}{Azure OpenAI} for GPT family models, and \href{https://www.siliconflow.com/pricing}{SiliconFlow} for Qwen3 models. (accessed February 2026)}
\iftoggle{full}
{
    Per-dataset costs can be found in Appendix~\ref{appendix:cost-analysis}.
}
{
    Per-dataset costs can be found in our full version~\citep{full}.
}

\at-Qwen is over $100\times$ cheaper than o3-pro and $29\times$ cheaper than Gemini 3 Pro while achieving higher mean quality than both. \at-GPT remains cost-competitive with o4-mini while substantially outperforming it in quality. 

Figure~\ref{fig:recall_vs_cost} summarizes the joint comparison: both \at variants form the quality--cost Pareto frontier and achieve substantial quality gains at less than 1\% of the cost of frontier models. 


\begin{table}[t]
\centering
\caption{Ablation study on ensemble strategies in \at-\texttt{Qwen}. \#$\mathcal{M}$ denotes the number of models used. \rev{Parentheses show drops from the full calibrated ensemble.}
}
\label{tab:composition_ablation}
\resizebox{\columnwidth}{!}{%
\begin{tabular}{l c cc cc c}
\toprule
\multirow{2}{*}{\textbf{Variant}} & \multirow{2}{*}{\textbf{\#$\mathcal{M}$}} & \multicolumn{2}{c}{\textbf{ID ($\uparrow$)}} & \multicolumn{2}{c}{\textbf{OOD ($\uparrow$)}} & \multirow{2}{*}{\textbf{Cost (\$) ($\downarrow$)}} \\
\cmidrule(lr){3-4} \cmidrule(lr){5-6}
 & & \textbf{R@P=0.9} & \textbf{pAUPRC} & \textbf{R@P=0.9} & \textbf{pAUPRC} & \\
\midrule
Hybrid Model  & 1 & 0.000 {\scriptsize($\downarrow$0.503)} & 0.000 {\scriptsize($\downarrow$0.488)} & 0.358 {\scriptsize($\downarrow$0.375)} & 0.357 {\scriptsize($\downarrow$0.360)} & \sci{5.45}{5} \\
Learned Router  & 2 & 0.158 {\scriptsize($\downarrow$0.345)} & 0.153 {\scriptsize($\downarrow$0.335)} & 0.423 {\scriptsize($\downarrow$0.310)} & 0.413 {\scriptsize($\downarrow$0.304)} & \sci{4.58}{4} \\
Classical ML  & 3 & 0.178 {\scriptsize($\downarrow$0.325)} & 0.156 {\scriptsize($\downarrow$0.332)} & 0.570 {\scriptsize($\downarrow$0.163)} & 0.565 {\scriptsize($\downarrow$0.152)} & \sci{1.48}{3} \\
\bottomrule
\end{tabular}
}
\end{table}

\subsection{Sensitivity analysis}
\label{sec:sensitivity}

\underline{Sensitivity to base model sizes.}
Table~\ref{tab:model_size} reports performance across Qwen3 (1.7B, 4B, 8B) and GPT-4.1 (Nano, Mini, Full) variants.\footnote{GPT-4.1 Mini and Nano are SLM-class models per OpenAI~\citep{openai2025gpt41}, while GPT-4.1 Full is included to assess the ceiling when using a frontier backbone.} 
%
Within each family, quality generally improves with model size. For Qwen3, performance improves consistently with size, though with diminishing OOD returns, suggesting that moderate model sizes are likely sufficient. Similarly, for GPT-4.1, Mini achieves comparable OOD performance to Full. 
This result shows that \at is robust to model architectures and sizes, allowing practitioners to select a backbone that fits their cost profiles using our framework. 
\rev{\underline{Sensitivity to Trace Construction Choices.}} \rev{Figure~\ref{fig:sensitivity}(a-c) varies three choices in our distillation pipeline. For \mr (Fig.~\ref{fig:sensitivity}a), quality improves with the number of teacher samples $k$ used for confidence aggregation. For \mc (Fig.~\ref{fig:sensitivity}b), the column-reconstruction threshold peaks at $t_r=0.8$: $t_r=0.0$ admits hard-coded snippets that fit only the masked row, while $t_r=1.0$ rejects tables with legitimate outliers such as sub-totals. Finally, selecting the shortest trace achieves comparable quality to random selection at slightly lower inference cost because it produces shorter student outputs (Fig.~\ref{fig:sensitivity}c).}

\rev{\underline{Sensitivity to Input-Table Row Sampling.}} \rev{Real-world tables can exceed an SLM's context window. To probe \at at this scale, Figure~\ref{fig:sensitivity}d sub-samples the input from 10, 30, 50, or all rows. Going from all rows down to 10 reduces R@P=0.9 only mildly ($\sim$0.05 for both backbones) while cutting input-token cost, indicating that uniform row sampling is a practical strategy for large tables.}


\begin{table}[t]
\caption{Ablation study on confidence extraction mechanisms. Each row replaces one 
specialist's confidence signal with random verbal confidence, keeping other 
components fixed.
}
\label{tab:confidence-ablation}
\centering
\resizebox{\columnwidth}{!}{%
\begin{tabular}{c l cc cc}
\toprule
\multirow{2}{*}{\textbf{Specialist}} & \multirow{2}{*}{\textbf{Ablation}} & \multicolumn{2}{c}{\textbf{ID}} & \multicolumn{2}{c}{\textbf{OOD}} \\
\cmidrule(lr){3-4} \cmidrule(lr){5-6}
 & & \textbf{R@P=0.9} & \textbf{pAUPRC} & \textbf{R@P=0.9} & \textbf{pAUPRC} \\
\midrule
$\mathcal{M}_K$ & Logprob
    & 0.290 {\scriptsize($\downarrow$0.213)} & 0.274 {\scriptsize($\downarrow$0.214)}
    & 0.625 {\scriptsize($\downarrow$0.108)} & 0.616 {\scriptsize($\downarrow$0.101)} \\
$\mathcal{M}_R$ & Avg.\ verbal
    & 0.467 {\scriptsize($\downarrow$0.036)} & 0.458 {\scriptsize($\downarrow$0.030)}
    & 0.715 {\scriptsize($\downarrow$0.018)} & 0.705 {\scriptsize($\downarrow$0.012)} \\
$\mathcal{M}_C$ & Execution
    & 0.466 {\scriptsize($\downarrow$0.037)} & 0.451 {\scriptsize($\downarrow$0.037)}
    & 0.706 {\scriptsize($\downarrow$0.027)} & 0.691 {\scriptsize($\downarrow$0.026)} \\
\bottomrule
\end{tabular}
}
\end{table}

\iftoggle{full}
{\underline{Sensitivity to RL fine-tuning}
Table~\ref{tab:grpo_ablation} reports the effect of adding RL on top of distillation-based SFT, with Qwen3-1.7B as the base. 
Unlike standard tasks where RL optimizes a single accuracy signal, our setting requires jointly improving prediction accuracy \emph{and} confidence calibration -- a harder target where gains on one can come at the expense of the other~\citep{leng2024taming,stangel2025rewarding}.
This challenge is reflected in the results: GRPO yields only modest gains,  and combining both \mr and \mc produces the most consistent improvement, with gains across all four metrics. Given the marginal overall gains and additional training cost, we treat GRPO as optional in \at, which is a potential area for future research.}
{

}

\subsection{Ablation studies}
\label{sec:ablation}

\underline{Contributions of individual SLMs.}
Table~\ref{tab:abstain_90_specialist_ablation_avg} ablates the contribution of each specialist.
Among single specialists, \mk achieves the strongest performance. 
\mc achieves the lowest overall recall, as programmatic column patterns exist in only a subset of tables. 
Among pairwise combinations, \mk+\mc performs best overall,
suggesting that knowledge and coding are especially complementary. 
Combining all three specialists achieves the best performance across both settings, confirming the benefit of the full design.
\iftoggle{full}
{
    Per-dataset breakdowns revealing further insights, such as \mr's collapse on knowledge-intensive datasets and \mc's near-perfect recall on relational ones, are discussed in Appendix~\ref{appendix:specialist-ablation}.
}
{
    Further per-dataset analysis is provided in our full version~\citep{full}.
}

\underline{Contributions of specialist composition strategy.}
Table~\ref{tab:composition_ablation} compares alternative composition strategies. The single hybrid model collapses to zero ID recall, likely because conflicting training signals prevent it from learning reliable confidence~\citep{yu2020gradient,shen2024mome}. The learned router also performs worse because it predicts which specialist to invoke rather than producing a calibrated probability. The classical ML ensemble improves over the router on OOD data but still trails calibration, suggesting that its learned confidence mapping does not generalize reliably. Overall, calibrated confidence selection provides the strongest and most consistent performance.

\underline{Contributions of confidence extraction.}
Table~\ref{tab:confidence-ablation} evaluates each specialist's confidence mechanism by replacing it with verbalized confidence at inference time, and with \T's verbalized confidence during training.
\mk's substitution incurs the largest drop, confirming the value of log-probabilities in the direct-answer setting. \mr drops moderately, as single-trace verbal confidence is noisier than $k$-run aggregation, and \mc drops comparably, since verbal confidence removes the execution-grounded correctness signal.

\section{Conclusions and Future Work}

We study the problem of predicting missing values in tables with calibrated precision estimates, and develop \at that post-trains specialist SLMs for knowledge/reasoning/coding, respectively, which are then combined using a dynamic calibrated ensemble that can abstain when no specialist SLM is confident. Extensive experiments show that \at achieves state-of-the-art accuracy, while operating at less than 1\% of the cost of frontier models. 
%
\rev{Future directions include jointly predicting interdependent missing cells 
and context pruning for large tables.}

\begin{acks}
    We sincerely thank Dr. Juliana Freire of New York University for her thoughtful feedback, as well as Tal Kariv, Tsofiya Aiello, Gil Kulish, Arnon Peretz, Dorli Hanson, Danielle Rifinski Fainman, Shir Zehavi Zoran, and many others on the Excel Clean Data team for their invaluable support.
\end{acks}

\iftoggle{full}
{
}
{
\clearpage
}

\balance

\bibliographystyle{ACM-Reference-Format}
\bibliography{Auto-Fill}
\clearpage

\iftoggle{full}
{
    \appendix 
    \section{Technical Details}

\subsection{Benchmark Construction Details}
\label{appendix:benchmark-construction}
For each dataset, we construct 200 distinct evaluation cases where exactly one cell is masked with the token \texttt{[MISSING]}. We employ two selection strategies based on dataset characteristics. (1) For datasets testing broad table comprehension or external knowledge (e.g., the public web-based and private-lake series), we uniformly sample non-null, non-empty cells as masking targets. This simulates realistic missing data scenarios where models must rely on surrounding context without structural information.
(2) For datasets derived from relationship detection tasks, we apply domain-specific selection criteria to generate evaluation cases that target cells participating in semantic relationships.
\begin{itemize}[leftmargin=*,noitemsep]
  \item \emph{Arithmetic Reasoning (\ar):} Tables in this dataset contain columns related through algebraic formulas.
  We strategically mask cells that participate in these algebraic relationships, testing whether models can perform quantitative reasoning by inferring missing values from learned arithmetic patterns.
  \item \emph{Functional Dependencies (\fd):} Tables contain columns with functional dependencies, where one column's value uniquely determines another's. 
  We mask cells in dependent columns, requiring models to exploit these deterministic relationships.
  \item \emph{Semantic Transformations (\st):} Tables contain columns derived through transformations of other columns, like concatenation, format conversion, etc. We mask cells in derived columns whose values can be reconstructed from source columns within the same table.
\end{itemize}

\subsection{Additional Fine-Tuning Details}
\at-Qwen models are fine-tuned for 2 epochs on 4 $\times$ A100 (80GB) GPUs, using a learning rate of 1e-5 for \mk and \mc, and 2e-5 for \mr. Training \mk requires $\sim$11 hours, while \mr and \mc each require $\sim$40 hours due to long chain-of-thought traces.

\subsection{RL Training Details}
\label{appendix:rl-training}
We apply GRPO-based RL fine-tuning~\cite{shao2024deepseekmath} on top of the distillation-based SFT checkpoints of \mr and \mc, using the ms-swift framework~\cite{zhao2024swiftascalablelightweightinfrastructure}. RL fine-tuning is not applied to \mk, as its confidence signal derives from token-level log-probabilities rather than a verifiable output suitable for reward-based training.

Both \mr and \mc are fine-tuned on 5{,}000 examples sampled from the same masking corpus used for SFT. Training runs for one epoch on 4$\times$A100 (80 GB) GPUs with a per-device batch size of 1 and gradient accumulation over 16 steps, yielding an effective batch size of 32. We use a learning rate of $1\times10^{-6}$ with cosine decay and a warmup ratio of 0.01. For GRPO, we sample $G=4$ completions per prompt and set the KL penalty coefficient to $\beta=0.01$. Rollout generation uses a sampling temperature of 0.8, consistent with SFT inference settings. The maximum completion length is 8{,}192 tokens, and the maximum total context length is 24{,}576 tokens; generation is performed using vLLM in colocated mode.

\underline{Reward function.} We use the RLCR reward described in Section~\ref{sec:grpo} (Eq.~\ref{eq:rlcr}). For \mr, $q$ is the verbalized integer confidence (0--100, normalized to $[0,1]$) parsed from the model's JSON output; when the confidence field is absent or unparseable, we default to $q=0.5$, the maximally uncertain prior, which penalizes the model symmetrically regardless of correctness and thereby incentivizes it to always emit an explicit confidence value. For \mc, $q$ is the execution-based column accuracy $\text{conf}_C$ (Eq.~\ref{eq:code}), computed by executing the generated code against the input table and measuring reproduction accuracy over observed column values. Completions that fail to produce a parseable prediction, or that result in a code execution error, receive a reward of $0$.

\subsection{\texorpdfstring{\rev{Baseline Details}}{Baseline Details}}
\label{appendix:baselines}
\rev{This section details our adaptation of one tabular ML baseline (\textbf{TabPFN}~\citep{hollmann2022tabpfn}) and two classical data-repair systems (\textbf{SCARE}~\citep{yakout2013don} and \textbf{Baran}~\citep{mahdavi2020baran}) to the single-cell imputation setting used in our experiments.}

\rev{\underline{TabPFN.}}
\rev{TabPFN~\citep{hollmann2022tabpfn} is a pretrained transformer that performs in-context tabular prediction. For a case with a \texttt{[MISSING]} cell at position $(i, j)$, we treat the rows where column $j$ is filled as the in-context training set and the row with the missing cell as the test instance. The target column is predicted with a classifier by default, and with a regressor when the column is numeric and most of its values are distinct, so that unseen numeric targets remain reachable. Because TabPFN-v2 supports at most $10$ output classes, target columns with more than $10$ distinct values require an additional adaptation, for which we evaluate two variants. \textbf{TabPFN-10class} restricts the prediction head to the $10$ most frequent training values in the target column, dropping any candidate outside this shortlist. \textbf{TabPFN-MultiClass} instead uses the \texttt{ManyClassClassifier} output-coding wrapper from \texttt{tabpfn-extensions}, which decomposes the $N$-way problem into $\lceil\log_{10}(N)\rceil$ sub-calls of arity $10$ and aggregates their outputs, keeping all training values reachable. Confidence is the top class probability for the classifier (or the aggregated wrapper probability under TabPFN-MultiClass); for the regressor branch, it is $\exp\bigl(-(q_{90} - q_{10}) / \sigma_y\bigr)$, where $\sigma_y$ is the training-target standard deviation and $q_{10}, q_{90}$ are the predicted decile bounds, so that a narrower predictive interval yields higher confidence. As shown in Table~\ref{tab:baseline-variants}, TabPFN-MultiClass slightly improves on TabPFN-10class on ID and outperforms on OOD, so we report TabPFN-MultiClass as \emph{TabPFN} in the main paper.}

\begin{table}[t!]
\centering
\small
\caption{\rev{R@P=0.9 of all classical-baseline variants, averaged across the 5 ID datasets and the 6 OOD datasets. The variant reported in the main paper (Table~\ref{tab:abstain_combined}) is marked with $^\star$.}}
\label{tab:baseline-variants}
\setlength{\tabcolsep}{6pt}
\rev{
\begin{tabular}{l|cc}
\toprule
\textbf{Method} & \textbf{ID} & \textbf{OOD} \\
\midrule
\rowcolor{black!5}\multicolumn{3}{c}{\textit{TabPFN variants}} \\
TabPFN-10class            & 0.104 & 0.250 \\
TabPFN-MultiClass$^\star$ & \textbf{0.106} & \textbf{0.260} \\
\midrule
\rowcolor{black!5}\multicolumn{3}{c}{\textit{SCARE variants}} \\
SCARE-Single$^\star$ & \textbf{0.093} & \textbf{0.270} \\
SCARE-Partition      & 0.089 & 0.260 \\
\midrule
\rowcolor{black!5}\multicolumn{3}{c}{\textit{Baran variants}} \\
Baran-A             & 0.115 & 0.290 \\
Baran-B-20$^\star$  & \textbf{0.140} & \textbf{0.310} \\
Baran-B-50          & 0.119 & 0.300 \\
Baran-B-100         & 0.115 & 0.290 \\
\bottomrule
\end{tabular}
}
\end{table}

\rev{\underline{SCARE.}}
\rev{SCARE~\citep{yakout2013don} repairs erroneous cells by chaining multinomial Naive-Bayes models that condition each ``flexible'' attribute on a small set of ``reliable'' attributes, partitioning the table by reliable-attribute values, fitting a per-partition model, and resolving multi-cell repairs with a hitting-set heuristic over a value-vote graph. Under our setting of one masked cell per case, the error-detection step and the graph pruning become vacuous, and the chain collapses to a single predictor. Our main adaptation, \textbf{SCARE-Single}, is therefore a single global multinomial Naive-Bayes classifier over the one-hot-encoded non-target columns, fit on rows where the target is filled and predicting over the full column domain; for purely numeric high-cardinality targets we substitute a Bayesian-ridge regressor on the same features. Confidence is the predicted class probability, or $\exp\bigl(-(q_{90} - q_{10}) / \sigma_y\bigr)$ in the regressor branch, with $q_{10}, q_{90}$ approximated from the regressor's posterior standard deviation under a Gaussian assumption. We additionally evaluate \textbf{SCARE-Partition}, which more faithfully preserves SCARE's partitioning: it selects the two lowest-cardinality non-target columns as reliable attributes, trains a per-partition Naive-Bayes model for each reliable-attribute value, and aggregates predictions across partitions weighted by partition size. As shown in Table~\ref{tab:baseline-variants}, SCARE-Partition does not outperform SCARE-Single---partitioning shrinks each per-partition training set and yields noisier estimates---so we report SCARE-Single as \emph{SCARE} in the main paper.}

\rev{\underline{Baran.}}
\rev{Baran~\citep{mahdavi2020baran} generates correction candidates from three corrector families---value-based, vicinity-based, and domain-based---and trains a per-column AdaBoost meta-classifier on user-labeled corrections to rank them. Two adaptations are required for our setting. First, we omit the value-based corrector, whose character-level edit transformations are learned from real $\langle\text{old}, \text{new}\rangle$ typo pairs and are undefined for the \texttt{[MISSING]} placeholder. Second, since our benchmark provides no user labels, we simulate the labeling budget by holding out $K$ filled cells from the target column as $\langle\text{missing}, \text{ground truth}\rangle$ corrections. We evaluate four label regimes. \textbf{Baran-A} uses no labels and takes the argmax over summed channel posteriors with the AdaBoost meta-classifier omitted, matching the original paper's $F_1=0$ at zero labels. \textbf{Baran-B-$K$} for $K \in \{20, 50, 100\}$ trains an AdaBoost meta-classifier ($100$ estimators) on the resulting candidate-score vectors and applies it to the masked cell, with the positive-class probability as confidence; $K=20$ matches the original paper's default labeling budget. As shown in Table~\ref{tab:baseline-variants}, Baran-B-20 is the strongest variant on both ID and OOD splits, with larger budgets ($K=50, 100$) giving no additional gain and Baran-A degenerating as expected. We therefore report Baran-B-20 as \emph{Baran} in the main paper.}
\section{Additional Results}
\label{appendix:additional_results}
This section provides supplementary results that extend the analysis in the main paper.

\subsection{Additional Sensitivity Analysis}
\label{appendix:sensitivity}

\rev{\underline{Sensitivity to Numerical-Matching Tolerance.}} \rev{Our results so far use strict exact-match. To verify our gains are not sensitive to this, Figure~\ref{fig:relative-error} re-computes R@P=0.9 under a relative-error tolerance $\epsilon \in \{0, 0.001, 0.01, 0.05\}$, where $\epsilon=0$ recovers exact-match\footnote{\rev{A numeric prediction counts as correct iff $|\text{pred} - \text{gt}| / |\text{gt}| \le \epsilon$; a non-numeric one iff its normalized Levenshtein distance to the ground truth is $\le \epsilon$.}}. Both \at variants stay almost flat across $\epsilon$, indicating their accepted predictions are already essentially exact, whereas baselines like o3-pro, GPT-5.2, and Qwen3-8B improve markedly under looser tolerance, suggesting frequent near-misses. Even at the loosest $\epsilon=0.05$, \at-GPT remains the strongest method.}

\subsection{Confidence Calibration}
\label{appendix:calibration}

We compare isotonic regression against Platt scaling~\citep{platt1999probabilistic}, beta 
calibration~\citep{kull2017beta}, and an uncalibrated baseline.
Besides overall task performance using R@P=0.9 and pAUPRC, we also assess calibration quality 
using \emph{Expected Calibration Error (ECE)}, which measures the gap between calibrated probability 
and empirical accuracy~\citep{naeini2015obtaining}.

\underline{Baselines.}
\begin{itemize}[leftmargin=*,noitemsep]
    \item \textit{Isotonic regression}, our default method, fits a non-parametric monotonic mapping between raw confidence and empirical accuracy as described in Section~\ref{sec:calibration}. 
    \item  \textit{Platt scaling}~\citep{platt1999probabilistic} fits a logistic regression model $\sigma(a \cdot \text{conf} + b)$ to map raw confidence to calibrated probability, assuming a sigmoid relationship. 
    \item  \textit{Beta calibration}~\citep{kull2017beta} extends Platt scaling by applying logistic regression in the log-odds space: $\text{logit}(p) = a \cdot \log(c) + b \cdot \log(1-c)$, where $c$ is the raw confidence, providing additional flexibility for probability-like inputs.
    \item \textit{None} applies no calibration, using raw confidence scores directly (with \mr's verbalized confidence normalized to $[0,1]$).
\end{itemize}

\begin{table}[!t]
\centering
\caption{Ablation study on confidence calibration methods.
}
\label{tab:calibration_ablation}
\resizebox{\columnwidth}{!}{%
\begin{tabular}{l|cc|cc|ccc|c}
\toprule
\multirow{2}{*}[-0.5em]{\textbf{\begin{tabular}{@{}c@{}}Calibration\\Method\end{tabular}}}
  & \multicolumn{2}{c|}{\textbf{ID ($\uparrow$)}}
  & \multicolumn{2}{c|}{\textbf{OOD ($\uparrow$)}}
  & \multicolumn{4}{c}{\textbf{ECE ($\downarrow$)}} \\
\cmidrule(lr){2-3} \cmidrule(lr){4-5} \cmidrule(lr){6-9}
& R@P=0.9 & pAUPRC
& R@P=0.9 & pAUPRC
& $\mathcal{M}_K$ & $\mathcal{M}_R$ & $\mathcal{M}_C$ & Mean \\
\midrule
None
  & 0.434 & 0.421
  & 0.694 & 0.675
  & 0.201 & 0.111 & 0.035 & 0.116 \\
Platt
  & 0.424 & 0.411
  & 0.683 & 0.676
  & 0.114 & 0.082 & 0.032 & 0.076 \\
Beta
  & 0.482 & 0.471
  & 0.715 & 0.706
  & \first{0.030} & 0.110 & 0.038 & 0.059 \\
\textbf{Isotonic (Ours)}
  & \first{0.503} & \first{0.488}
  & \first{0.733} & \first{0.717}
  & 0.033 & \first{0.021} & \first{0.029} & \first{0.028} \\
\bottomrule
\end{tabular}
}
\end{table}


\begin{figure}[t]
\centering
    \includegraphics[width=\columnwidth]{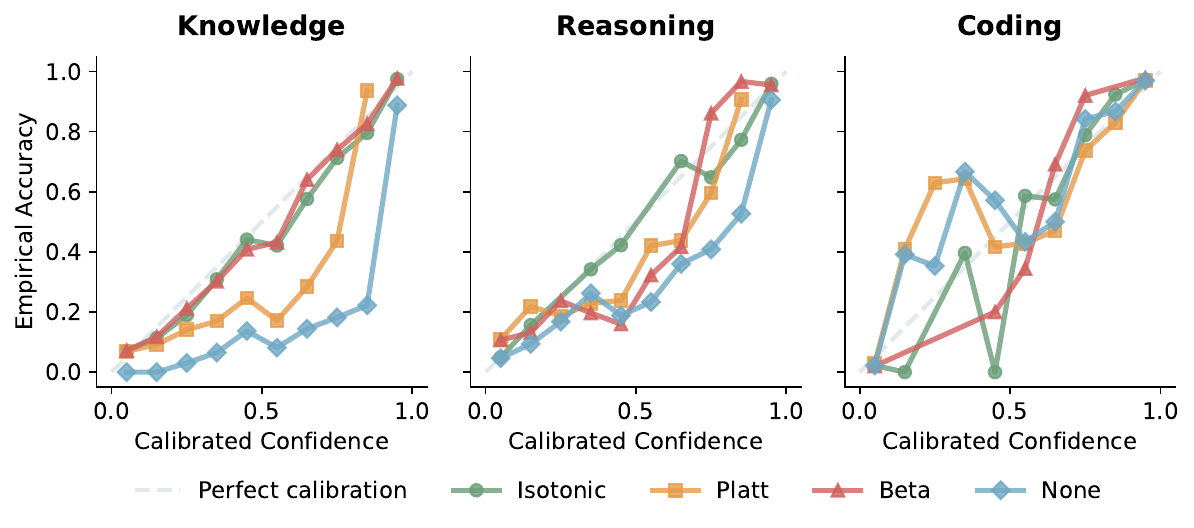}
\caption{calibration reliability diagram}
\label{fig:calibration_diagram}
\end{figure}

\underline{Expected Calibration Error (ECE)~\cite{naeini2015obtaining}.} This metric quantifies how well a model's predicted confidence scores align with its empirical accuracy. Intuitively, a well-calibrated model should be correct $p\%$ of the time on predictions made with confidence $p$.

Formally, predictions are grouped into $m$ equal-width bins based on their calibrated confidence scores. ECE is then computed as the weighted average of the absolute difference between mean confidence and empirical accuracy across bins:
\begin{equation}
  \text{ECE} = \sum_{i=1}^{m} \frac{n_i}{N} \left| \text{conf}_i - \text{acc}_i \right|
\end{equation}
\noindent where $n_i$ is the number of predictions in bin $i$, $N$ is the total number of predictions, $\text{conf}_i$ is the mean confidence of predictions in bin $i$, and $\text{acc}_i$ is the empirical accuracy (fraction of correct predictions) in bin $i$. We use $m = 10$ equal-width bins in all experiments. A lower ECE indicates better calibration, with ECE $= 0$ representing perfect calibration.

Table~\ref{tab:calibration_ablation} shows that isotonic regression achieves the best overall performance. 
Beta calibration excels on \mk (ECE=0.030), where its log-odds parameterization aligns well with log-probability-based confidence, but degrades on \mr (ECE=0.110), where integer-valued verbalized confidence violates its smoothness assumptions. Platt scaling reduces ECE relative to no calibration (0.076 vs.\ 0.116) but underperforms on task metrics, as its sigmoid assumption is too rigid. 
Isotonic regression's non-parametric monotonic fitting 
achieves the most consistent performance across all metrics and both splits. 

Figure~\ref{fig:calibration_diagram} compares the calibration reliability of the three methods, which further confirms that isotonic regression produces well-calibrated confidence across all specialists, while beta calibration shows systematic miscalibration for \mr, and the uncalibrated baseline severely underestimates true correctness probability for \mk.

\begin{figure}[t]
\centering
    \includegraphics[width=.8\columnwidth]{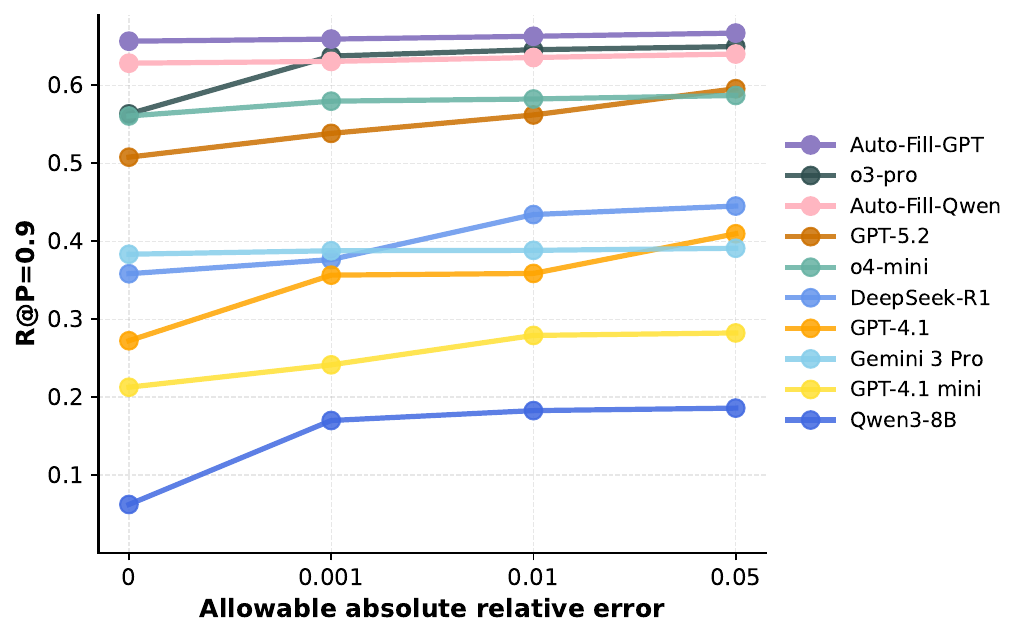}
\caption{%
  \rev{R@P=0.9 as a function of the relative-error tolerance $\epsilon$ used to judge correctness ($\epsilon=0$ is exact-match).}
}
\label{fig:relative-error}
\end{figure}

\subsection{Detailed Cost Analysis}
\label{appendix:cost-analysis}
Tables~\ref{tab:model_pricing} and~\ref{tab:final_cost} give pricing details and a full per-dataset cost breakdown for all evaluated models, complementing the mean-cost figures reported in
Table~\ref{tab:abstain_combined}. Table~\ref{tab:model_pricing}
reports the lowest publicly available API prices we found for each model at the time of writing (February 2026), measured in USD per 1M tokens. The per-dataset breakdown reveals that cost variation across datasets is driven primarily by differences in average table size, which ranges from $89 \pm 110$ cells (\webwiki) to $1{,}359 \pm 774$ cells (\fd), with an overall average of $494 \pm 560$ cells across all 2{,}200 cases. \at-Qwen remains the most cost-efficient option across every dataset, spending on average \$0.00148 per query---over $100\times$ less than o3-pro and $29\times$ less than Gemini~3~Pro.

\subsection{Precision-Recall Curves}
\label{appendix:pr-curve}
Figures~\ref{fig:pr-curve-avg} and~\ref{fig:pr-curve-all} present
the macro-averaged and per-dataset precision-recall curves in the
high-precision regime, offering a curve-level view of the
quality-coverage trade-off summarized by R@P=0.9 and pAUPRC in the main text. For each dataset, predictions are sorted by descending calibrated confidence, and precision and recall are computed at each distinct confidence threshold, where recall is measured as over all 200 cases. The macro-averaged curve averages recall across all 11 datasets at each precision level, using the same contiguous-prefix convention as R@P=0.9 (break at the first drop below the threshold), so that the macro-averaged recall at precision $= 0.9$ matches the per-dataset R@P=0.9 values reported in Table~\ref{tab:abstain_combined}.
Both \at variants extend their curves to the highest recall values on the macro-averaged plot, confirming that calibrated specialization achieves superior coverage across all datasets. On the per-dataset plot, the coding specialist's execution-backed confidence enables near-perfect recall on relational datasets (\ar, \fd, \st), while \at-GPT's broader world knowledge gives it a clear advantage on the hardest OOD dataset (\webtable), where most baselines collapse to zero recall.

\subsection{Additional Precision Thresholds}
\label{appendix:full-and-rp80}
While our primary evaluation targets the high-precision regime (P$\geq$0.9), we include R@P=0.8 (Table~\ref{tab:r08}) and full-accuracy results (Table~\ref{tab:accuracy}) for completeness.
As discussed in Section~\ref{sec:intro}, our problem setting prioritizes high precision because inaccurate suggestions burden users with manual verification and can contaminate downstream analytics; relaxed precision thresholds and unconstrained prediction therefore \emph{do not} reflect our target deployment scenario.

Under R@P=0.8, \at-GPT remains the top-performing method (0.698), while \at-Qwen (0.657) is outperformed by o3-pro (0.679) and o4-mini (0.667) as overconfident frontier models benefit
from the relaxed threshold. However, \at-Qwen achieves this competitive recall at over $100\times$ lower cost than o3-pro, making it a substantially more practical option.

Full accuracy, which applies no abstention, isolates raw prediction quality from calibration quality. Gemini~3~Pro (0.786) and o3-pro (0.757) lead on this metric, yet neither translates this advantage to R@P=0.9 (Table~\ref{tab:abstain_combined}), confirming that confidence calibration---not raw prediction ability---is the key bottleneck in our high-precision setting. Both \at-Qwen (0.690) and \at-GPT (0.727) remain competitive with frontier models at a fraction of their cost. Among the \at-Qwen variants, the Learned Router (0.634) and Classical ML (0.626) trail the main \at setting by a smaller margin than in the high-precision setting, offering cheaper alternatives when strict precision guarantees are not required. The Hybrid Model (0.558) remains the weakest variant, consistent with its collapse under abstention.

\begin{figure}[t]
\centering
    \includegraphics[width=\columnwidth]{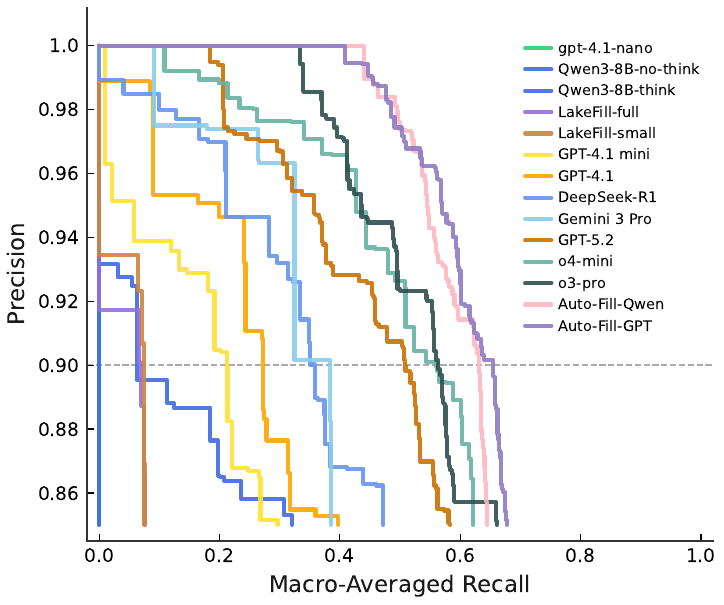}
\caption{%
  Macro-averaged PR curve in the high-precision regime
  (precision $\geq$ 0.85 shown).
}
\label{fig:pr-curve-avg}
\end{figure}

\subsection{Detailed Specialist Combination Ablation}
\label{appendix:specialist-ablation}
Tables~\ref{tab:abstain_90_specialist_ablation}  provides per-dataset breakdowns of the specialist-combination ablation under R@P=0.9, extending the averaged results reported in Table~\ref{tab:abstain_90_specialist_ablation_avg}. 
\mk alone achieves the strongest individual performance (mean 0.542/0.612 for Qwen/GPT), as world knowledge suffices for many ID tables and even some relational tasks. However, it underperforms on \ar (0.550/0.755), where answers require arithmetic computation.
\mr alone shows a more uneven profile: competitive on pattern-heavy and relational datasets (\gov: 0.455/0.420, \st: 0.975/0.945) but collapsing to 0.000 on knowledge-intensive datasets like \pbi and \webtable, where chain-of-thought reasoning cannot compensate for missing world knowledge and may even lead to confidently wrong predictions~\citep{turpin2023language}. 
\mc alone achieves the lowest mean recall (0.344/0.320), as programmatic column patterns exist in only a subset of datasets. Yet it provides an irreplaceable contribution: near-perfect recall on \ar (0.955/0.940), where execution-based confidence is grounded in verifiable column relationships rather than model self-assessment.

Among pairwise combinations, \mk+\mc performs best overall (0.604/0.647), combining strong factual coverage with execution-grounded relational reasoning. \mk+\mr provides complementary gains on knowledge-intensive OOD datasets like \webtable. Neither pairwise combination consistently dominates the other, and both fall short of the full ensemble. Combining all three specialists achieves the best or near-best recall on every dataset (mean 0.628/0.656), confirming that missing-value prediction requires a distinct combination of knowledge, reasoning, and coding capabilities that no single specialist can fully cover.

Table~\ref{tab:accuracy_specialist_ablation} shows the
specialist-combination ablation under full accuracy, where the overall trends are consistent. Notably, \mk+\mr now performs better than \mk+\mc (0.667/0.712 vs.\ 0.664/0.703). The full ensemble still achieves the best mean accuracy (0.690/0.727), confirming that all three specialists contribute to overall prediction quality even beyond the high-precision regime.

\section{Qualitative Examples}
\label{appendix:examples}
This section presents qualitative examples of teacher-generated training traces and specialist predictions at inference time.
\section{\rev{Residual Error Analysis}}
\label{appendix:error-analysis}

\rev{We audit what is left after specialisation, calibration, and ensembling. A full-ensemble failure is a test case on which the highest-confidence specialist of \at-Qwen disagrees with the ground truth; across the 11 datasets this yields 682 failures (\st\ has zero). We draw a stratified random sample of 100, proportional to each dataset's failure count with a floor of one per non-empty dataset, and label each by hand from the input table, the ground truth, and the three specialists' full trajectories. Each failure is assigned to one of three categories according to what would have to change to recover the ground truth. A failure is \emph{within-\mk/\mr/\mc} if a stronger version of the corresponding specialist would solve it; \emph{mis-routed} if one of the other two specialists was already individually correct, so that the ensemble contained the right answer but calibration ranked the wrong one on top; and \emph{unrecoverable} if neither the observed rows and columns of the table nor external knowledge point to the ground truth, leaving any predictor reduced to random guessing. The label refers to the mode of fix, not the chosen specialist: a case in which \mr\ was selected but the ground truth is a factual lookup is counted as within-\mk, because what would fix it is a stronger knowledge specialist rather than stronger reasoning.}

\rev{Table~\ref{tab:error-analysis-breakdown} reports the per-dataset breakdown. The \mk/\mr/\mc\ decomposition absorbs 89 of the 100 failures (52 within-\mk, 23 within-\mr, 4 within-\mc, and 10 mis-routes), indicating that the residual is dominated by capability gaps within individual specialists rather than by missing modalities. Within-\mk\ failures are predictions whose ground truth is a fact about the world that the specialist failed to recall. Some involve facts common enough that a stronger \mk\ would memorise them: on \webtable\ the specialist predicts \codeq{Nintendo DS} as the platform of \emph{Tamagotchi: Party On!} (GT \codeq{Wii}). Others involve facts too specific to be reliably memorized at any realistic scale but reachable by a targeted web query, such as the IAU coordinates of the lunar crater \emph{Cleomedes~S} (\codeq{59.0$^{\circ}$ E}), the tracklist position of the Hindi-film song \emph{Mission Tadofier}, or a competitor's surname on a small French triathlon roster. This subset could in principle be recovered by augmenting \mk\ with a retrieval module. Retrieval belongs to the knowledge modality: the ground truth is a fact about the world, and an external index simply broadens \mk's knowledge store beyond model parameters without altering the solving strategy. We leave retrieval integration to future work, as it would substantially increase per-query cost for a modest aggregate gain.}

\rev{Within-\mr\ and within-\mc\ failures share a common character: a coherent solution attempt that misses on a single step. Within-\mr\ failures are reasoning chains that land one step off the ground truth. On \webwiki, \mr\ correctly enforces a sum-to-100\,\% constraint over an election table and returns \codeq{33.96\%}, off GT \codeq{33.95\%} by a 0.01-point rounding. On \webtable, \mr\ computes offense-per-game with the wrong season length (assuming 8 games where 10 is implied) and returns \codeq{179.6} instead of \codeq{143.7}. Within-\mc\ failures commit to a column-level rule that fits part of the column but breaks on the masked row: on \excelFiftyK, \mc\ uses \codeq{(Max+Min)/2} even though its own trace notes that the formula does not hold across other rows. Mis-routes are concentrated on \enterpriseexcel\ and \webtable.}

\rev{The remaining 11 cases are unrecoverable: the visible rows and columns simply do not determine the ground truth. It arises in tables (particularly private enterprise data) where ad-hoc entries coexist with structured-looking ones, so that the masked cell either contradicts an apparent column-wide rule or takes a value for which the observed rows provide no signal. On \excelFiftyK, a row whose non-target columns are identical to another row labelled \codeq{Moyen} is itself labelled \codeq{Bien}, breaking the only deducible rule (``same features $\Rightarrow$ same label''). On \parquet, the visible sequence \codeq{Talk01, Talk02} extrapolates to \codeq{Talk03}, but the GT \codeq{Thanks00} breaks the naming convention. On \pbi, six dispatchers appear in the previous week and only one continues at a different rate the next week, with nothing in the row indicating which. On \enterpriseexcel, a column otherwise populated with percentages contains the sentinel \codeq{Already Met}, an override unsupported by any column-level rule. These are precisely the cases where, under our high-precision setting (R@P${=}0.9$), a well-calibrated system should abstain rather than commit, and the abstention mechanism (Section~\ref{sec:calibration}) is designed to filter exactly this regime by driving calibrated confidence low when no specialist's evidence is reliable. Counting them as silent deferrals rather than ensemble errors is consistent with the deployment scenario the benchmark targets: a missing-value tool that offers a prediction only when the evidence warrants it, and defers to the user otherwise.}

\rev{Taken together, this analysis confirms that knowledge, reasoning, and code partition the space of solving strategies for missing-value prediction. The residual error lies not in missing modalities but in within-modality capacity, which stronger base models, more sophisticated training, or specialised modules such as retrieval would help close.}

\begin{table}[t!]
\centering
\small
\rev{
\caption{\rev{Per-dataset breakdown of 100 stratified-sampled \at-Qwen full-ensemble failures, labelled by mode of fix. \mk/\mr/\mc\ count within-mode failures---the residual is recoverable by improving the corresponding specialist (\mk\ includes both short-tail recall slips and long-tail lookups recoverable by a retrieval-augmented \mk, which is an extension of the knowledge specialist, not a fourth modality). \emph{Mis-route} marks failures in which another specialist was individually correct (a calibration error, not a capability gap). \emph{Unrecoverable} marks ground truths that cannot be derived from the prompt by any deducible rule and are not retrievable from external knowledge. The first four columns account for 89/100 of the residuals.}}
\label{tab:error-analysis-breakdown}
\setlength{\tabcolsep}{3.5pt}
\begin{tabular}{lcccccc}
\toprule
Dataset & \mk & \mr & \mc & Mis-route & Unrecoverable & $n$ \\
\midrule
\excelFiftyK & 3 & 5 & 1 & 1 & 2 & 12 \\
\pbi & 6 & 1 & 0 & 0 & 2 & 9 \\
\webwiki & 14 & 3 & 0 & 0 & 0 & 17 \\
\gov & 6 & 4 & 0 & 1 & 1 & 12 \\
\parquet & 5 & 3 & 1 & 1 & 1 & 11 \\
\enterprisecosmos & 8 & 1 & 0 & 0 & 1 & 10 \\
\enterpriseexcel & 1 & 3 & 0 & 3 & 1 & 8 \\
\webtable & 8 & 1 & 1 & 4 & 3 & 17 \\
\ar & 0 & 0 & 1 & 0 & 0 & 1 \\
\fd & 1 & 2 & 0 & 0 & 0 & 3 \\
\midrule
\textbf{Total} & \textbf{52} & \textbf{23} & \textbf{4} & \textbf{10} & \textbf{11} & \textbf{100} \\
\bottomrule
\end{tabular}
\label{tab:error_analysis_breakdown}
}
\end{table}

    \begin{table*}[!t]
    \centering
    \caption{\textbf{Model pricing} (USD per 1M tokens).}
    \label{tab:model_pricing}
    \resizebox{\textwidth}{!}{%
        \begin{tabular}{l|rrrr|rrrrrrr|rr}
            \toprule
            & \multicolumn{4}{c|}{\emph{Open-source}} & \multicolumn{9}{c}{\emph{Commercial}} \\
            & Qwen3-1.7B & Qwen3-4B & Qwen3-8B & DeepSeek-R1 & GPT-4o & GPT-4.1 nano & GPT-4.1 mini & GPT-4.1 & o4-mini & o3-pro & GPT-5.2 & \multicolumn{2}{c}{Gemini 3 Pro} \\
            \midrule
            \textbf{Input}  & --- & 0.01 & 0.04 & 0.70 & 2.50 & 0.10 & 0.40 & 2.00 & 1.10 & 20.00 & 1.75 & 2.00 ($\leq$200K) & 4.00 ($>$200K) \\
            \textbf{Output} & --- & 0.03 & 0.14 & 2.50 & 10.00 & 0.40 & 1.60 & 8.00 & 4.40 & 80.00 & 14.00 & 12.00 ($\leq$200K) & 18.00 ($>$200K) \\
            \bottomrule
        \end{tabular}%
        }
\vspace{5pt}
\end{table*}

\begin{table*}[!t]
    \centering
    \caption{Per-dataset inference cost (\$). Values $< \$0.01$ use small scientific notation (\sci{x}{y}). Lower is better. The second row reports the average number of cells per table (mean $\pm$ std) for each benchmark.}
    \label{tab:final_cost}
    
    \resizebox{\textwidth}{!}{%
        \begin{tabular}{l|ccccc|cccccc|c}
            \toprule
            & \multicolumn{5}{c}{\textbf{In-Distribution (ID)}} & \multicolumn{6}{c}{\textbf{Out-of-Distribution (OOD)}} & \\
            \textbf{Model} & \textbf{\excelFiftyK} & \textbf{\pbi} & \textbf{\webwiki} & \textbf{\gov} & \textbf{\parquet} & \textbf{\enterprisecosmos} & \textbf{\enterpriseexcel} & \textbf{\webtable} & \textbf{\ar} & \textbf{\fd} & \textbf{\st} & \textbf{Mean} \\
            \textit{\#cells} &
            \textit{235{$\pm$265}} & \textit{114{$\pm$196}} & \textit{89{$\pm$110}} & \textit{343{$\pm$394}} & \textit{386{$\pm$403}} &
            \textit{790{$\pm$408}} & \textit{524{$\pm$595}} & \textit{135{$\pm$129}} & \textit{671{$\pm$360}} & \textit{1359{$\pm$774}} & \textit{789{$\pm$509}} & \textit{494{$\pm$560}} \\
            \midrule
            DeepSeek-R1 & 
            \sci{6.15}{3} & \sci{4.09}{3} & \sci{4.33}{3} & \sci{6.12}{3} & \sci{6.88}{3} & 
            0.010 & \sci{6.86}{3} & \sci{4.32}{3} & \sci{6.04}{3} & \sci{8.15}{3} & \sci{5.11}{3} & \sci{6.21}{3} \\
            
            o4-mini & 
            \sci{8.89}{3} & \sci{6.10}{3} & 0.012 & \sci{9.93}{3} & 0.012 & 
            0.017 & 0.014 & 0.010 & \sci{9.29}{3} & 0.013 & \sci{7.70}{3} & 0.011 \\
            
            GPT-5.2 & 
            \sci{3.07}{3} & \sci{1.72}{3} & \sci{1.39}{3} & \sci{4.38}{3} & \sci{7.01}{3} & 
            0.020 & \sci{7.09}{3} & \sci{1.82}{3} & \sci{7.88}{3} & 0.018 & \sci{8.89}{3} & \sci{7.31}{3} \\
            
            Gemini 3 Pro & 
            0.045 & 0.031 & 0.031 & 0.034 & 0.047 & 
            0.080 & 0.046 & 0.032 & 0.038 & 0.067 & 0.024 & 0.043 \\
            
            o3-pro & 
            0.162 & 0.098 & 0.134 & 0.148 & 0.204 & 
            0.313 & 0.203 & 0.139 & 0.196 & 0.224 & 0.132 & 0.177 \\
            
            GPT-4.1 & 
            \sci{3.65}{3} & \sci{2.07}{3} & \sci{1.59}{3} & \sci{5.23}{3} & \sci{8.15}{3} & 
            0.021 & \sci{8.24}{3} & \sci{2.15}{3} & \sci{9.28}{3} & 0.028 & 0.010 & \sci{9.02}{3} \\
            
            \midrule
            
            \textbf{\at-Qwen} & 
            \first{\sci{1.30}{3}} & \first{\sci{9.00}{4}} & \first{\sci{9.00}{4}} & \first{\sci{1.30}{3}} & \first{\sci{1.90}{3}} & 
            \first{\sci{2.20}{3}} & \first{\sci{1.70}{3}} & \first{\sci{9.00}{4}} & \first{\sci{1.40}{3}} & \first{\sci{2.50}{3}} & \first{\sci{1.30}{3}} & \first{\sci{1.48}{3}} \\
            
            \textbf{\at-GPT} & 
            0.011 & \sci{8.12}{3} & \sci{8.69}{3} & 0.011 & 0.013 & 
            0.020 & 0.013 & \sci{9.34}{3} & \sci{8.52}{3} & 0.014 & \sci{7.94}{3} & 0.011 \\
            
            \bottomrule
        \end{tabular}%
    }
\vspace{3cm}
\end{table*}

\begin{figure*}[!t]
\centering
    \includegraphics[width=\textwidth]{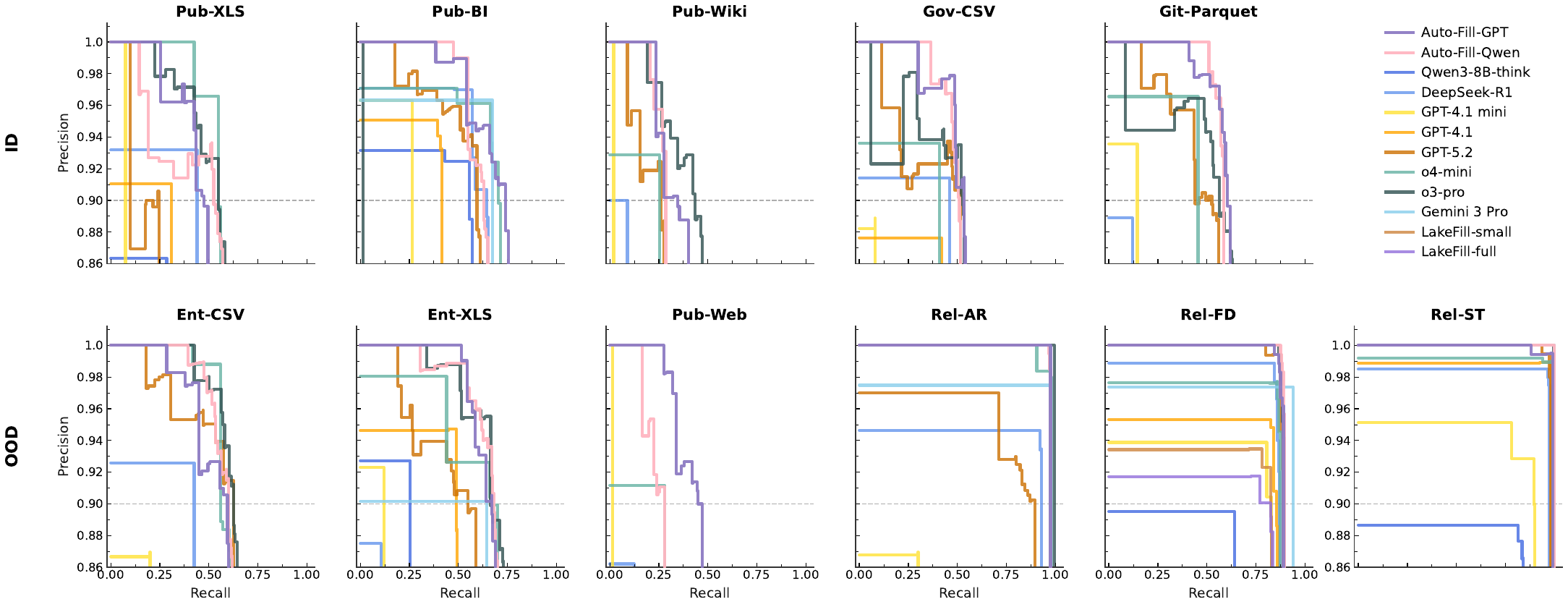}
\caption{%
  Per-dataset PR curves in the high-precision regime for all 11 benchmarks.
}
\label{fig:pr-curve-all}
\end{figure*}

\clearpage

\begin{table*}[!t]
    \centering
    \caption{R@P=0.8 performance of all methods. Cost reflects LLM inference per query only. \firstT{First}, \secondT{second}, and \thirdT{third} best results per column are highlighted.}
    \label{tab:r08}
    \resizebox{\textwidth}{!}{%
        \begin{tabular}{l|ccccc|cccccc|c|c}
            \toprule
            & \multicolumn{5}{c}{\textbf{In-Distribution (ID)}} & \multicolumn{6}{c}{\textbf{Out-of-Distribution (OOD)}} & & \\
            \textbf{Method} & \textbf{\excelFiftyK} & \textbf{\pbi} & \textbf{\webwiki} & \textbf{\gov} & \textbf{\parquet} & \textbf{\enterprisecosmos} & \textbf{\enterpriseexcel} & \textbf{\webtable} & \textbf{\ar} & \textbf{\fd} & \textbf{\st} & \textbf{Mean ($\uparrow$)} & \textbf{Cost (\$) ($\downarrow$)} \\
            \midrule
            \rowcolor{black!5}\multicolumn{14}{c}{\textbf{\textit{Non-LLM baselines}}} \\
            LakeFill-small &
            0.000 & 0.000 & 0.000 & 0.000 & 0.000 &
            0.000 & 0.000 & 0.000 & 0.000 & 0.830 & 0.790 & 0.147 & --- \\
            LakeFill-full &
            0.000 & 0.000 & 0.000 & 0.000 & 0.000 &
            0.000 & 0.515 & 0.000 & 0.000 & 0.835 & 0.785 & 0.194 & --- \\
            \rev{TabPFN} &
            \rev{0.090} & \rev{0.155} & \rev{0.000} & \rev{0.220} & \rev{0.350} &
            \rev{0.525} & \rev{0.445} & \rev{0.000} & \rev{0.120} & \rev{0.690} & \rev{0.035} & \rev{0.239} & \rev{---} \\
            \rev{Baran} &
            \rev{0.185} & \rev{0.170} & \rev{0.000} & \rev{0.290} & \rev{0.375} &
            \rev{0.545} & \rev{0.505} & \rev{0.000} & \rev{0.145} & \rev{0.695} & \rev{0.030} & \rev{0.267} & \rev{---} \\
            \rev{SCARE} &
            \rev{0.080} & \rev{0.065} & \rev{0.000} & \rev{0.220} & \rev{0.290} &
            \rev{0.475} & \rev{0.450} & \rev{0.000} & \rev{0.115} & \rev{0.660} & \rev{0.030} & \rev{0.217} & \rev{---} \\
            \midrule
            \rowcolor{black!5}\multicolumn{14}{c}{\textbf{\textit{LLM baselines}}} \\
            Qwen3-8B &
            0.285 & 0.595 & 0.000 & 0.285 & 0.285 &
            0.145 & 0.350 & 0.220 & 0.805 & 0.660 & 0.840 & 0.406 & \first{\sci{5.36}{4}} \\
            GPT-4.1 mini &
            0.075 & 0.265 & 0.100 & 0.285 & 0.340 &
            0.200 & 0.430 & 0.125 & 0.300 & 0.830 & 0.900 & 0.350 & \third{\sci{1.81}{3}} \\
            GPT-4.1 &
            0.330 & 0.590 & 0.155 & 0.420 & 0.475 &
            0.475 & 0.495 & 0.290 & 0.775 & 0.860 & 0.985 & 0.532 & \sci{9.02}{3} \\
            DeepSeek-R1 &
            0.495 & 0.730 & 0.090 & 0.525 & 0.515 &
            0.495 & 0.540 & \third{0.390} & 0.930 & 0.870 & 0.975 & 0.596 & \sci{6.21}{3} \\
            o4-mini &
            \second{0.615} & 0.685 & 0.255 & \third{0.550} & \third{0.625} &
            0.620 & 0.715 & \second{0.450} & 0.970 & 0.865 & 0.985 & \third{0.667} & \sci{1.09}{2} \\
            GPT-5.2 &
            0.490 & 0.695 & \third{0.310} & 0.525 & 0.615 &
            \second{0.645} & 0.725 & 0.000 & 0.895 & \third{0.890} & 0.980 & 0.615 & \sci{7.31}{3} \\
            Gemini 3 Pro &
            \third{0.560} & \second{0.770} & 0.000 & 0.000 & 0.615 &
            0.570 & \first{0.785} & 0.000 & 0.975 & \first{0.940} & \first{1.000} & 0.565 & \sci{4.30}{2} \\
            o3-pro &
            \first{0.625} & \first{0.790} & \first{0.540} & \second{0.570} & \second{0.655} &
            \first{0.660} & \second{0.740} & 0.000 & \first{0.995} & \second{0.895} & \second{0.995} & \second{0.679} & \sci{1.77}{1} \\
            \midrule
            \rowcolor{black!5}\multicolumn{14}{c}{\textbf{\textit{Ours}}} \\
            \textbf{\at-Qwen} &
            \third{0.560} & 0.680 & 0.305 & 0.540 & 0.600 &
            \third{0.625} & 0.690 & 0.355 & \second{0.990} & \third{0.890} & \second{0.995} & 0.657 & \second{\sci{1.48}{3}} \\
            \textbf{\at-GPT} &
            \third{0.560} & \third{0.750} & \second{0.430} & \first{0.580} & \first{0.660} &
            0.620 & \third{0.730} & \first{0.480} & \second{0.990} & \third{0.890} & 0.985 & \first{0.698} & \sci{1.13}{2} \\
            \bottomrule
        \end{tabular}%
    }
\end{table*}

\begin{table*}[!t]
    \centering
    \caption{Full accuracy (no abstention) performance of all methods. The \emph{FD upper-bound} row reports full-distribution recall. Cost reflects LLM inference per query only. \firstT{First}, \secondT{second}, and \thirdT{third} best results per column are highlighted.}
    \label{tab:accuracy}
    \resizebox{\textwidth}{!}{%
        \begin{tabular}{l|ccccc|cccccc|c|c}
            \toprule
            & \multicolumn{5}{c}{\textbf{In-Distribution (ID)}} & \multicolumn{6}{c}{\textbf{Out-of-Distribution (OOD)}} & & \\
            \textbf{Method} & \textbf{\excelFiftyK} & \textbf{\pbi} & \textbf{\webwiki} & \textbf{\gov} & \textbf{\parquet} & \textbf{\enterprisecosmos} & \textbf{\enterpriseexcel} & \textbf{\webtable} & \textbf{\ar} & \textbf{\fd} & \textbf{\st} & \textbf{Mean ($\uparrow$)} & \textbf{Cost (\$) ($\downarrow$)} \\
            \midrule
            \rowcolor{black!5}\multicolumn{14}{c}{\textbf{\textit{Non-LLM baselines}}} \\
            FD upper-bound$^*$ &
            0.060 & 0.160 & 0.080 & 0.070 & 0.230 &
            0.340 & 0.130 & 0.070 & 0.060 & 0.480 & 0.000 & 0.153 & --- \\
            LakeFill-small &
            0.320 & 0.320 & 0.305 & 0.345 & 0.380 &
            0.530 & 0.540 & 0.245 & 0.560 & 0.830 & 0.790 & 0.470 & --- \\
            LakeFill-full &
            0.285 & 0.295 & 0.240 & 0.390 & 0.355 &
            0.520 & 0.630 & 0.265 & 0.585 & 0.835 & 0.785 & 0.471 & --- \\
            \rev{TabPFN} &
            \rev{0.220} & \rev{0.290} & \rev{0.170} & \rev{0.330} & \rev{0.405} &
            \rev{0.530} & \rev{0.505} & \rev{0.150} & \rev{0.160} & \rev{0.690} & \rev{0.090} & \rev{0.322} & \rev{---} \\
            \rev{Baran} &
            \rev{0.235} & \rev{0.300} & \rev{0.210} & \rev{0.345} & \rev{0.420} &
            \rev{0.550} & \rev{0.515} & \rev{0.180} & \rev{0.180} & \rev{0.695} & \rev{0.110} & \rev{0.340} & \rev{---} \\
            \rev{SCARE} &
            \rev{0.210} & \rev{0.285} & \rev{0.165} & \rev{0.340} & \rev{0.440} &
            \rev{0.510} & \rev{0.525} & \rev{0.165} & \rev{0.165} & \rev{0.670} & \rev{0.095} & \rev{0.325} & \rev{---} \\
            \midrule
            \rowcolor{black!5}\multicolumn{14}{c}{\textbf{\textit{LLM baselines}}} \\
            Qwen3-8B &
            0.500 & 0.670 & 0.330 & 0.485 & 0.515 &
            0.455 & 0.525 & 0.435 & 0.805 & 0.665 & 0.840 & 0.566 & \third{\sci{5.36}{4}} \\
            GPT-4.1 mini &
            0.470 & 0.670 & 0.395 & 0.470 & 0.555 &
            0.565 & 0.610 & 0.440 & 0.700 & 0.830 & 0.900 & 0.600 & \sci{1.81}{3} \\
            GPT-4.1 &
            0.535 & 0.705 & 0.450 & 0.540 & 0.600 &
            0.600 & 0.645 & 0.535 & 0.785 & 0.860 & 0.985 & 0.658 & \sci{9.02}{3} \\
            DeepSeek-R1 &
            0.590 & 0.735 & \third{0.510} & 0.600 & 0.625 &
            0.585 & 0.645 & 0.565 & 0.930 & 0.870 & 0.975 & 0.694 & \sci{6.21}{3} \\
            o4-mini &
            \third{0.620} & 0.750 & 0.470 & 0.605 & 0.645 &
            0.650 & 0.720 & 0.555 & 0.970 & 0.865 & 0.985 & 0.712 & \sci{1.09}{2} \\
            GPT-5.2 &
            0.580 & 0.730 & 0.505 & 0.600 & 0.655 &
            \third{0.655} & 0.735 & \third{0.575} & 0.895 & \third{0.890} & 0.980 & 0.709 & \sci{7.31}{3} \\
            Gemini 3 Pro &
            \first{0.660} & \second{0.775} & \first{0.725} & \first{0.665} & \first{0.720} &
            \first{0.690} & \first{0.785} & \first{0.710} & \third{0.975} & \first{0.940} & \first{1.000} & \first{0.786} & \sci{4.30}{2} \\
            o3-pro &
            \second{0.655} & \first{0.790} & \second{0.645} & \second{0.630} & \third{0.660} &
            \second{0.665} & \second{0.745} & \second{0.650} & \first{0.995} & \second{0.895} & \second{0.995} & \second{0.757} & \sci{1.77}{1} \\
            \midrule
            \rowcolor{black!5}\multicolumn{14}{c}{\textbf{\textit{Ours}}} \\
            \textbf{\at-Qwen} &
            0.595 & 0.725 & 0.425 & 0.565 & 0.625 &
            \third{0.655} & 0.695 & 0.425 & \second{0.990} & \third{0.890} & \first{1.000} & 0.690 & \sci{1.48}{3} \\
            \textbf{\at-GPT} &
            0.600 & \third{0.770} & \third{0.510} & \third{0.610} & \second{0.675} &
            \third{0.655} & \third{0.740} & 0.565 & \second{0.990} & \second{0.895} & \third{0.990} & \third{0.727} & \sci{1.13}{2} \\
            \midrule
            \rowcolor{black!5}\multicolumn{14}{c}{\textbf{\textit{\at-Qwen Variants}}} \\
            Hybrid Model &
            0.495 & 0.655 & 0.250 & 0.500 & 0.540 &
            0.445 & 0.605 & 0.320 & 0.600 & 0.765 & 0.960 & 0.558 & \first{\sci{5.45}{5}} \\
            Learned Router &
            0.550 & 0.660 & 0.345 & 0.565 & 0.605 &
            0.575 & 0.610 & 0.405 & 0.860 & 0.830 & 0.965 & 0.634 & \second{\sci{4.58}{4}} \\
            Classical ML &
            0.520 & 0.655 & 0.340 & 0.510 & 0.560 &
            0.545 & 0.585 & 0.400 & 0.960 & 0.850 & 0.960 & 0.626 & \sci{1.48}{3} \\
            \bottomrule
        \end{tabular}%
    }
\end{table*}

\clearpage

\begin{table*}[!t]
    \centering
    \caption{Full ablation study on specialist combinations across 11 datasets (R@P=0.9).}
    \label{tab:abstain_90_specialist_ablation}
    \resizebox{\textwidth}{!}{%
        \begin{tabular}{ll|ccccc|cccccc|c}
            \toprule
            & & \multicolumn{5}{c}{\textbf{In-Distribution (ID)}} & \multicolumn{6}{c}{\textbf{Out-of-Distribution (OOD)}} & \\
            \textbf{Base Model} & \textbf{Specialist(s)} & \textbf{\excelFiftyK} & \textbf{\pbi} & \textbf{\webwiki} & \textbf{\gov} & \textbf{\parquet} & \textbf{\enterprisecosmos} & \textbf{\enterpriseexcel} & \textbf{\webtable} & \textbf{\ar} & \textbf{\fd} & \textbf{\st} & \textbf{Mean} \\
            \midrule
            \multirow{7}{*}{\makecell[l]{\textbf{Qwen3-8B}}} 
            & \mk only & 0.335 & 0.580 & 0.210 & 0.450 & 0.565 & 0.570 & 0.640 & 0.235 & 0.550 & 0.865 & 0.960 & 0.542 \\
            & \mr only & 0.360 & 0.000 & 0.000 & 0.455 & 0.475 & 0.115 & 0.535 & 0.000 & 0.815 & 0.865 & 0.975 & 0.418 \\
            & \mc only & 0.235 & 0.000 & 0.110 & 0.150 & 0.305 & 0.215 & 0.315 & 0.100 & 0.955 & 0.590 & 0.805 & 0.344 \\
            \cmidrule{2-14}
            & \mk + \mr & 0.460 & 0.570 & \second{0.265} & \second{0.470} & 0.565 & 0.555 & \third{0.650} & \second{0.240} & 0.835 & \second{0.885} & 0.985 & 0.589 \\
            & \mk + \mc & \second{0.465} & \second{0.590} & 0.245 & 0.450 & \second{0.585} & \first{0.590} & \first{0.660} & \second{0.240} & \second{0.975} & 0.865 & 0.975 & \second{0.604} \\
            & \mr + \mc & 0.430 & 0.465 & 0.000 & 0.465 & 0.495 & 0.215 & 0.550 & 0.000 & 0.965 & 0.870 & \second{0.990} & 0.495 \\
            \cmidrule{2-14}
            & \textbf{All (Ours)} & \first{0.525} & \first{0.615} & \first{0.285} & \first{0.500} & \first{0.590} & \second{0.585} & \first{0.660} & \first{0.275} & \first{0.990} & \first{0.890} & \first{0.995} & \first{0.628} \\
            \midrule
            \multirow{7}{*}{\makecell[l]{\textbf{GPT-4.1 mini}}}
            & \mk only & 0.425 & 0.645 & 0.325 & 0.530 & 0.580 & \third{0.590} & 0.665 & 0.365 & 0.755 & \second{0.875} & \second{0.980} & 0.612 \\
            & \mr only & 0.100 & 0.570 & 0.100 & 0.420 & 0.500 & 0.015 & 0.440 & 0.145 & 0.910 & 0.825 & 0.945 & 0.452 \\
            & \mc only & 0.245 & 0.320 & 0.000 & 0.150 & 0.220 & 0.175 & 0.000 & 0.100 & 0.940 & 0.550 & 0.815 & 0.320 \\
            \cmidrule{2-14}
            & \mk + \mr & 0.420 & \second{0.695} & \first{0.340} & 0.515 & \third{0.590} & 0.540 & 0.645 & \first{0.440} & 0.895 & \second{0.875} & \second{0.980} & 0.630 \\
            & \mk + \mc & \first{0.510} & 0.680 & 0.305 & \second{0.535} & \first{0.600} & \first{0.605} & \first{0.690} & 0.370 & 0.970 & \second{0.875} & \second{0.980} & \second{0.647} \\
            & \mr + \mc & 0.240 & 0.580 & 0.000 & 0.375 & 0.540 & 0.360 & 0.000 & 0.180 & \second{0.985} & 0.840 & 0.965 & 0.460 \\
            \cmidrule{2-14}
            & \textbf{All (Ours)} & \second{0.470} & \first{0.705} & \second{0.330} & \first{0.545} & \first{0.600} & \first{0.605} & \second{0.685} & \second{0.415} & \first{0.990} & \first{0.890} & \first{0.985} & \first{0.656} \\
            \bottomrule
        \end{tabular}%
    }
    \vspace{3cm}
\end{table*}

\begin{table*}[!t]
    \centering
    \caption{Full ablation study on specialist combinations across 11 datasets (full accuracy).}
    \label{tab:accuracy_specialist_ablation}
    \resizebox{\textwidth}{!}{%
        \begin{tabular}{ll|ccccc|cccccc|c}
            \toprule
            & & \multicolumn{5}{c}{\textbf{In-Distribution (ID)}} & \multicolumn{6}{c}{\textbf{Out-of-Distribution (OOD)}} & \\
            \textbf{Base Model} & \textbf{Specialist(s)} & \textbf{\excelFiftyK} & \textbf{\pbi} & \textbf{\webwiki} & \textbf{\gov} & \textbf{\parquet} & \textbf{\enterprisecosmos} & \textbf{\enterpriseexcel} & \textbf{\webtable} & \textbf{\ar} & \textbf{\fd} & \textbf{\st} & \textbf{Mean} \\
            \midrule
            \multirow{7}{*}{\makecell[l]{\textbf{Qwen3-8B}}}
            & \mk only & 0.495 & 0.695 & 0.365 & 0.495 & 0.595 & 0.625 & 0.670 & 0.365 & 0.580 & 0.865 & 0.960 & 0.610 \\
            & \mr only & 0.555 & 0.670 & 0.355 & \first{0.575} & 0.610 & 0.575 & 0.610 & 0.410 & 0.855 & 0.865 & 0.975 & 0.641 \\
            & \mc only & 0.315 & 0.365 & 0.165 & 0.245 & 0.305 & 0.320 & 0.415 & 0.255 & 0.955 & 0.615 & 0.805 & 0.433 \\
            \cmidrule{2-14}
            & \mk + \mr & \second{0.575} & \second{0.715} & \second{0.380} & \second{0.565} & \first{0.630} & 0.625 & \second{0.690} & \second{0.420} & 0.865 & \second{0.885} & 0.985 & \second{0.667} \\
            & \mk + \mc & 0.545 & 0.710 & \second{0.380} & 0.505 & 0.605 & \second{0.645} & 0.685 & 0.390 & \second{0.985} & 0.875 & 0.980 & 0.664 \\
            & \mr + \mc & 0.570 & 0.680 & 0.365 & \second{0.565} & 0.610 & 0.575 & 0.630 & \second{0.420} & \second{0.985} & 0.870 & \second{0.990} & 0.660 \\
            \cmidrule{2-14}
            & \textbf{All (Ours)} & \first{0.595} & \first{0.725} & \first{0.425} & \second{0.565} & \second{0.625} & \first{0.655} & \first{0.695} & \first{0.425} & \first{0.990} & \first{0.890} & \first{1.000} & \first{0.690} \\
            \midrule
            \multirow{7}{*}{\makecell[l]{\textbf{GPT-4.1 mini}}}
            & \mk only & 0.545 & 0.725 & 0.460 & 0.595 & 0.650 & 0.630 & 0.705 & 0.500 & 0.775 & 0.875 & \second{0.980} & 0.676 \\
            & \mr only & 0.565 & 0.710 & 0.485 & 0.580 & 0.625 & 0.580 & 0.610 & 0.480 & 0.910 & 0.835 & 0.945 & 0.666 \\
            & \mc only & 0.265 & 0.355 & 0.115 & 0.170 & 0.230 & 0.180 & 0.295 & 0.195 & 0.940 & 0.550 & 0.815 & 0.374 \\
            \cmidrule{2-14}
            & \mk + \mr & \first{0.605} & \second{0.760} & \first{0.515} & \first{0.625} & \first{0.675} & 0.635 & 0.705 & \second{0.560} & 0.895 & \second{0.880} & \second{0.980} & \second{0.712} \\
            & \mk + \mc & 0.585 & 0.740 & 0.460 & 0.605 & 0.660 & \second{0.640} & \second{0.715} & 0.495 & \third{0.975} & 0.875 & \second{0.980} & 0.703 \\
            & \mr + \mc & 0.580 & 0.715 & 0.495 & 0.580 & 0.595 & 0.590 & 0.645 & 0.500 & \second{0.985} & 0.865 & \second{0.980} & 0.685 \\
            \cmidrule{2-14}
            & \textbf{All (Ours)} & \second{0.600} & \first{0.770} & \second{0.510} & \second{0.610} & \first{0.675} & \first{0.655} & \first{0.740} & \first{0.565} & \first{0.990} & \first{0.895} & \first{0.990} & \first{0.727} \\
            \bottomrule
        \end{tabular}%
    }
\end{table*}

    \definecolor{promptcolor}{HTML}{2471A3}      
\definecolor{completioncolor}{HTML}{2471A3}  
\definecolor{thinkcolor}{HTML}{FF69B4}       
\definecolor{missingcolor}{HTML}{FF69B4}
\definecolor{outputcolor}{HTML}{2471A3}

\lstset{
  basicstyle=\ttfamily\small,
  breaklines=true,
  breakatwhitespace=false,
  frame=single,
  escapeinside={(*}{*)},   
  title=\lstname,
  inputencoding=utf8,
  extendedchars=true,
  literate={≈}{{\ensuremath{\approx}}}1,
}
\renewcommand{\lstlistingname}{Example}
\providecommand*{\lstlistingautorefname}{Example}

\clearpage
\onecolumn

\subsection{Teacher-Generated Training Traces}
\label{appendix:teacher-traces}
We show example traces generated by $\mathcal{T}$ during training data generation, covering reasoning traces (with and without verbalized confidence) and coding traces.
\vspace{5pt}

\begin{lstlisting}[
  title={Example of one trace generated by $\mathcal{T}$ without confidence on \parquet for \mr training.},
  basicstyle=\ttfamily\footnotesize,
  breaklines=true,
  breakatwhitespace=false,
  columns=fullflexible,
]
(*\textbf{\textcolor{promptcolor}{\underline{Prompt:}}}*)
Please fill in the missing value in the input table. The missing value is denoted by '(*\textcolor{missingcolor}{[MISSING]}*)'. Please return the value filled in JSON format: {"value": "filled_value"}.

Input Table:
| Sku | jan | feb | mar | apr | may | jun | jul | aug | sep | oct | nov | dec | unit cost | lead-time | retail_price | quantity_on_hand | backlog |
|:----------|:------|:------|:------|:------|:------|:------|:------|:------|:------|:------|:------|:------|:------------|:------------
|:---------------|:-------------------|:----------|
| KR202-209 | 1509 | 1855 | 2665 | 1841 | 1231 | 2598 | 1988 | 1988 | 2927 | 2707 | 731 | 2598 | 1001 | 2 | 5000 | 1003 | 10 |
| KR202-210 | 1006 | 206 | 2588 | 670 | 2768 | 2809 | 1475 | 1537 | 919 | 2525 | 440 | 2691 | 394 | 2 | 1300 | 3224 | 10 |
| (*\textcolor{missingcolor}{[MISSING]}*) | 1840 | 2284 | 850 | 983 | 2737 | 1264 | 2002 | 1980 | 235 | 1489 | 218 | 525 | 434 | 4 | 1200 | 390 | 10 |
| KR202-212 | 104 | 2262 | 350 | 528 | 2570 | 1216 | 1101 | 2755 | 2856 | 2381 | 1867 | 2743 | 474 | 3 | 10 | 390 | 10 |
| KR202-213 | 489 | 954 | 1112 | 199 | 919 | 330 | 561 | 2372 | 921 | 1587 | 1532 | 1512 | 514 | 1 | 2000 | 2095 | 10 |
| KR202-214 | 2416 | 2010 | 2527 | 1409 | 1059 | 890 | 2837 | 276 | 987 | 2228 | 1095 | 1396 | 554 | 2 | 1800 | 55 | 10 |
| KR202-215 | 403 | 1737 | 753 | 1982 | 2775 | 380 | 1561 | 1230 | 1262 | 2249 | 824 | 743 | 594 | 1 | 2500 | 4308 | 10 |
| KR202-216 | 2908 | 929 | 684 | 2618 | 1477 | 1508 | 765 | 43 | 2550 | 2157 | 937 | 1201 | 634 | 3 | 3033 | 34 | 10 |
| KR202-217 | 2799 | 2197 | 1647 | 2263 | 224 | 2987 | 2366 | 588 | 1140 | 869 | 1707 | 1180 | 674 | 3 | 5433 | 390 | 10 |
| KR202-218 | 1333 | 402 | 804 | 318 | 1408 | 830 | 1028 | 534 | 1871 | 2730 | 2022 | 94 | 714 | 2 | 3034 | 3535 | 10 |
| KR202-219 | 813 | 969 | 745 | 1001 | 2732 | 1987 | 717 | 599 | 2722 | 171 | 639 | 2108 | 754 | 3 | 5000 | 334 | 10 |
| KR202-220 | 1481 | 905 | 1067 | 2513 | 861 | 1670 | 650 | 2630 | 1245 | 997 | 1936 | 2780 | 794 | 3 | 7500 | 3434 | 10 |
| KR202-221 | 771 | 2941 | 1360 | 2714 | 1801 | 1744 | 1428 | 1660 | 436 | 578 | 1956 | 1101 | 834 | 2 | 4938 | 4433 | 10 |
| KR202-222 | 2349 | 4 | 345 | 524 | 340 | 2698 | 2137 | 1164 | 498 | 1583 | 1241 | 2965 | 874 | 2 | 4922 | 3435 | 10 |
| KR202-223 | 2045 | 2055 | 552 | 81 | 2780 | 176 | 2316 | 1475 | 2566 | 1678 | 1553 | 2745 | 914 | 1 | 4894 | 34533 | 10 |
| KR202-224 | 2482 | 1887 | 1911 | 1446 | 2939 | 1241 | 1281 | 692 | 119 | 627 | 1941 | 1383 | 954 | 2 | 2942 | 33 | 10 |
| KR202-225 | 2744 | 2770 | 2697 | 1726 | 1776 | 2264 | 332 | 2420 | 2722 | 1161 | 1986 | 2587 | 994 | 6 | 8999 | 2000 | 10 |
| KR202-226 | 2509 | 914 | 903 | 877 | 1859 | 2263 | 383 | 593 | 236 | 189 | 920 | 1686 | 1034 | 3 | 4342 | 4344 | 10 |
| KR202-227 | 368 | 2502 | 2955 | 2994 | 1270 | 2884 | 2208 | 699 | 854 | 877 | 2320 | 160 | 1074 | 3 | 4920 | 489 | 10 |
| KR202-228 | 1468 | 1109 | 2464 | 2799 | 948 | 589 | 2858 | 1140 | 501 | 2691 | 93 | 1060 | 1114 | 2 | 15000 | 9439 | 10 |
| KR202-229 | 2114 | 198 | 1479 | 1249 | 1475 | 744 | 407 | 2280 | 226 | 2285 | 796 | 1948 | 1154 | 2 | 13000 | 8939 | 10 |
| KR202-230 | 1023 | 1150 | 1672 | 2026 | 1590 | 441 | 2484 | 2300 | 2928 | 1082 | 2064 | 2412 | 1194 | 2 | 10000 | 349 | 10 |
| KR202-231 | 482 | 546 | 299 | 2304 | 2953 | 1029 | 1863 | 2809 | 454 | 927 | 2488 | 2341 | 1234 | 4 | 9999 | 3434 | 10 |
| KR202-232 | 614 | 2138 | 962 | 2017 | 2398 | 2963 | 2189 | 1804 | 414 | 2016 | 1350 | 2464 | 1274 | 2 | 7500 | 234 | 10 |
| KR202-233 | 2395 | 2521 | 2157 | 728 | 1028 | 43 | 138 | 826 | 570 | 2825 | 181 | 787 | 1314 | 4 | 6000 | 349 | 10 |
| KR202-234 | 1336 | 1478 | 865 | 533 | 1562 | 422 | 2287 | 1302 | 1230 | 1059 | 1153 | 399 | 1354 | 2 | 20000 | 324 | 10 |
| KR202-235 | 2565 | 2762 | 2721 | 1431 | 845 | 2163 | 2413 | 2227 | 1753 | 740 | 1139 | 2300 | 1394 | 3 | 59500 | 850 | 10 |
| KR202-236 | 1912 | 1726 | 1569 | 316 | 71 | 2082 | 108 | 174 | 1974 | 609 | 2896 | 566 | 1434 | 3 | 2300 | 4930 | 10 |
| KR202-237 | 2153 | 1112 | 16 | 130 | 590 | 2619 | 2576 | 2390 | 2567 | 1531 | 842 | 242 | 1474 | 2 | 4500 | 9483 | 10 |
| KR202-238 | 1417 | 2044 | 1981 | 1936 | 2377 | 780 | 1544 | 1521 | 51 | 1056 | 1876 | 1356 | 1514 | 3 | 8000 | 839 | 10 |
| KR202-239 | 2717 | 2186 | 2300 | 677 | 2157 | 2328 | 1917 | 2519 | 561 | 281 | 1162 | 1146 | 1554 | 2 | 39000 | 433 | 10 |
| KR202-240 | 1015 | 741 | 2754 | 2925 | 2302 | 695 | 2869 | 440 | 406 | 1083 | 2334 | 1015 | 1594 | 3 | 3943 | 390 | 10 |
| KR202-241 | 3050 | 1507 | 3637 | 1112 | 1963 | 1675 | 898 | 1986 | 2262 | 3895 | 1229 | 2904 | 769 | 5 | 8007 | 2125 | 10 |
| KR202-242 | 1875 | 2368 | 830 | 823 | 868 | 1409 | 1845 | 3095 | 3247 | 1894 | 2558 | 3048 | 1819 | 1 | 13225 | 1253 | 10 |
| KR202-243 | 1717 | 593 | 3006 | 2935 | 3139 | 2753 | 3247 | 3845 | 1720 | 3413 | 3399 | 2799 | 1120 | 3 | 14682 | 1128 | 10 |
| KR202-244 | 2383 | 2046 | 2487 | 3827 | 1674 | 3118 | 2849 | 2233 | 3888 | 2566 | 2216 | 3817 | 1067 | 5 | 11997 | 1191 | 10 |
| KR202-245 | 1115 | 2694 | 3038 | 3366 | 1058 | 2724 | 2863 | 1930 | 1787 | 838 | 3087 | 1565 | 1623 | 2 | 12876 | 611 | 10 |
| KR202-246 | 3108 | 1197 | 2472 | 1264 | 3179 | 3638 | 1268 | 1581 | 3456 | 1630 | 1788 | 2288 | 608 | 2 | 6548 | 2192 | 10 |
| KR202-247 | 3439 | 1854 | 652 | 1827 | 1645 | 2257 | 2733 | 1337 | 2034 | 2106 | 877 | 2409 | 1578 | 2 | 10463 | 1017 | 10 |

(*\textbf{\textcolor{completioncolor}{\underline{Completion:}}}*)
(*\textit{\textcolor{thinkcolor}{<think>}}*)
Okay, let's see. I need to find the missing SKU in the table. The existing SKUs are from KR202-209 up to KR202-247. Let me check the sequence.

Looking at the list, the first one is KR202-209, then 210, then the missing one, followed by 212, 213, and so on up to 247. So the missing SKU should be between 210 and 212. That would be KR202-211. Let me confirm that there's no other gaps. The next after the missing is 212, so yes, 211 is missing. The other entries are in order, so the missing value is KR202-211.
(*\textit{\textcolor{thinkcolor}{</think>}}*)
(*\textit{\textcolor{outputcolor}{\{"value": "KR202-211"\}}}*)
\end{lstlisting}

\clearpage

\begin{lstlisting}[
  title={Example of a trace generated by $\mathcal{T}$ with verbalized confidence on \gov for \mr training. The reasoning trace is discarded after training data generation; only the confidence score is used in computing Eq.~\ref{eq:reason} for $\text{conf}^{\text{train}}_R$},
  basicstyle=\ttfamily\footnotesize,
  breaklines=true,
  breakatwhitespace=false,
  columns=fullflexible,
]
(*\textbf{\textcolor{promptcolor}{\underline{Prompt:}}}*)
Please fill in the missing value in the input table and provide your confidence level as an integer between 0 (no confidence) and 100 (full confidence). The missing value is denoted by '(*\textcolor{missingcolor}{[MISSING]}*)'. Please return the value filled in JSON format: {"value": "filled_value", "confidence": "confidence_level"}.

Input Table:
| Financial year | Period | Class | Tests | Pass after rectification at MOT test station | Fails | Initial fail rate | Final fail rate | Tests failed with at least one dangerous item | Initial fail rate for tests with at least one dangerous item | Tests failed with no dangerous items | Initial fail rate for tests with no dangerous item |
|:-----------------|:-----------------------------|:------------------------------------------------------------------------|:-----------
|:-----------------------------------------------|:----------|:--------------------|:------------------
|:------------------------------------------------|:---------------------------------------------------------------
|:---------------------------------------|:-----------------------------------------------------|
| 2019 to 2020 | Quarter 1: April to June | Classes 1 & 2: Motorcycles | 367,128 | 24,885 | 33,227 | 15.83% | 9.05% | 17,279 | 4.71% | 40,833 | 11.12% |
| 2019 to 2020 | Quarter 1: April to June | Classes 3 & 4: Cars, vans and passenger vehicles with up to 12 seats | 7,902,799 | 571,809 | 2,000,009 | 32.54% | 25.31% | 740,449 | 9.37% | 1,831,369 | 23.17% |
| 2019 to 2020 | Quarter 1: April to June | Class 5: Private passenger vehicles with more than 12 seats | 12,179 | 671 | 2,903 | 29.35% | 23.84% | 945 | 7.76% | 2,629 | 21.59% |
| 2019 to 2020 | Quarter 1: April to June | Class 7: Goods vehicles between 3,000 and 3,500 kg gross vehicle weight | 198,217 | 16,237 | 65,221 | 41.10% | 32.90% | 25,662 | 12.95% | 55,796 | 28.15% |
| 2019 to 2020 | Quarter 1: April to June | Total | 8,480,323 | 613,602 | 2,101,360 | 32.01% | 24.78% | 784,335 | 9.25% | 1,930,627 | 22.77% |
| 2018 to 2019 | Total | Classes 1 & 2: Motorcycles | 980,543 | 68,778 | 97,204 | 16.90% | 9.90% | nan | nan | nan | nan |
| 2018 to 2019 | Total | Classes 3 & 4: Cars, vans and passenger vehicles with up to 12 seats | 29,560,831 | 2,198,630 | 7,731,619 | 33.60% | 26.20% | nan | nan | nan | nan |
| 2018 to 2019 | Total | Class 5: Private passenger vehicles with more than 12 seats | 47,862 | 2,719 | 11,555 | 29.80% | 24.10% | nan | nan | nan | nan |
| 2018 to 2019 | Total | Class 7: Goods vehicles between 3,000 and 3,500 kg gross vehicle weight | 746,539 | 60,538 | 254,800 | 42.20% | 34.10% | nan | nan | nan | nan |
| 2018 to 2019 | Total | Total | 31,335,775 | 2,330,665 | 8,095,178 | 33.30% | 25.80% | nan | nan | nan | nan |
| 2018 to 2019 | 20 May 2018 to 31 March 2019 | Classes 1 & 2: Motorcycles | 751,027 | 53,524 | 76,668 | 17.30% | 10.20% | nan | nan | nan | nan |
| 2018 to 2019 | 20 May 2018 to 31 March 2019 | Classes 3 & 4: Cars, vans and passenger vehicles with up to 12 seats | 25,306,281 | 1,859,502 | 6,586,903 | 33.40% | 26.00% | nan | nan | nan | nan |
| 2018 to 2019 | 20 May 2018 to 31 March 2019 | Class 5: Private passenger vehicles with more than 12 seats | 41,053 | 2,307 | 9,922 | 29.80% | 24.20% | nan | nan | nan | nan |
| 2018 to 2019 | 20 May 2018 to 31 March 2019 | Class 7: Goods vehicles between 3,000 and 3,500 kg gross vehicle weight | 644,570 | 51,461 | 219,193 | 42.00% | 34.00% | nan | nan | nan | nan |
| 2018 to 2019 | 20 May 2018 to 31 March 2019 | Total | 26,742,931 | 1,966,794 | 6,892,686 | 33.10% | 25.80% | nan | nan | nan | nan |
| 2018 to 2019 | 1 April 2018 to 19 May 2018 | Classes 1 & 2: Motorcycles | 229,516 | 15,254 | 20,536 | 15.60% | 6.60% | nan | nan | nan | nan |
| 2018 to 2019 | 1 April 2018 to 19 May 2018 | Classes 3 & 4: Cars, vans and passenger vehicles with up to 12 seats | 4,254,550 | 339,128 | 1,144,716 | 34.90% | 8.00% | nan | nan | nan | nan |
| 2018 to 2019 | 1 April 2018 to 19 May 2018 | Class 5: Private passenger vehicles with more than 12 seats | 6,809 | 412 | 1,633 | 30.00% | 6.10% | nan | nan | nan | nan |
| 2018 to 2019 | 1 April 2018 to 19 May 2018 | Class 7: Goods vehicles between 3,000 and 3,500 kg gross vehicle weight | 101,969 | 9,077 | 35,607 | 43.80% | 8.90% | nan | nan | nan | nan |
| 2018 to 2019 | 1 April 2018 to 19 May 2018 | Total | 4,592,844 | 363,871 | 1,202,492 | 34.10% | 7.90% | nan | nan | nan | nan |
| 2017 to 2018 | Total | Classes 1 & 2: Motorcycles | 968,338 | 68,982 | 96,408 | 17.10% | 10.00% | nan | nan | nan | nan |
| 2017 to 2018 | Total | Classes 3 & 4: Cars, vans and passenger vehicles with up to 12 seats | 28,877,225 | 2,384,216 | 7,571,216 | 34.50% | 26.20% | nan | nan | nan | nan |
| 2017 to 2018 | Total | Class 5: Private passenger vehicles with more than 12 seats | 47,816 | 2,980 | 11,500 | 30.30% | 25.00% | nan | nan | nan | nan |
| 2017 to 2018 | Total | Class 7: Goods vehicles between 3,000 and 3,500 kg gross vehicle weight | (*\textcolor{missingcolor}{[MISSING]}*) | 64,986 | 241,378 | 43.75% | 34.50% | nan | nan | nan | nan |
| 2017 to 2018 | Total | Total | 30,594,038 | 2,521,164 | 7,920,502 | 34.10% | 25.90% | nan | nan | nan | nan |
| 2016 to 2017 | Total | Classes 1 & 2: Motorcycles | 1,011,080 | 75,240 | 103,734 | 17.70% | 10.30% | nan | nan | nan | nan |
| 2016 to 2017 | Total | Classes 3 & 4: Cars, vans and passenger vehicles with up to 12 seats | 28,684,053 | 2,451,012 | 7,699,812 | 35.40% | 26.80% | nan | nan | nan | nan |
| 2016 to 2017 | Total | Class 5: Private passenger vehicles with more than 12 seats | 47,853 | 3,175 | 11,997 | 31.70% | 25.10% | nan | nan | nan | nan |
| 2016 to 2017 | Total | Class 7: Goods vehicles between 3,000 and 3,500 kg gross vehicle weight | 680,461 | 64,691 | 244,496 | 45.40% | 35.90% | nan | nan | nan | nan |
| 2016 to 2017 | Total | Total | 30,423,447 | 2,594,118 | 8,060,039 | 35.00% | 26.50% | nan | nan | nan | nan |
| 2015 to 2016 | Total | Classes 1 & 2: Motorcycles | 1,003,500 | 75,212 | 107,048 | 18.20% | 10.70% | nan | nan | nan | nan |
| 2015 to 2016 | Total | Classes 3 & 4: Cars, vans and passenger vehicles with up to 12 seats | 28,027,320 | 2,474,710 | 7,827,865 | 36.80% | 27.90% | nan | nan | nan | nan |
| 2015 to 2016 | Total | Class 5: Private passenger vehicles with more than 12 seats | 45,611 | 3,121 | 11,433 | 31.90% | 25.10% | nan | nan | nan | nan |
| 2015 to 2016 | Total | Class 7: Goods vehicles between 3,000 and 3,500 kg gross vehicle weight | 644,353 | 61,547 | 239,971 | 46.80% | 37.20% | nan | nan | nan | nan |
| 2015 to 2016 | Total | Total | 29,720,784 | 2,614,590 | 8,186,317 | 36.30% | 27.50% | nan | nan | nan | nan |
| 2014 to 2015 | Total | Classes 1 & 2: Motorcycles | 1,008,577 | 77,979 | 115,969 | 19.20% | 11.50% | nan | nan | nan | nan |
| 2014 to 2015 | Total | Classes 3 & 4: Cars, vans and passenger vehicles with up to 12 seats | 27,688,292 | 2,542,733 | 8,056,025 | 38.30% | 29.10% | nan | nan | nan | nan |
| 2014 to 2015 | Total | Class 5: Private passenger vehicles with more than 12 seats | 44,805 | 3,258 | 11,673 | 33.30% | 26.10% | nan | nan | nan | nan |
| 2014 to 2015 | Total | Class 7: Goods vehicles between 3,000 and 3,500 kg gross vehicle weight | 613,769 | 59,870 | 240,534 | 48.90% | 39.20% | nan | nan | nan | nan |
| 2014 to 2015 | Total | Total | 29,355,443 | 2,683,840 | 8,424,201 | 37.80% | 28.70% | nan | nan | nan | nan |
| 2013 to 2014 | Total | Classes 1 & 2: Motorcycles | 1,027,707 | 82,650 | 122,345 | 19.90% | 11.90% | nan | nan | nan | nan |
| 2013 to 2014 | Total | Classes 3 & 4: Cars, vans and passenger vehicles with up to 12 seats | 27,481,013 | 2,561,673 | 8,424,279 | 40.00% | 30.70% | nan | nan | nan | nan |
| 2013 to 2014 | Total | Class 5: Private passenger vehicles with more than 12 seats | 47,044 | 3,406 | 13,576 | 36.10% | 28.90% | nan | nan | nan | nan |
| 2013 to 2014 | Total | Class 7: Goods vehicles between 3,000 and 3,500 kg gross vehicle weight | 601,942 | 56,557 | 246,213 | 50.30% | 40.90% | nan | nan | nan | nan |
| 2013 to 2014 | Total | Total | 29,157,706 | 2,704,286 | 8,806,413 | 39.50% | 30.20% | nan | nan | nan | nan |

(*\textbf{\textcolor{completioncolor}{\underline{Completion:}}}*)
(*\textit{\textcolor{thinkcolor}{<think>}}*)
Okay, let's tackle this problem. The missing value is in the 2017 to 2018 financial year, under Class 7: Goods vehicles between 3,000 and 3,500 kg gross vehicle weight, in the "Tests" column. The other years have data in that class, so maybe I can look for a pattern or use the data from adjacent years to estimate the missing value.

Looking at the "Tests" for Class 7 in other years:

- 2013-2014: 601,942
- 2014-2015: 613,769
- 2015-2016: 644,353
- 2016-2017: 680,461
- 2017-2018: [MISSING]
- 2018-2019: 746,539 (Total for the year)
- 2019-2020: 198,217 (But that's only for Quarter 1, so maybe not helpful here)

Wait, the 2018-2019 total is 746,539. Let's check the trend. The numbers are increasing each year from 2013 to 2017. For example:

2013-2014: 601,942
2014-2015: 613,769 (+11,827)
2015-2016: 644,353 (+30,584)
2016-2017: 680,461 (+36,108)
So the trend is an increase each year. If we follow that, the next year (2017-2018) might be around 680,461 + similar increment. But the 2018-2019 total is 746,539. Let's see the difference between 2016-2017 and 2018-2019: 746,539 - 680,461 = 66,078 over two years, so about 33,039 per year. If 2017-2018 is between 680,461 and 746,539, maybe around 680,461 + 33,039 = 713,500? But wait, the 2017-2018's Class 7 Tests is missing, and the Total for 2017-2018 is 30,594,038. Let's check if the sum of the other classes in 2017-2018 adds up to the total minus the missing value.

In 2017-2018 Total row, Tests is 30,594,038. The other classes in that year are:

- Classes 1 & 2: 968,338
- Classes 3 & 4: 28,877,225
- Class 5: 47,816
- Class 7: [MISSING]

So sum of the known classes: 968,338 + 28,877,225 + 47,816 = 29,893,379. The total is 30,594,038. So the Class 7 Tests would be 30,594,038 - 29,893,379 = 700,659. That's the exact value needed to make the total add up. Therefore, the missing value should be 700,659. Let me check if that makes sense with the trend. The previous year (2016-2017) was 680,461, so 700,659 would be an increase of 20,198, which is a bit less than the previous increases (30k and 36k). But since the total is given, this must be the correct value. The confidence here is high because it's derived from the total and the other values. So confidence level 100.
(*\textit{\textcolor{thinkcolor}{</think>}}*)
(*\textit{\textcolor{outputcolor}{\{"value": "700,659", "confidence": "100"\}}}*)
\end{lstlisting}

\clearpage

\begin{lstlisting}[
  title={Example of one trace generated by $\mathcal{T}$ on \gitcsvsmall for \mc training.},
  basicstyle=\ttfamily\footnotesize,
  breaklines=true,
  breakatwhitespace=false,
  columns=fullflexible,
]
(*\textbf{\textcolor{promptcolor}{\underline{Prompt:}}}*)
Please write a concise Python pandas snippet to fill in the missing value in the input table. The missing value is denoted by '(*\textcolor{missingcolor}{[MISSING]}*)'. Assume the table is in a DataFrame named `df`
Only output non-empty `code` if you can write **one** pandas statement that fills **all** values in the target column using a column-level rule (i.e., it generalizes to every value in that column and depends on other column(s), not row indices). If such a general rule does not exist, leave the `code` field empty. The code must be vectorized (no loops or row-wise indexing) and must not hard-code row-specific constants. Do NOT use `.fillna()` with a scalar literal or literal assignment for trivial cases.
During evaluation, all existing values in the target column are masked to `nan`. Therefore do not read from that column in your computation. Return exactly one JSON object:
{"code": "<CODE or empty string>"}

Example:
Input Table:
| A | B | Total |
|---|---|-------|
| 1 | 2 | 3 |
| 4 | (*\textcolor{missingcolor}{[MISSING]}*) | 9 |
{"code": "df['B'] = df['Total'] - df['A']"}

Input Table:
| cut | color | clarity | carat | depth | table | price | x | y | z |
|:----------|:--------|:----------|:--------|:--------|:--------|:--------|:-----|:----------|:-----|
| Fair | E | VS2 | 0.22 | 65.1 | 61.0 | 337 | 3.87 | 3.78 | 2.49 |
| Good | E | VS1 | 0.23 | 56.9 | 65.0 | 327 | 4.05 | (*\textcolor{missingcolor}{[MISSING]}*) | 2.31 |
| Good | J | SI2 | 0.31 | 63.3 | 58.0 | 335 | 4.34 | 4.35 | 2.75 |
| Ideal | E | SI2 | 0.23 | 61.5 | 55.0 | 326 | 3.95 | 3.98 | 2.43 |
| Premium | E | SI1 | 0.21 | 59.8 | 61.0 | 326 | 3.89 | 3.84 | 2.31 |
| Premium | I | VS2 | 0.29 | 62.4 | 58.0 | 334 | 4.2 | 4.23 | 2.63 |
| Very Good | H | SI1 | 0.26 | 61.9 | 55.0 | 337 | 4.07 | 4.11 | 2.53 |
| Very Good | H | VS1 | 0.23 | 59.4 | 61.0 | 338 | 4.0 | 4.05 | 2.39 |
| Very Good | I | VVS1 | 0.24 | 62.3 | 57.0 | 336 | 3.95 | 3.98 | 2.47 |
| Very Good | J | VVS2 | 0.24 | 62.8 | 57.0 | 336 | 3.94 | 3.96 | 2.48 |

(*\textbf{\textcolor{completioncolor}{\underline{Completion:}}}*)
(*\textit{\textcolor{thinkcolor}{<think>}}*)
Okay, let's see. The user wants a pandas code snippet to fill the missing value in the input table. The missing value is in the 'y' column. The code needs to be a single pandas statement that fills all missing values in the target column using a column-level rule, without hardcoding row-specific constants. Also, during evaluation, all existing values in the target column are masked as NaN, so the code shouldn't rely on existing values in that column.

Looking at the input table, the columns include 'x', 'y', 'z', and others. The missing value is in the 'y' column. Let me think about possible relationships between these columns. In diamond datasets, sometimes x, y, z are dimensions (length, width, depth), and carat is weight. But maybe there's a formula that relates these. For example, maybe the product of x, y, z relates to carat. But I'm not sure. Alternatively, perhaps the missing 'y' can be calculated from other columns.

Wait, in the first example, the missing 'B' was filled by Total - A. So maybe here, there's a similar relationship. Let's check the existing rows. For example, in the first row, x=3.87, y=3.78, z=2.49. Let's see if x * y * z relates to carat. 3.87 * 3.78 * 2.49 ≈ let's calculate that. 3.87 *3.78 is about 14.6286, multiplied by 2.49 gives around 36.44. The carat is 0.22. Hmm, but 36.44 is way larger than 0.22. Maybe that's not the case. Alternatively, maybe the product of x, y, z multiplied by some factor equals the carat. But that might not be a standard formula.

Alternatively, maybe the depth is calculated as (z * 2) / (x + y) * 100. Wait, depth percentage is calculated as (z / mean(x, y)) * 100. Let me check. For the first row, depth is 65.1. z is 2.49. x is 3.87, y is 3.78. The average of x and y is (3.87 +3.78)/2 = 3.825. Then z / (average) *100 = 2.49 / 3.825 *100 ≈ 65.1, which matches the depth. So depth = (z / ((x + y)/2 )) * 100. So the formula for depth is (2*z)/(x + y) *100. Wait, no. Wait, depth is calculated as (z / ((x + y)/2 )) * 100. So depth = (2*z)/(x + y) * 100. So rearranging that formula, we can solve for y. Let's see.

Given that depth = (2*z)/(x + y) * 100, then solving for y:

depth = 200*z / (x + y)

=> x + y = 200*z / depth

=> y = (200*z / depth) - x

So if that's the case, then the missing 'y' can be calculated using this formula. Let's check the existing rows to see if this holds.

Take the first row: depth=65.1, z=2.49, x=3.87.

Compute (200 * 2.49) / 65.1 = 498 /65.1 ≈ 7.65. Then subtract x (3.87) gives 7.65 -3.87 = 3.78, which matches the y value.

Another row: the second row has depth=56.9, z=2.31, x=4.05. Let's compute (200*2.31)/56.9 = 462 /56.9 ≈ 8.12. Subtract x (4.05) gives 8.12-4.05=4.07. But the missing y is in this row. Wait, the existing y in the second row is missing. Wait, the second row's y is missing. But according to the formula, if we compute (200*z/depth) -x, then for the second row, that would be (200*2.31)/56.9 -4.05. Let's compute that. 200*2.31=462. 462/56.9 ≈ 8.12. 8.12 -4.05=4.07. So the missing y should be 4.07. But in the input table, the second row's y is missing, so the code would fill it with that value.

Looking at other rows. For example, the third row: x=4.34, y=4.35, z=2.75, depth=63.3. Let's check. (200*2.75)/(4.34 +4.35) = 550 /8.69 ≈63.3. Which matches the depth. So the formula holds.

So the general rule is y = (200 * z / depth) - x. Therefore, the missing 'y' can be filled using this formula.

So the code would be: df['y'] = (200 * df['z'] / df['depth']) - df['x']

This is a column-level rule, vectorized, and doesn't use any row-specific constants. Also, during evaluation, existing values in 'y' are masked, but since the code doesn't read from 'y', it's okay.

Therefore, the correct code is as above.
(*\textit{\textcolor{thinkcolor}{</think>}}*)
(*\textit{\textcolor{outputcolor}{\{"code": "df['y'] = (200 * df['z'] / df['depth']) - df['x']"\}}}*)
\end{lstlisting}

\clearpage
\subsection{Specialist Inference Examples}
\label{appendix:specialist-examples}
We show example predictions produced by each specialist (\mk, \mr, \mc) 
at inference time.
\vspace{5pt}

\begin{lstlisting}[
  label={lst:specialist-knowledge},
  title={Example of Knowledge 
  Specialist (\mk) on \excelFiftyK benchmark. This table records Summer Olympic Games statistics from 1896 to 2008, including the number of events, participating countries, and competitors by gender. The missing value of \codeq{161} events for the \emph{1964 Tokyo Games} is recoverable from external knowledge},
  basicstyle=\ttfamily\footnotesize,
  breaklines=true,
  breakatwhitespace=false,
  columns=fullflexible,
]
(*\textbf{\textcolor{promptcolor}{\underline{Prompt:}}}*)
Please fill in the missing value in the input table. The missing value is denoted by '(*\textcolor{missingcolor}{[MISSING]}*)'. Please return the value filled in JSON format: {"value": "filled_value"}.

Input Table:
| Year | Participating Countries | Events | Competitors (Men) | Competitors (Women) |
|:-------|:--------------------------|:----------|:--------------------|:----------------------|
| 1896 | 14 | 43 | 241 | 0 |
| 1900 | 24 | 85 | 975 | 22 |
| 1904 | 12 | 94 | 645 | 6 |
| 1908 | 22 | 110 | 1971 | 37 |
| 1912 | 28 | 102 | 2359 | 48 |
| 1920 | 29 | 156 | 2561 | 65 |
| 1924 | 44 | 126 | 2954 | 135 |
| 1928 | 46 | 109 | 2606 | 277 |
| 1932 | 37 | 117 | 1206 | 126 |
| 1936 | 49 | 129 | 3632 | 331 |
| 1948 | 59 | 136 | 3714 | 390 |
| 1952 | 69 | 149 | 4436 | 519 |
| 1956 | 72 | 151 | 2938 | 376 |
| 1960 | 83 | 150 | 4727 | 611 |
| 1964 | 93 | (*\textcolor{missingcolor}{[MISSING]}*) | 4473 | 678 |
| 1968 | 112 | 172 | 4735 | 781 |
| 1972 | 121 | 195 | 6075 | 1059 |
| 1976 | 92 | 198 | 4824 | 1260 |
| 1980 | 80 | 203 | 4064 | 1115 |
| 1984 | 140 | 221 | 5263 | 1566 |
| 1988 | 159 | 237 | 6197 | 2194 |
| 1992 | 169 | 257 | 6652 | 2704 |
| 1996 | 197 | 271 | 6806 | 3512 |
| 2000 | 199 | 300 | 6582 | 4069 |
| 2004 | 201 | 301 | 6296 | 4329 |
| 2008 | 204 | 302 | 6305 | 4637 |

(*\textbf{\textcolor{completioncolor}{\underline{Completion:}}}*)
(*\textit{\textcolor{thinkcolor}{<think>}}*)

(*\textit{\textcolor{thinkcolor}{</think>}}*)
(*\textit{\textcolor{outputcolor}{\{"value": "161"\}}}*)
\end{lstlisting}

\clearpage
\begin{lstlisting}[
  label={lst:specialist-reasoning},
  title={Example of Reasoning Specialist (\mr) on \pbi benchmark},
  basicstyle=\ttfamily\footnotesize,
  breaklines=true,
  breakatwhitespace=false,
  columns=fullflexible,
]
(*\textbf{\textcolor{promptcolor}{\underline{Prompt:}}}*)
Please fill in the missing value in the input table and provide your confidence level as an integer between 0 (no confidence) and 100 (full confidence). The missing value is denoted by '(*\textcolor{missingcolor}{[MISSING]}*)'. Please return the value filled in JSON format: {"value": "filled_value", "confidence": "confidence_level"}.

Input Table:
| VesselVisitCode | UnitCount | SummedUnitCount |
|:------------------|:------------|:------------------|
| SCB3129 | 377 | 642 |
| SCB3129 | 265 | 642 |
| ALO3372 | 304 | 533 |
| ALO3372 | 229 | 533 |
| MDY3433 | 574 | 1415 |
| MDY3433 | 841 | 1415 |
| INN3414 | 253 | 563 |
| INN3414 | 310 | 563 |
| MBU3696 | (*\textcolor{missingcolor}{[MISSING]}*) | 443 |
| MBU3696 | 270 | 443 |
| BLI3440 | 235 | 235 |

(*\textbf{\textcolor{completioncolor}{\underline{Completion:}}}*)
(*\textit{\textcolor{thinkcolor}{<think>}}*)
Okay, let's see. I need to find the missing UnitCount value for MBU3696 where the SummedUnitCount is 443. The table has two entries for each VesselVisitCode, and the SummedUnitCount is the total of those two entries.

Looking at the other rows, like SCB3129: 377 + 265 = 642, which matches the SummedUnitCount. Same with ALO3372: 304 + 229 = 533. So the pattern is that each VesselVisitCode has two entries, and their UnitCount values add up to the SummedUnitCount.

For MBU3696, there are two entries. One is missing, the other is 270, and the SummedUnitCount is 443. So the missing value plus 270 should equal 443. Let me do the math: 443 - 270 = 173. So the missing UnitCount is 173. Let me check the other entries to confirm. BLI3440 only has one entry with SummedUnitCount 235, which matches the single UnitCount. So the pattern holds. Therefore, the missing value is 173.
(*\textit{\textcolor{thinkcolor}{</think>}}*)
(*\textit{\textcolor{outputcolor}{\{"value": "173", "confidence": 98\}}}*)
\end{lstlisting}

\clearpage
\begin{lstlisting}[
  label={lst:specialist-coding},
  title={Example of Coding Specialist (\mc) on \ar benchmark},
  basicstyle=\ttfamily\footnotesize,
  breaklines=true,
  breakatwhitespace=false,
  columns=fullflexible,
]
(*\textbf{\textcolor{promptcolor}{\underline{Prompt:}}}*)
Please write a concise Python pandas snippet to fill in the missing value in the input table. The missing value is denoted by '(*\textcolor{missingcolor}{[MISSING]}*)'. Assume the table is in a DataFrame named `df`
Only output non-empty `code` if you can write **one** pandas statement that fills **all** values in the target column using a column-level rule (i.e., it generalizes to every value in that column and depends on other column(s), not row indices). If such a general rule does not exist, leave the `code` field empty. The code must be vectorized (no loops or row-wise indexing) and must not hard-code row-specific constants. Do NOT use `.fillna()` with a scalar literal or literal assignment for trivial cases.
During evaluation, all existing values in the target column are masked to `nan`. Therefore do not read from that column in your computation. Return exactly one JSON object:
{"code": "<CODE or empty string>"}

Example:
Input Table:
| A | B | Total |
|---|---|-------|
| 1 | 2 | 3 |
| 4 | (*\textcolor{missingcolor}{[MISSING]}*) | 9 |
{"code": "df['B'] = df['Total'] - df['A']"}

Input Table:
| DISTRICT | COUNTY | HIGHWAY | C C S J | DATE FINAL ESTIMATE PAID | CONTRACT AWARD | CHANGE ORDERS | AMOUNT PAID | UNDER/OVER BUDGET | CONTRACT DAYS | DAYS ADDED | DAYS USED | UNDER/OVER SCHEDULE |
|:-----------|:----------|:----------|:----------|:---------------------------|:-----------------|:------------------|:------------------
|:--------------------|:----------------|:-------------|:------------|:----------------------|
| ABILENE | CALLAHAN | FM 18 | 611022 | 43405 | 1810774.89 | 209253.46 | 2054601.63 | 34573.28 | 63 | 2 | 64 | -1 |
| ABILENE | CALLAHAN | IH 20 | 701054 | 43480 | 8248492.44 | 77820.66 | 8964788.599999998 | 638475.4999999992 | 303 | 0 | 259 | -44 |
| ABILENE | CALLAHAN | CR | 90834022 | 43371 | 1472447.5 | 54790.0 | 1527827.5 | 590.0 | 241 | 0 | 201 | -40 |
| ABILENE | HASKELL | US 277 | 15704051 | 43367 | 708783.0 | 0.0 | 708851.25 | 68.25 | 88 | 15 | 100 | -3 |
| ABILENE | HOWARD | IH 20 | 506119 | 43353 | 1664669.16 | 28652.91 | 1644893.16 | -48428.91 | 93 | 0 | 93 | 0 |
| ABILENE | HOWARD | IH 20 | 506120 | 43409 | 1637144.66 | 0.0 | 1637988.27 | 843.6100000001024 | 80 | 0 | 43 | -37 |
| ABILENE | JONES | US 83 | 3305091 | 43371 | 5078876.0 | 140087.92 | 5786525.67 | 567561.7499999999 | 110 | 0 | 82 | -28 |
| ABILENE | JONES | US 83 | 3305092 | 43493 | 9187304.69 | 231846.87 | 8856936.859999998 | -562214.7000000001 | 88 | 0 | 82 | -6 |
| ABILENE | MITCHELL | BS 208B | 33202026 | 43399 | 1291681.52 | 123625.96 | 1354788.57 | -60518.90999999996 | 50 | 0 | 46 | -4 |
| ABILENE | NOLAN | BI 20-L | 614004 | 43468 | 1848432.22 | 100935.04 | 1956428.5 | 7061.240000000033 | 90 | 50 | 140 | 0 |
| ABILENE | TAYLOR | IH 20 | 606099 | 43378 | 14639000.0 | 225180.5 | 14719347.87 | -144832.63000000082 | 445 | 0 | 361 | -84 |
| ABILENE | TAYLOR | SH 351 | 1101036 | 43501 | 1708497.38 | -1630.0 | 1740552.03 | 33684.65000000014 | 67 | 0 | 67 | 0 |
| ABILENE | TAYLOR | VA | 90800087 | 43511 | 1386588.15 | 963832.16 | 2412691.41 | 62271.10000000021 | 60 | 0 | 56 | -4 |
| ABILENE | TAYLOR | SL 322 | 239801051 | 43416 | 763962.2 | 33932.75 | 795254.2 | -2640.75 | 93 | 0 | 77 | -16 |
| AMARILLO | LIPSCOMB | SH 15 | 35501048 | 43453 | 8150619.8 | 352604.3 | 8589168.42 | 85944.32000000012 | 77 | 19 | 159 | 63 |
| AMARILLO | OCHILTREE | US 83 | 3002044 | 43399 | 2999163.7 | 169845.75 | 3410361.06 | 241351.60999999987 | 131 | 4 | 136 | 1 |
| AMARILLO | POTTER | VA | 90400180 | 43355 | 482197.48 | 0.0 | 467648.43 | -14549.049999999988 | 36 | 0 | 26 | -10 |
| ATLANTA | BOWIE | US 67 | 1011069 | 43487 | 1398421.5 | 70170.79 | 1473788.14 | 5195.849999999904 | 192 | 47 | 235 | -4 |
| ATLANTA | BOWIE | US 67 | 1011070 | 43454 | 3481443.5 | 236157.9 | 3939018.01 | 221416.60999999972 | 55 | 0 | 49 | -6 |
| ATLANTA | BOWIE | US 259 | 8504035 | 43448 | 5371369.95 | 75945.64999999998 | 5652720.5 | 205404.89999999985 | 317 | 0 | 248 | -69 |
| ATLANTA | BOWIE | US 71 | 21702035 | 43487 | 435854.53 | 0.0 | 438497.75 | 2643.219999999972 | 160 | 0 | 115 | -45 |
| ATLANTA | BOWIE | US 59 | 21801095 | 43354 | 175785.35 | 0.0 | 176388.91 | 603.5599999999977 | 72 | 0 | 54 | -18 |
| ATLANTA | BOWIE | IH 30 | 61006080 | 43354 | 640525.98 | 0.0 | 640256.7 | -269.28000000002794 | 96 | 0 | 91 | -5 |
| ATLANTA | BOWIE | IH 30 | 61006088 | 43509 | 798693.0 | 20194.1 | 843502.68 | 24615.580000000053 | 69 | 0 | 61 | -8 |
| ATLANTA | CASS | SH 77 | 27703027 | 43515 | 3512191.16 | 4833.0 | 3672660.12 | 155635.95999999996 | 70 | 0 | 76 | 6 |
| ATLANTA | HARRISON | US 59 | 6301095 | 43367 | 2387610.51 | 2100.0 | 2533641.74 | 143931.23000000045 | 50 | 0 | 47 | -3 |
| ATLANTA | HARRISON | IH 20 | 49508100 | 43474 | 13599302.66 | 661867.06 | 15034808.84 | 773639.1199999996 | 250 | 0 | 231 | -19 |
| ATLANTA | HARRISON | FM 2625 | 84307016 | 43474 | 2284633.31 | 35126.0 | 2282445.54 | -37313.77000000002 | 124 | 0 | 124 | 0 |
| ATLANTA | HARRISON | FM 1997 | 191902037 | 43354 | 823436.65 | 8796.0 | 831627.29 | -605.359999999986 | 114 | 1 | 108 | -7 |
| ATLANTA | PANOLA | US 59 | 6310013 | 43371 | 260188.2 | 0.0 | 260229.03 | 40.8299999999872 | 90 | 0 | 49 | -41 |
| ATLANTA | PANOLA | SH 43 | 20704036 | 43360 | 2083697.62 | 25113.86 | 2109751.84 | 940.3599999997384 | 40 | 0 | 38 | -2 |
| ATLANTA | PANOLA | SH 149 | 39303034 | 43453 | 4016607.46 | 18360.0 | 4098014.15 | 63046.68999999994 | 82 | 5 | 99 | 12 |
| ATLANTA | TITUS | SH 49 | 24801072 | 43383 | 418186.6 | 0.0 | 418276.6 | 90.0 | 161 | 0 | 143 | -18 |
| ATLANTA | TITUS | IH 30 | 61003079 | 43476 | 33725327.54 | 4178284.48 | 40740908.03 | 2837296.010000002 | 595 | 166 | 742 | -19 |
| ATLANTA | UPSHUR | US 271 | 24804068 | 43419 | 370150.0 | 34408.9 | 385304.5 | -19254.4 | 90 | 25 | 100 | -15 |
| ATLANTA | UPSHUR | SH 154 | 40201023 | 43368 | 1158627.94 | 0.0 | 1115842.27 | -42785.669999999925 | 76 | 0 | 108 | 32 |
| AUSTIN | BASTROP | SH 71 | 26505079 | 43515 | 5983021.72 | -272929.33 | 6182508.67 | 472416.2800000002 | 95 | 9 | 104 | 0 |
| AUSTIN | BURNET | SH 29 | 15101051 | 43500 | 612820.24 | 22165.28 | 652240.67 | 17255.150000000052 | 70 | 0 | 69 | -1 |
| AUSTIN | BURNET | US 183 | 27302024 | 43507 | 1383923.15 | -72930.0 | 1189443.7 | -121549.44999999995 | 17 | 0 | 9 | -8 |
| AUSTIN | GILLESPIE | US 87 | 7201052 | 43504 | 963063.11 | 1950.0 | 1017530.5 | 52517.390000000014 | 240 | 0 | 179 | -61 |
| AUSTIN | GILLESPIE | US 87 | 7201053 | 43500 | 885285.24 | -43185.51 | 789603.32 | -52496.41000000004 | 37 | 0 | 24 | -13 |
| AUSTIN | HAYS | IH 35 | 1602145 | 43420 | 8969400.82 | 801053.36 | 10083115.66 | 312661.47999999986 | 270 | 112 | 382 | 0 |
| AUSTIN | HAYS | IH 35 | 1602148 | 43500 | 1054131.89 | 730537.39 | 1860307.11 | 75637.83000000019 | 28 | 8 | 28 | -8 |
| AUSTIN | HAYS | SH 21 | 47102071 | 43473 | 630023.0 | -65171.97 | 584321.82 | (*\textcolor{missingcolor}{[MISSING]}*) | 30 | 0 | 28 | -2 |
| AUSTIN | HAYS | CR | 91433070 | 43385 | 149765.0 | -3427.0 | 149797.0 | 3459.0 | 38 | 0 | 38 | 0 |
| AUSTIN | LEE | US 290 | 11407080 | 43417 | 844010.32 | 17208.17 | 894330.52 | 33112.03000000007 | 100 | 0 | 74 | -26 |
| AUSTIN | LEE | US 290 | 11407082 | 43353 | 1336982.72 | 57131.21 | 1378273.43 | -15840.500000000036 | 118 | 0 | 85 | -33 |
| AUSTIN | LEE | CR | 91421031 | 43370 | 313217.77 | -10103.3 | 293098.71 | -10015.759999999998 | 72 | 0 | 61 | -11 |
| AUSTIN | TRAVIS | IH 35 | 1601114 | 43376 | 630022.0 | -686.0 | 625084.0 | -4252.0 | 120 | 0 | 86 | -34 |
| AUSTIN | TRAVIS | IH 35 | 1601117 | 43381 | 9232872.16 | 1445059.43 | 11062681.36 | 384749.76999999926 | 306 | 41 | 347 | 0 |
| AUSTIN | TRAVIS | US 290 | 11308077 | 43524 | 1856165.1 | 459907.97 | 2538897.15 | 222824.07999999984 | 87 | 64 | 149 | -2 |
| AUSTIN | TRAVIS | US 290 | 11402100 | 43448 | 834000.0 | 0.0 | 829619.8 | -4380.199999999953 | 120 | 0 | 109 | -11 |
| AUSTIN | TRAVIS | US 183 | 15109145 | 43349 | 2711157.55 | -275101.91 | 2539755.7 | 103700.06000000036 | 54 | 0 | 74 | 20 |
| AUSTIN | TRAVIS | SH 130 | 44006014 | 43396 | 642322.0 | 0.0 | 662521.0 | 20199.0 | 60 | 0 | 60 | 0 |
| AUSTIN | TRAVIS | VA | 91400319 | 43445 | 1113486.73 | 13123.36 | 1191395.92 | 64785.829999999936 | 365 | 0 | 365 | 0 |
| AUSTIN | TRAVIS | VA | 91400320 | 43418 | 452778.7 | -6395.0 | 529117.09 | 82733.38999999996 | 77 | 0 | 47 | -30 |
| AUSTIN | TRAVIS | VA | 91400393 | 43462 | 496481.33 | 2507.73 | 520193.97 | 21204.90999999996 | 180 | 0 | 129 | -51 |

(*\textbf{\textcolor{completioncolor}{\underline{Completion:}}}*)
(*\textit{\textcolor{thinkcolor}{<think>}}*)
Okay, let's see. The user provided a table with a missing value in the 'UNDER/OVER BUDGET' column for one of the rows. The task is to write a pandas code snippet that fills in the missing value using a column-level rule, without hardcoding row-specific values and using vectorized operations.

First, I need to figure out how the 'UNDER/OVER BUDGET' is calculated based on the other columns. Looking at the existing data, maybe it's derived from other columns. Let's check the example rows.

For instance, in the first row, 'AMOUNT PAID' is 2054601.63, 'CONTRACT AWARD' is 1810774.89, and 'CHANGE ORDERS' is 209253.46. Adding 'CONTRACT AWARD' and 'CHANGE ORDERS' gives 1810774.89 + 209253.46 = 2020028.35. The 'AMOUNT PAID' is 2054601.63, which is higher. The 'UNDER/OVER BUDGET' is 34573.28, which is 2054601.63 - (1810774.89 + 209253.46) = 2054601.63 - 2020028.35 = 34573.28. So that matches. Another row: 'CONTRACT AWARD' 8248492.44 + 'CHANGE ORDERS' 77820.66 = 8326313.1. 'AMOUNT PAID' is 8964788.599999998. The difference is 8964788.6 - 8326313.1 = 638475.5, which matches the 'UNDER/OVER BUDGET' value. So the formula seems to be 'AMOUNT PAID' minus ('CONTRACT AWARD' + 'CHANGE ORDERS').

So the missing value in the 'UNDER/OVER BUDGET' column for the row where it's [MISSING] can be calculated as 'AMOUNT PAID' - (CONTRACT AWARD + CHANGE ORDERS).

But the problem states that during evaluation, all existing values in the target column are masked. So the code must not read from the 'UNDER/OVER BUDGET' column. Therefore, the code should compute the value for the entire column using the formula, replacing the missing values.

So the pandas code would be:

df['UNDER/OVER BUDGET'] = df['AMOUNT PAID'] - (df['CONTRACT AWARD'] + df['CHANGE ORDERS'])

This is a vectorized operation, applies to all rows, doesn't use row indices, and doesn't hardcode any values. It uses the columns that are related. Therefore, this should be the correct code.
(*\textit{\textcolor{thinkcolor}{</think>}}*)
(*\textit{\textcolor{outputcolor}{\{"code": "df['UNDER/OVER BUDGET'] = df['AMOUNT PAID'] - (df['CONTRACT AWARD'] + df['CHANGE ORDERS'])"\}}}*)
\end{lstlisting}
}


\end{sloppy}

\end{document}